%% file: main.tex
\documentclass[sigconf]{acmart}
\usepackage{tikz}
\usepackage[edges]{forest}

\usepackage{listings}
\usepackage{color}
\usepackage{fancyvrb}
\usepackage{tcolorbox}
\usepackage{enumitem}
\usepackage{pifont}
\usepackage{booktabs}
\usepackage{multirow}
\usepackage{graphicx}

\usepackage[normalem]{ulem}
\useunder{\uline}{\ul}{}

\usepackage{xcolor}
\usepackage{xcolor,colortbl}
\definecolor{hidden-draw}{RGB}{0,0,0}

\AtBeginDocument{%
  }

\setcopyright{acmlicensed}
\copyrightyear{2018}
\acmYear{2018}
\acmDOI{XXXXXXX.XXXXXXX}
\acmConference[Conference acronym 'XX]{Make sure to enter the correct
  conference title from your rights confirmation email}{June 03--05,
  2018}{Woodstock, NY}
\acmISBN{978-1-4503-XXXX-X/2018/06}

\begin{document}

\title{A Survey of Foundation Models for IoT: Taxonomy and Criteria-Based Analysis}


\author{Hui Wei}
\email{huiwei2@ucmerced.edu}
\affiliation{%
  \institution{Univ. of California, Merced}
  \city{Merced}
  \state{CA}
  \country{USA}
}

\author{Dong Yoon Lee}
\authornote{Both authors contributed equally to this research.}
\email{dlee267@ucmerced.edu}
\affiliation{%
  \institution{Univ. of California, Merced}
  \city{Merced}
  \state{CA}
  \country{USA}
}

\author{Shubham Rohal}
\authornotemark[1]
\email{srohal@ucmerced.edu}
\affiliation{%
  \institution{Univ. of California, Merced}
  \city{Merced}
  \state{CA}
  \country{USA}
}

\author{Zhizhang Hu}
\email{zhu42@ucmerced.edu}
\affiliation{%
  \institution{Univ. of California, Merced}
  \city{Merced}
  \state{CA}
  \country{USA}
}

\author{Ryan Rossi}
\email{ryrossi@adobe.com}
\affiliation{%
  \institution{Adobe Research}
  \city{San Jose}
  \state{CA}
  \country{USA}
}

\author{Shiwei Fang}
\email{shfang@augusta.edu}
\affiliation{%
  \institution{Augusta University}
  \city{Augusta}
  \state{GA}
  \country{USA}
}

\author{Shijia Pan}
\email{span24@ucmerced.edu}
\affiliation{%
  \institution{Univ. of California, Merced}
  \city{Merced}
  \state{CA}
  \country{USA}
}

\renewcommand{\shortauthors}{Wei et al.}

\input{sections/1abstract}

\ccsdesc[500]{General and reference~Surveys and overviews}
\ccsdesc[500]{Computing methodologies~Artificial intelligence}
\ccsdesc[500]{Computer systems organization~Embedded and cyber-physical systems}

\keywords{Foundation Model, Internet-of-Things, Survey, Critieria-Based Analysis}


\maketitle

\input{sections/2introduction}
\input{sections/taxonomy/tree_chart}
\input{sections/3related_work}
\input{sections/4taxonomy}
\input{sections/5discussion}
\input{sections/6conclusion}

\input{sections/7acknowledgements}

\bibliographystyle{ACM-Reference-Format}
\bibliography{references}

\appendix
\input{sections/8appendices}

\end{document}

%% file: sections/1abstract.tex
\begin{abstract}
  Foundation models have gained growing interest in the IoT domain due to their reduced reliance on labeled data and strong generalizability across tasks, which address key limitations of traditional machine learning approaches. However, most existing foundation model based methods are developed for specific IoT tasks, making it difficult to compare approaches across IoT domains and limiting guidance for applying them to new tasks. This survey aims to bridge this gap by providing a comprehensive overview of current methodologies and organizing them around four shared performance objectives by different domains: \textbf{efficiency}, \textbf{context-awareness}, \textbf{safety}, and \textbf{security \& privacy}. For each objective, we review representative works, summarize commonly-used techniques and evaluation metrics. This objective-centric organization enables meaningful cross-domain comparisons and offers practical insights for selecting and designing foundation model based solutions for new IoT tasks. We conclude with key directions for future research to guide both practitioners and researchers in advancing the use of foundation models in IoT applications.
\end{abstract}

%% file: sections/2introduction.tex
\section{Introduction}
\label{sec:introduction}

Machine learning (ML) models have been widely adopted in IoT applications to enable more convenient, automated, and intelligent solutions across diverse domains (e.g., smart cities \citep{ullah2020applications, mehta2022machine}, autonomous driving \citep{bachute2021autonomous, grigorescu2020survey, dogan2011autonomous}, smart agriculture \citep{benos2021machine, sharma2020machine, liakos2018machine}, and precision health \citep{javaid2022significance, habehh2021machine, shailaja2018machine}).
However, existing approaches that rely on traditional ML models (e.g., models trained on \emph{task-specific} datasets \cite{xu2024llm}; deep learning models are also considered as traditional ML in this work) face two key limitations: \textbf{(C1) Dependence on Labeled data}: Most of proposed methods depend on supervised learning, which require large volumes of labeled data to achieve high performance \citep{xu2021limu, bian2022machine, bzai2022machine}. However, IoT data are often not human-interpretable, making it difficult to obtain sufficient high-quality annotations. \textbf{(C2) Poor Generalization Across Contexts}: Even when labeled data are available, IoT data are highly heterogeneous. As a result, models trained in one context (e.g., specific environments or applications) often fail to generalize to others, limiting scalability and cross-domain applicability \cite{ji2024hargpt, chen2024towards, xue2024leveraging}.

In order to address these two challenges, foundation models \citep{bommasani2021opportunities, zhou2024comprehensive} (e.g., large language models) have gained increasing attention and adoption in the IoT domain. First, foundation models (FMs) rely on self-supervised training instead of traditional supervised approaches, which addresses Challenge C1 by reducing the dependency on large labeled IoT datasets.
Second, the training data for FMs typically span multiple domains and contexts, which allows FMs to learn generalizable, context-invariant representations, making them more adaptable to a wide range of downstream tasks. This mitigates Challenge C2.

To advance the field and guide future research, prior surveys \citep{baris2025foundation, xu2024llm, kok2024iot, siam2025artificial, bhat2025llm, sarhaddi2025llms} have summarized FM-based approaches for IoT tasks, along with evaluation metrics and benchmark datasets, primarily from the perspective of \emph{specific application domains} (e.g., healthcare, robotics, smart homes). However, our review of the cited papers reveals a key but previously overlooked insight: 
\emph{while tasks within the same domain may require different techniques, similar techniques can often be applied across domains to address shared objectives}
(e.g., reducing training time, personalizing outputs). Organizing the literature solely by application area obscures these cross-domain goals and the research opportunities they present. Moreover, proposed methods targeting the same objective are often evaluated on different tasks, without comparison to other approaches targeting the same objective. \emph{This inconsistent evaluation makes it difficult to assess which approaches are most effective across different scenarios, providing limited insights for addressing the same objective in new tasks.}

To address the limitations identified above, this paper categorizes current research on FMs for IoT around four key performance criteria that serve as shared objectives across diverse application domains:
(1) \textbf{Efficiency},
(2) \textbf{Context-awareness},
(3) \textbf{Safety} (different from Security in this paper. Please see Section \ref{sec:safety} for more details), and
(4) \textbf{Security and privacy}.
By organizing the literature around these common objectives rather than specific tasks, we aim to provide a clearer understanding of how foundation models are being leveraged across the IoT landscape. This approach enables the community to (i) identify shared performance objectives across diverse IoT tasks, (ii) facilitate meaningful cross-domain comparisons, and (iii) promote the design of more generalizable and effective solutions.
To support this goal, we address the following research questions:
\begin{enumerate}
    \item What \emph{methodologies} have been proposed to improve each performance criterion?
    \item What \emph{metrics} are used to evaluate these criteria?
    \item What \emph{evaluation strategies} are commonly adopted when applying FMs to IoT tasks?
\end{enumerate}
Finally, based on our analysis, we identify additional gaps in the literature and suggest future research directions to further advance the field.

The structure of this paper is as follows: Section \ref{sec:related_work} reviews and compares our survey with existing related work. Section \ref{sec:fm_iot_foundations} introduces three fundamental paradigms and frameworks for applying foundation models to IoT tasks, providing essential background for readers new to the field. Sections \ref{sec:efficiency} through \ref{sec:security_privacy} examine the four key performance criteria along with commonly used approaches to improve each. Section \ref{sec:evaluation} reviews the evaluation metrics used in the current literature for each criterion and discusses the strategies employed to assess FM-based solutions in IoT applications. Finally, Section \ref{sec:discussion} discusses existing research limitations and outlines future directions. Figure \ref{fig:taxonomy} presents a taxonomy of this paper and representative techniques. 

For readers who are already familiar with foundation models, we recommend proceeding directly to the sections of interest.
For those who are familiar with IoT but new to foundation models, we encourage a thorough reading of the entire paper, with particular emphasis on Section \ref{sec:fm_iot_foundations}, which introduces the essential background relevant to this emerging area.

%% file: sections/taxonomy/tree_chart.tex
\tikzstyle{my-box}= [
    rectangle,
    draw=hidden-draw,
    rounded corners,
    text opacity=1,
    minimum height=1.5em,
    minimum width=5em,
    inner sep=2pt,
    align=center,
    fill opacity=.5,
]
\tikzstyle{leaf}=[my-box, minimum height=1.5em,
    fill=blue!15, text=black, align=left,font=\large,
    inner xsep=2pt,
    inner ysep
=4pt,
]
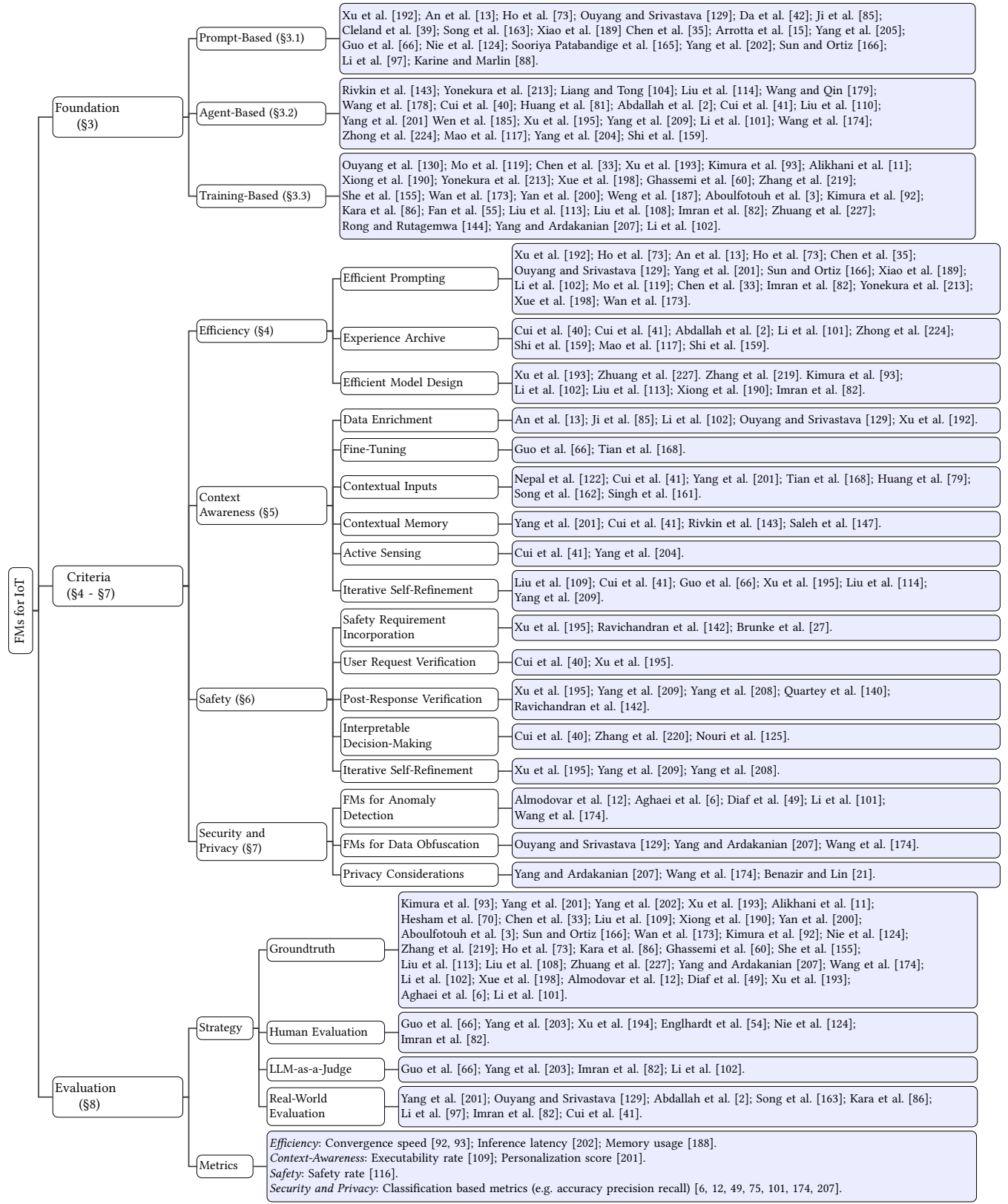
\begin{figure*}[htbp!]
    \centering
    \resizebox{\textwidth}{!}{
        \begin{forest}
            forked edges,
            for tree={
                grow=east,
                reversed=true,
                anchor=base west,
                parent anchor=east,
                child anchor=west,
                base=left,
                font=\Large,
                rectangle,
                draw=hidden-draw,
                rounded corners,
                align=left,
                minimum width=4em,
                edge+={darkgray, line width=1pt},
                s sep=3pt,
                inner xsep=2pt,
                inner ysep=3pt,
                ver/.style={rotate=90, child anchor=north, parent anchor=south, anchor=center},
            },
            where level=1{text width=11.0em,font=\Large,, align=center}{},
            where level=2{text width=11.0em,font=\large,}{},
            where level=3{text width=11.0em,font=\large,}{},
            where level=4{text width=11.0em,font=\large,}{},
        [ \;\;FMs for IoT\;\; , ver
                [Foundation\\(\S \ref{sec:fm_iot_foundations})
                    [Prompt-Based (\S \ref{sec:prompt_based_methods}),text width=10.9em
                        [
                            \citet{xu2024penetrative};
                            \citet{an2024iot};
                            \citet{ho2024remoni};
                            \citet{ouyang2024llmsense};
                            \citet{da2024prompt};
                            \citet{ji2024hargpt}; \\
                            \citet{cleland2024leveraging};
                            \citet{song2023pre};
                            \citet{xiao2024efficient}
                            \citet{chen2024towards}; 
                            \citet{arrotta2024contextgpt};
                            \citet{yang2024llm}; \\
                            \citet{guo2024sensor2scene};
                            \citet{nie2024llm};
                            \citet{sooriya2023poster};
                            \citet{yang2023edgefm};
                            \citet{sun2024ai}; \\
                            \citet{li2024llmcount};
                            \citet{karine2025using}.
                        ,leaf ,text width=55.5em]
                    ]
                    [Agent-Based (\S \ref{sec:agent_based_methods}),text width=10.9em
                        [
                            \citet{rivkin2024aiot}; 
                            \citet{yonekura2024generating};
                            \citet{liang2025llm};
                            \citet{liu2024agents4plc}; 
                            \citet{wang2024intelligent};\\
                            \citet{wang2024omnidrive};
                            \citet{cui2024drive};
                            \citet{huang2024drivlme};
                            \citet{abdallah2024netorchllm}; 
                            \citet{cui2024llmind};
                            \citet{liu2024tasking};\\
                            \citet{yang2025socialmind}
                            \citet{wen2024poster};
                            \citet{xu2024assuring};
                            \citet{yang2024plug};
                            \citet{li2024ids};
                            \citet{wang2024privacyoracle}; \\
                            \citet{zhong2024casit};
                            \citet{mao2023language};
                            \citet{yang2025contextagent};
                            \citet{shi2024ehragent}. 
                        ,leaf ,text width=55.5em]
                    ]
                    [Training-Based (\S \ref{sec:training_based_methods}),text width=10.9em
                        [
                         \citet{ouyang2025mmbind};
                         \citet{mo2024iot};
                         \citet{chen2024sensor2text};
                         \citet{xu2021limu};
                         \citet{kimura2024efficiency};
                         \citet{alikhani2024large}; \\
                         \citet{xiong2024novel};
                         \citet{yonekura2024generating};
                         \citet{xue2024leveraging};
                         \citet{ghassemi2024multi};
                         \citet{zhang2024mambareid}; \\
                         \citet{she2024llmdiff};
                         \citet{wan2024meit};
                         \citet{yan2024language};
                         \citet{weng2024large};
                         \citet{aboulfotouh2024building};
                         \citet{kimura2024vibrofm}; \\
                         \citet{kara2024phymask};
                         \citet{fan2024llmair};
                         \citet{liu2025llm4wm};
                         \citet{liu2024llm4cp};
                         \citet{imran2024llasa};
                         \citet{zhuang2024litemoe}; \\
                         \citet{rong2024leveraging};
                         \citet{yang2023privacy};
                         \citet{li2024sensorllm}.
                        ,leaf ,text width=55.5em]
                    ]
                ]
                [\;\;Criteria \\\;\;\;(\S \ref{sec:efficiency} - \S\ref{sec:security_privacy})
                    [Efficiency (\S \ref{sec:efficiency})                      
                        [Efficient Prompting,text width=13.4em
                            [
                            \citet{xu2024penetrative}; 
                            \citet{ho2024remoni};
                            \citet{an2024iot}; 
                            \citet{ho2024remoni};
                            \citet{chen2024towards}; \\
                            \citet{ouyang2024llmsense}; 
                            \citet{yang2025socialmind};
                            \citet{sun2024ai}; 
                            \citet{xiao2024efficient}; \\
                            \citet{li2024sensorllm}; 
                            \citet{mo2024iot}; 
                            \citet{chen2024sensor2text}; 
                            \citet{imran2024llasa};
                            \citet{yonekura2024generating}; \\
                            \citet{xue2024leveraging};
                            \citet{wan2024meit}. 
                            , leaf, text width= 42.5em]
                        ]
                        [Experience Archive,text width=13.4em
                            [ \citet{cui2024drive};
                              \citet{cui2024llmind};
                              \citet{abdallah2024netorchllm};
                              \citet{li2024ids};
                              \citet{zhong2024casit}; \\
                              \citet{shi2024ehragent};
                              \citet{mao2023language};
                              \citet{shi2024ehragent}.
                            ,leaf, text width = 42.5em]
                        ]
                        [Efficient Model Design,text width=13.4em
                            [
                            \citet{xu2021limu};
                            \citet{zhuang2024litemoe}. 
                            \citet{zhang2024mambareid}. 
                            \citet{kimura2024efficiency}; \\
                            \citet{li2024sensorllm}; \citet{liu2025llm4wm}; \citet{xiong2024novel};
                            \citet{imran2024llasa}.
                            ,leaf, text width = 42.5em]
                        ]
                    ]
                    [Context \\ Awareness (\S \ref{sec:context_awareness})
                        [Data Enrichment,text width=13.4em
                            [
                            \citet{an2024iot};
                            \citet{ji2024hargpt};
                            \citet{li2024sensorllm};
                            \citet{ouyang2024llmsense};
                            \citet{xu2024penetrative}.
                            ,leaf, text width=42.5em]
                        ]
                        [Fine-Tuning,text width=13.4em
                            [\citet{guo2024sensor2scene};
                             \citet{tian2025dailyllm}.
                            ,leaf, text width= 42.5em]
                        ]
                        [Contextual Inputs,text width=13.4em
                            [\citet{nepal2024mindscape};
                             \citet{cui2024llmind};
                             \citet{yang2025socialmind};
                             \citet{tian2025dailyllm};
                             \citet{huang2022inner}; \\
                             \citet{song2023llm};
                             \citet{singh2022progprompt}.
                            ,leaf, text width= 42.5em]
                        ]
                        [Contextual Memory,text width=13.4em
                            [\citet{yang2025socialmind};
                             \citet{cui2024llmind};
                             \citet{rivkin2024aiot};
                             \citet{saleh2025usercentrix}.
                            ,leaf, text width= 42.5em]
                        ]   
                        [Active Sensing,text width=13.4em
                            [\citet{cui2024llmind};
                             \citet{yang2025contextagent}.
                            ,leaf, text width= 42.5em]
                        ]   
                        [Iterative Self-Refinement,text width=13.4em
                            [
                            \citet{liu2024chainstream};
                            \citet{cui2024llmind};
                            \citet{guo2024sensor2scene};
                            \citet{xu2024assuring};
                            \citet{liu2024agents4plc}; \\
                            \citet{yang2024plug}.
                            ,leaf, text width= 42.5em]
                        ]   
                    ]
                    [Safety (\S \ref{sec:safety})
                        [Safety Requirement \\Incorporation,text width=13.4em
                            [\citet{xu2024assuring};
                             \citet{ravichandran2025safety};
                             \citet{brunke2025semantically}.  
                            ,leaf, text width= 42.5em]
                        ]
                        [User Request Verification,text width=13.4em
                            [\citet{cui2024drive};
                             \citet{xu2024assuring}.
                            ,leaf, text width= 42.5em]
                        ]
                        [Post-Response Verification,text width=13.4em
                            [\citet{xu2024assuring};
                             \citet{yang2024plug};
                             \citet{yang2024joint};
                             \citet{quartey2024verifiably}; \\
                             \citet{ravichandran2025safety}.
                            ,leaf, text width= 42.5em]
                        ]
                        [Interpretable \\Decision-Making,text width=13.4em
                            [\citet{cui2024drive};
                             \citet{zhang2024large};
                             \citet{nouri2024engineering}.
                            ,leaf, text width= 42.5em]
                        ]
                        [Iterative Self-Refinement,text width=13.4em
                            [\citet{xu2024assuring};
                             \citet{yang2024plug};
                             \citet{yang2024joint}.
                            ,leaf, text width= 42.5em]
                        ]
                    ] 
                    [Security and \\Privacy (\S \ref{sec:security_privacy})
                        [FMs for Anomaly \\Detection,text width=13.4em
                            [\citet{almodovar2022can};  
                             \citet{aghaei2022securebert};
                             \citet{diaf2024bartpredict};
                             \citet{li2024ids}; \\
                             \citet{wang2024privacyoracle}.
                            ,leaf, text width= 42.5em]
                        ]
                        [FMs for Data Obfuscation,text width=13.4em
                            [\citet{ouyang2024llmsense};
                            \citet{yang2023privacy};
                            \citet{wang2024privacyoracle}. 
                            ,leaf, text width= 42.5em]
                        ]
                        [Privacy Considerations,text width=13.4em
                            [\citet{yang2023privacy};
                             \citet{wang2024privacyoracle};
                             \citet{benazir2024maximizing}. 
                            ,leaf, text width= 42.5em]
                        ]
                    ]
                ]
                [Evaluation \\\;\;(\S \ref{sec:evaluation})
                    [Strategy, text width=4.5em
                        [Groundtruth,text width=9.9em
                            [\citet{kimura2024efficiency};
                             \citet{yang2025socialmind};
                             \citet{yang2023edgefm};
                             \citet{xu2021limu};
                             \citet{alikhani2024large}; \\
                             \citet{hesham2024localingua};
                             \citet{chen2024sensor2text};
                             \citet{liu2024chainstream};
                             \citet{xiong2024novel};
                             \citet{yan2024language}; \\
                             \citet{aboulfotouh2024building};
                             \citet{sun2024ai};
                             \citet{wan2024meit};
                             \citet{kimura2024vibrofm};
                             \citet{nie2024llm}; \\
                             \citet{zhang2024mambareid};
                             \citet{ho2024remoni};
                             \citet{kara2024phymask};
                             \citet{ghassemi2024multi};
                             \citet{she2024llmdiff}; \\
                             \citet{liu2025llm4wm};
                             \citet{liu2024llm4cp};
                             \citet{zhuang2024litemoe};
                             \citet{yang2023privacy};
                             \citet{wang2024privacyoracle}; \\
                             \citet{li2024sensorllm};
                             \citet{xue2024leveraging};
                             \citet{almodovar2022can};
                             \citet{diaf2024bartpredict};
                             \citet{xu2021limu}; \\
                             \citet{aghaei2022securebert};
                             \citet{li2024ids}. 
                            ,leaf ,text width=50.5em]
                        ]
                        [Human Evaluation,text width=9.9em
                            [\citet{guo2024sensor2scene};
                             \citet{yang2024drhouse};
                             \citet{xu2024large};
                             \citet{englhardt2024classification};
                             \citet{nie2024llm}; \\
                             \citet{imran2024llasa}. 
                            ,leaf ,text width=50.5em]
                        ]
                        [LLM-as-a-Judge,text width=9.9em
                            [\citet{guo2024sensor2scene};
                             \citet{yang2024drhouse};
                             \citet{imran2024llasa};
                             \citet{li2024sensorllm}.
                            ,leaf ,text width=50.5em]
                        ]
                        [Real-World \\Evaluation,text width=9.9em
                            [\citet{yang2025socialmind};
                             \citet{ouyang2024llmsense};
                             \citet{abdallah2024netorchllm};
                             \citet{song2023pre};
                             \citet{kara2024phymask}; \\
                             \citet{li2024llmcount};
                             \citet{imran2024llasa};
                             \citet{cui2024llmind}.
                            ,leaf ,text width=50.5em]
                        ]
                    ]
                    [Metrics, text width=4.5em
                        [
                        \emph{Efficiency}: 
                        Convergence speed \citep{kimura2024vibrofm, kimura2024efficiency};
                        Inference latency \citep{yang2023edgefm};
                        Memory usage \citep{worae2024unified}. \\
                        \emph{Context-Awareness}:
                        Executability rate \citep{liu2024chainstream};
                        Personalization score \citep{yang2025socialmind}. \\
                        \emph{Safety}:
                        Safety rate \citep{ma2025safety}. \\
                        \emph{Security and Privacy}: 
                        Classification based metrics (e.g. accuracy precision recall) \citep{li2024ids, houssel2024towards, diaf2024bartpredict, almodovar2022can, aghaei2022securebert, yang2023privacy, wang2024privacyoracle}.
                        ,leaf ,text width=62em]
                    ]
                ]               
        ]
        \end{forest}
    }
    \caption{\textbf{Taxonomy of Foundation Models for IoT.} The structure of this paper is as follows: Section \ref{sec:related_work} reviews and compares our survey with existing related work. Section \ref{sec:fm_iot_foundations} introduces three fundamental paradigms and frameworks for applying foundation models to IoT tasks, providing essential background for readers new to the field. Sections \ref{sec:efficiency} through \ref{sec:security_privacy} examine the four key performance criteria along with commonly used approaches to improve each. Section \ref{sec:evaluation} reviews the evaluation metrics used in the current literature for each criterion and discusses the strategies employed to assess FM-based solutions in IoT applications. Finally, Section \ref{sec:discussion} discusses existing research limitations and outlines future directions.}
    \label{fig:taxonomy}
\end{figure*}

%% file: sections/3related_work.tex
\section{Related Work}
\label{sec:related_work}

\input{sections/tables/survey_comparison}

Existing survey papers on foundation models for IoT have organized their reviews around the following three perspectives (Please also refer to Table \ref{tab:survey_comparison} for a comparison between prior survey studies and our work.): 1. \textbf{IoT Application Domains}: Nearly all prior surveys \citep{baris2025foundation, xu2024llm, kok2024iot, bhat2025llm, khatiwada2025large, sarhaddi2025llms} construct their taxonomies based on specific IoT application domains, including robotics, autonomous vehicles, industrial systems, experimental platforms, smart cities, smart homes, smart devices, healthcare, agriculture, communication, education, environmental monitoring, cybersecurity, and privacy. This broad coverage reflects the widespread integration of foundation models across diverse IoT sectors that significantly impact human life. As noted in the introduction, this adoption is largely driven by foundation models’ reduced reliance on labeled data and their ability to generalize across different contexts, which are the advantages over traditional machine learning approaches.
2.	\textbf{IoT System Architecture}: Several surveys \citep{kok2024iot, bhat2025llm} examine where foundation models can be deployed within the IoT infrastructure: specifically, at the edge, fog, or cloud layers. These taxonomies align with earlier studies \citep{zhang2020empowering, seng2022artificial, dou2023towards} on the deployment of machine learning models, recognizing foundation models as a subclass of machine learning systems with unique deployment considerations.
3.	\textbf{Sensor Modalities and Task Types}: \citet{baris2025foundation} analyze foundation models in cyber-physical systems (CPS) and IoT along two axes: sensor modalities (e.g., single, multi-modal, and flexible) and task types (e.g., fixed, configurable, selectable, and runtime-specifiable). This focus reflects the nature of IoT tasks, which often involve heterogeneous sensors and require models capable of handling multiple, dynamically defined tasks to ensure generalization and adaptability.

Additionally, two papers specifically address the evaluation of foundation models for IoT. \citet{xu2024llm} summarize commonly used evaluation metrics such as accuracy, RMSE, and success rate. \citet{sarhaddi2025llms} introduce the emerging “LLM-as-a-Judge” strategy, where large language models are used to assess the performance of other foundation models or LLMs in IoT tasks, which is an increasingly popular approach following the widespread adoption of LLMs.

Compared to prior surveys, our paper introduces a novel taxonomy based on performance criteria. As discussed in the introduction, identifying shared performance criteria across diverse IoT scenarios (e.g.,  application domains, deployment locations, sensor types, and task types) enables systematic cross-domain comparison of methods targeting specific performance goals. This approach also supports the informed selection of appropriate techniques for new scenarios and fosters the development of more generalizable methods. These insights and the associated research opportunities have been largely overlooked in previous surveys.

Furthermore, to facilitate comparison and selection of methods for each performance criterion, our survey summarizes a broader set of evaluation metrics for each criterion beyond general metrics outlined in \citet{xu2024llm}. We also provide a comprehensive discussion of four evaluation strategies: ground-truth comparison, human evaluation, and LLM-as-a-Judge, including their advantages, limitations, and applicable performance criteria. This analysis offers significantly greater depth and coverage than that presented in \citet{sarhaddi2025llms}.

We also summarize three fundamental paradigms of foundation models for IoT tasks (i.e., prompt-based, agent-based, and training-based methods), which have been overlooked in previous surveys. Understanding these paradigms is crucial because they offer distinct trade-offs in terms of computational efficiency, adaptability to diverse IoT environments, ease of integration with external systems, and requirements for training data and infrastructure. In addition, by clarifying the strengths and limitations of each approach as well as presenting a decision tree to help select the most suitable paradigm, our survey equips researchers and practitioners with the necessary framework to select, adapt, or develop solutions that are best suited to their specific IoT applications and resource constraints.

%% file: sections/tables/survey_comparison.tex
\begin{table*}[t!]
\caption{\textbf{Comparison with Prior Survey Papers.} Unlike previous survey papers that primarily focus on specific IoT application domains (e.g., healthcare, smart homes), this survey introduces a taxonomy centered on performance criteria. We also provide a summary of key foundation model paradigms, a multidimensional comparison of these paradigms, and a detailed overview of evaluation metrics for each performance criterion as well as commonly used evaluation strategies. The content in parentheses reflects the alternative perspectives adopted in prior surveys.}
\label{tab:survey_comparison}
\resizebox{0.8\textwidth}{!}{%
\begin{tabular}{@{}llccc@{}}
\toprule
\multirow{2}{*}{\textbf{\begin{tabular}[c]{@{}c@{}}Survey \\ Papers\end{tabular}}} &
  \multicolumn{1}{c}{\multirow{2}{*}{\textbf{\begin{tabular}[c]{@{}c@{}}Performance Criteria \\ Perspective\end{tabular}}}} &
  \multirow{2}{*}{\textbf{\begin{tabular}[c]{@{}c@{}}FM Paradigms \\ \& Comparison\end{tabular}}} &
  \multicolumn{2}{c}{\textbf{Evaluation}} \\ \cmidrule(l){4-5} 
                   & \multicolumn{1}{c}{}                     &     & \textbf{Metrics} & \textbf{Strategies} \\ \midrule
{\citet{baris2025foundation}}            & \ding{56}  (Sensor modalities, task types)      & \ding{56}  & \ding{56}               & \ding{56}                  \\
{\citet{xu2024llm}}            & \ding{56} (IoT application domains)             & \ding{56}  & \ding{52}              & \ding{56}                  \\
{\citet{kok2024iot}}            & \ding{56} (IoT application domains, IoT system architecture)             & \ding{56}  & \ding{56}               & \ding{56}                  \\
{\citet{bhat2025llm}}            & \ding{56} (IoT application domains, IoT system architecture)             & \ding{56}  & \ding{56}               & \ding{56}                  \\
{\citet{khatiwada2025large}}            & \ding{56} (IoT application domains)             & \ding{56}  & \ding{56}               & \ding{56}                  \\
{\citet{sarhaddi2025llms}}            & \ding{56} (IoT application domains) & \ding{56}  & \ding{56}               & \ding{52}                 \\ \midrule
\textbf{Our Paper} & \multicolumn{1}{c}{\ding{52}}                  & \ding{52} & \ding{52}              & \ding{52}                 \\ \bottomrule
\end{tabular}%
}
\end{table*}

%% file: sections/4taxonomy.tex
\input{sections/taxonomy/1foundation_fm_iot}

\input{sections/taxonomy/6efficiency}

\input{sections/taxonomy/7context_awareness}

\input{sections/taxonomy/8safety}

\input{sections/taxonomy/9security_and_privacy}

\input{sections/taxonomy/10evaluation}

%% file: sections/taxonomy/1foundation_fm_iot.tex
\section{Foundations of FMs for IoT }
\label{sec:fm_iot_foundations}

In this section, we categorize existing approaches that leverage foundation models into three core paradigms: \emph{prompt-based methods}, \emph{agent-based methods}, and \emph{training-based methods}. These approaches are used to: (1) perform end-to-end processing by perceiving, reasoning over, and making decisions based on sensor data, currently the primary focus of most reviewed studies, and (2) support reinforcement learning (RL) to improve policy learning for IoT tasks (e.g., HVAC control \citep{yang2024llm}, mobile health \citep{karine2025using}, beam management \citep{ghassemi2024multi}), an area that remains relatively underexplored.
We then compare these paradigms across key dimensions, highlight their respective strengths and limitations, and provide practical guidance for selecting the most appropriate approach for practitioners and researchers. 

\input{sections/taxonomy/2prompt_based}
\input{sections/taxonomy/3agent_based}
\input{sections/taxonomy/4training_based}

\input{sections/taxonomy/5comparison}

%% file: sections/taxonomy/2prompt_based.tex
\vspace{-0.1in}
\subsection{Paradigm I: Prompt-Based Methods}
\label{sec:prompt_based_methods}

Prompt-based methods leverage commercial or open-source \emph{pretrained large language models (LLMs)} (e.g., GPT-4 \citep{achiam2023gpt}, LLaMA-2 \citep{touvron2023llama}, DeepSeek-V2 \citep{liu2024deepseek}) to perform IoT tasks through textual inputs without requiring additional training \citep{sahoo2024systematic, liu2023pre}. These prompts typically serve as task-specific instructions (e.g., “You are a knowledgeable IoT expert. Please classify the human activity described by the following IMU data.”), which the model interprets to generate a response in the specified format \citep{sahoo2024systematic, liu2023pre} (please refer to the example prompt for using LLMs to address the IoT task in Appendix \ref{sec:appendix_example_prompt}). Because prompt-based approaches rely solely on pretrained models, they enable rapid prototyping and deployment, particularly in settings with limited computational resources.

Prompt-based methods offer several advantages for applying foundation models to IoT tasks:
(1)	\textbf{Efficiency}: Unlike training-based methods that require substantial labeled data and computational power to update model parameters (see Section \ref{sec:training_based_methods}), prompt-based approaches leverage already trained LLMs which exhibit strong few-shot learning capabilities \citep{song2023comprehensive, brown2020language}, eliminating the need for fine-tuning or retraining. This significantly reduces deployment costs and time.
(2) \textbf{General Knowledge}: LLMs trained on large-scale web data possess extensive general and domain-relevant knowledge, which can be leveraged to interpret IoT data and support tasks (e.g., human activity recognition from IMU signals) \citep{xu2024penetrative, an2024iot, ho2024remoni, ouyang2024llmsense, da2024prompt, ji2024hargpt, cleland2024leveraging, arrotta2024contextgpt, chen2024towards, song2023pre}.
(3) \textbf{Enhanced Reasoning}: Prompting techniques like chain-of-thought (CoT) \citep{wei2022chain} and program-of-thought (PoT) \citep{chen2022program} can activate commonsense reasoning abilities in LLMs and improve their ability to interpret complex or ambiguous IoT data. This has been shown to significantly boost performance in tasks such as human activity recognition, industrial monitoring, and sensor data analysis \citep{kok2024iot, xiao2024efficient}.
(4) \textbf{Flexibility with Heterogeneous Data}: Prompt-based methods can accommodate both structured (e.g., tabular sensor readings) and unstructured data (e.g., geospatial graphs) by explicitly specifying input and output formats in the prompt \citep{zong2025integrating, fan2024llmair}. This flexibility makes them well-suited to the diverse data types found in IoT systems.
(5) \textbf{Natural Language Interfaces}: LLMs can serve as intuitive interfaces for querying or controlling IoT systems using natural language. This lowers the barrier to entry for non-experts and enhances system usability and accessibility.
(6) \textbf{Interpretability}: Language-based outputs, especially when coupled with reasoning prompts (e.g., “Explain your answer in a few sentences.”), enhance transparency by revealing the model’s reasoning process. This allows users to verify the outputs and increases trust in the system, unlike many traditional ML models that operate as black boxes \citep{cui2024drive}.
(7) \textbf{Real-Time Insights}: Prompted LLMs are capable of processing large streams of sensor data and producing actionable responses in real time, making them suitable for rapid prototyping and deployment in dynamic IoT environments \citep{zong2025integrating}.

%% file: sections/taxonomy/3agent_based.tex
\subsection{Paradigm II: Agent-Based Methods}
\label{sec:agent_based_methods}

While prompt-based methods leveraging pretrained LLMs offer a lightweight and accessible way to address IoT tasks, they face several inherent limitations stemming from the constraints of the underlying pretrained models:

(1)	\textbf{Hallucination and Knowledge Limitations}: Despite being trained on large-scale internet data, LLMs still suffer from hallucination (i.e., generating responses that are factually incorrect or nonsensical \citep{huang2025survey}). This issue arises from outdated or inaccurate information in the pretraining corpus and is particularly problematic in the IoT domain, where LLMs are rarely trained on sensor data or task-specific information. As a result, they may struggle to interpret some real-world signals and make accurate decisions in high-stakes applications (e.g., healthcare or autonomous driving) \citep{kim2025medical, agarwal2024medhalu, ahmad2023creating, wang2024hallucination, dona2024llms}. While embedding domain-specific knowledge into prompts can help, it is often incomplete and difficult to scale or adapt to dynamic user needs. Without access to accurate, up-to-date external knowledge, prompt-based methods remain fundamentally limited.

(2) \textbf{Lack of Specialized Capabilities and Active Perception}: Although LLMs demonstrate strong zero- and few-shot generalization, they often underperform relative to specialized modules on complex, domain-specific tasks (e.g., object detection, wireless resource allocation, numerical reasoning) \citep{abdallah2024netorchllm, cui2024drive}. Furthermore, LLMs lack the ability to actively gather additional information when input data is ambiguous or incomplete, limiting their ability to resolve uncertainty and optimize task performance. This passivity contrasts significantly with the dynamic nature of many IoT environments.

(3)	\textbf{Limited Reasoning and Planning for Complex Tasks}: Many IoT applications require multi-step reasoning and decision-making beyond basic question-answering. For instance, predictive maintenance in smart factories may involve interpreting sensor data, diagnosing failures, ordering components, and scheduling repairs. Solving such tasks requires the ability to decompose problems into subtasks, reason about their dependencies, and interact with external tools or systems \citep{abdallah2024netorchllm, cui2024drive}. While prompt-based methods with manually embedded task decompositions offer partial solutions, they lack the adaptability and autonomy needed for real-time reasoning and planning across diverse scenarios, ultimately limiting the system’s generalizability and effectiveness.

To overcome the limitations of prompt-based methods, LLM agents \citep{wang2024survey, luo2025large, weng2023agent}, which augment pretrained language models with additional modules: \emph{external memory}, \emph{tool integration}, and \emph{planning capabilities}, are gaining increasing adoption across various IoT domains (e.g., smart homes \citep{rivkin2024aiot, yonekura2024generating}, industrial monitoring \citep{liang2025llm, liu2024agents4plc, wang2024intelligent}, autonomous driving \citep{wang2024omnidrive, cui2024drive, huang2024drivlme}). These agents retain the strengths of prompt-based approaches while addressing their key shortcomings:

(1) \textbf{Access to External Knowledge}: By integrating external memory or online search tools, LLM agents can retrieve up-to-date, task-relevant information beyond what is stored in their static model parameters \citep{lewis2020retrieval, zhang2024survey}. This significantly mitigates hallucination and improves decision-making accuracy, especially in fast-evolving or safety-critical IoT applications.

(2) \textbf{Advanced Reasoning and Task Execution}: LLM agents can autonomously decompose complex tasks into subtasks \citep{huang2024understanding, wei2025plangenllms, aghzal2025survey, li2024lasp}, perform multi-step reasoning \citep{plaat2024reasoning, zhang2024llm, ferrag2025llm, yao2023tree, wei2022chain}, and invoke specialized tools or hardware systems as needed \citep{shen2024llm, qu2025tool}. This enables them to handle sophisticated IoT workflows (e.g., predictive maintenance or adaptive control) that go beyond the capabilities of purely prompt-driven models.

(3) \textbf{Adaptive Learning and Feedback Integration}: Through the use of external memory and feedback loops, LLM agents can store and recall prior experiences to improve performance over time. They can also adjust their outputs in response to real-time feedback from humans or the environment, allowing them to learn and adapt without retraining the base model \citep{abdallah2024netorchllm, cui2024llmind, cui2024drive}.

By integrating the generalization capabilities of LLMs with domain-specific knowledge, real-time interactivity, and advanced reasoning, LLM agents offer a more robust and scalable solution for building intelligent, adaptable, and context-aware IoT systems, particularly for addressing complex tasks, that exceed the capabilities of prompt-based methods.

%% file: sections/taxonomy/4training_based.tex
\subsection{Paradigm III: Training-Based Methods}
\label{sec:training_based_methods}
Although prompt-based and agent-based methods demonstrate promise in addressing IoT tasks, their effectiveness is limited by the constraints of pretrained LLMs, which typically lack task- and environment-specific knowledge \citep{li2024sensorllm, mo2024iot}. In contrast, training-based methods can update model parameters using domain-specific data, enabling higher accuracy and better adaptation to the target IoT application.

Training-based methods of foundation models for IoT tasks include two stages: \emph{pretraining} and \emph{fine-tuning}.

Pretraining involves training a large model (often not limited to language models) on extensive volumes of unlabeled data collected from diverse devices and environments. In the IoT context, this includes raw sensor data (e.g., vibration, acoustic, temperature signals) collected from many devices and environments \citep{ouyang2025mmbind, mo2024iot, chen2024sensor2text, xu2021limu, yan2024language, weng2024large, kimura2024efficiency, yonekura2024generating, kimura2024vibrofm, kara2024phymask, li2024sensorllm}. The objective is to learn general, environment-invariant representations that are robust to variations in deployment conditions and noise levels. In practice, pretraining may not need to be performed from scratch if task-relevant pretrained models are already available (e.g., LIMU-BERT \citep{xu2021limu} for IMU data or VibroFM \citep{kimura2024vibrofm} for vibration signals).

Fine-tuning involves adapting the pretrained model to a specific application by updating all or part of its parameters using a smaller, labeled dataset relevant to the target task (e.g., fall detection within a specific building). This process tailors the general representations learned during pretraining to the unique characteristics of the application domain, sensor setup, or environmental context \citep{xiong2024novel, imran2024llasa, liu2025llm4wm, hesham2024localingua, kimura2024efficiency}.

Compared to prompt-based and agent-based methods that rely on pretrained LLMs, training-based methods offer significant performance improvements on IoT tasks by updating model parameters and directly learning from raw sensor data. Also, unlike LLM-based approaches, they do not depend on converting numerical sensor data into natural language tokens. This tokenization process can introduce information loss or misinterpretation due to internal LLM limitations (e.g., inconsistent tokenization) \citep{an2024iot, zhang2024large} or prompt engineering constraints (e.g., downsampling or quantization of long sensor inputs) \citep{xu2024penetrative, mo2024iot, ho2024remoni}.
Additionally, training-based methods can effectively leverage both labeled and unlabeled data, continuously improving through exposure to new domain-specific inputs by updating their internal representations.

%% file: sections/taxonomy/5comparison.tex
\subsection{Method Comparison}
\label{sec:method_comparison}
In this section, we compare the three most commonly used frameworks for applying foundation models to IoT tasks: \emph{prompt-based methods}, \emph{agent-based methods}, and \emph{training-based methods}, as well as \emph{traditional supervised learning methods}. The comparison is made across six dimensions: \emph{computation requirement} (CR), \emph{error rate on specific tasks} (ER), \emph{task specificity} (TS), \emph{development time} (DT), \emph{labeled data requirement} (LDR), and \emph{unlabeled data requirement} (UDR). For training-based methods, we focus on models after \emph{fine-tuning}. Task specificity refers to the breadth of tasks a method can address; a lower specificity indicates broader applicability. Based on this comparison, we highlight the strengths and weaknesses of each method and offer practical guidance to help researchers and practitioners choose the most appropriate approach for their specific IoT applications.

\begin{table*}[t!]
\label{tab:method_comparison}
\caption{\textbf{Comparison of Different Approaches.} The first three methods apply foundation models, while the last represents traditional machine learning models with supervised learning. 
\textbf{CR}: computation requirement; 
\textbf{ER}: error rate; 
\textbf{TS}: task specificity; 
\textbf{DT}: development time;
\textbf{LDR}: labeled data requirement, 
\textbf{UDR}: unlabeled data requirement.
For training-based methods, we focus on models after \textbf{fine-tuning}.}
\label{tab:method_comparison}
\resizebox{0.8\textwidth}{!}{
\begin{tabular}{@{}lcccccc@{}}
\toprule
\textbf{Methods}                     & \textbf{CR} & \textbf{ER} & \textbf{TS} & \textbf{DT} & \textbf{LDR} & \textbf{UDR} \\ \midrule
Prompt-based   & Low      & High     & Moderate & Low      & Low      & Low  \\
Agent-based    & Moderate & Moderate & Low     & Moderate & Low      & Low  \\
Training-based & High     & Low      & High      & High     & Moderate & High \\
Traditional ML (supervised) & Moderate    & High         & Low         & Moderate    & High         & Low          \\ \bottomrule
\end{tabular}
}
\end{table*}

Table \ref{tab:method_comparison} summarizes the comparison results, highlighting the strengths and limitations of each method.
\emph{Prompt-based methods} offer rapid development, low computational and data requirements, and easy adaptability to new or evolving IoT tasks. However, they typically yield higher error rates on complex or domain-specific tasks, which are limited by the fixed knowledge of the underlying LLM, and cannot improve over time through additional data exposure.
\emph{Agent-based methods} extend prompt-based approaches by enabling more complex, multi-step, and multi-device workflows, often achieving higher accuracy for orchestrated tasks without requiring the computational demands of full model training. Nonetheless, like prompt-based methods, they remain constrained by the static knowledge of the base LLM and cannot learn from new data.
\emph{Training-based methods} achieve the high accuracy and robustness for domain-specific IoT tasks, particularly when handling heterogeneous or noisy data. They can evolve through exposure to new data via pretraining and fine-tuning. However, they are the most resource- and time-intensive, requiring large volumes of unlabeled and labeled data, substantial computational resources, and longer training times. Compared to traditional ML methods, they often require less labeled data due to their ability to leverage generalizable knowledge from pretraining.
\emph{Traditional ML methods} are efficient for narrow, well-defined tasks with sufficient labeled data and offer low inference costs. However, they lack the flexibility and adaptability of FM-based approaches.

In conclusion, we offer the following practical guidance for selecting methods based on task requirements and resource constraints: use \emph{prompt-based methods} for rapid prototyping or when computational resources and labeled data are limited; adopt \emph{agent-based methods} for complex, multi-step IoT workflows requiring tool orchestration; apply \emph{training-based methods} when high accuracy and robustness are critical, especially in domain-specific or mission-critical settings with abundant unlabeled data; and rely on \emph{traditional ML approaches} for simple, well-defined tasks with sufficient labeled data and low computational demands. We also provide a decision tree for method selection based on the evaluated dimensions in Appendix \ref{sec:appendix_method_selection}.

\textbf{Note}: The comparison in Table \ref{tab:method_comparison} does not account for security and privacy constraints. We assume that all pretrained LLMs are deployed in the cloud, computational resources are solely considered from the user’s perspective, and inference costs on the cloud side are ignored. Additionally, we assume low time requirements are supported by high-speed network access. If these assumptions do not hold, the comparison results may differ. For instance, in privacy-sensitive applications (e.g., involving personal data), data may not be allowed to leave local devices \citep{xiao2024efficient}. In such cases, LLMs must be deployed locally, increasing the computational burden on edge devices and raising the resource requirements for prompt-based methods to a moderate level.

%% file: sections/taxonomy/6efficiency.tex
\vspace{-0.06in}
\section{Criterion I: Efficiency}
\label{sec:efficiency}

In the context of IoT tasks, efficiency in foundation models encompasses several key objectives: (1) \emph{reducing inference time}, 
(2) \emph{decreasing computational resources (e.g., memory) and energy consumption}, and 
(3) \emph{lowering network load}.  

Improving the efficiency of foundation models for IoT applications is essential due to their unique constraints of these environments. First, many IoT tasks (e.g., fall detection and autonomous driving) demand fast inference to ensure real-time responsiveness and prevent harmful delays \citep{karar2022survey, mozaffari2019practical, yacchirema2019fall, dai2019hybridnet, choi2019gaussian, neven2017fast}. 
Second, most IoT devices have limited memory and battery capacity, requiring models to be both lightweight and energy-efficient \citep{pereira2020challenges, georgiou2017iot}. This constraint becomes even more critical when data must be processed locally on edge devices due to privacy and security concerns (e.g., in hospital settings where uploading sensitive patient data to the cloud is not viable) \citep{xiao2024efficient}. 
Finally, even when cloud processing is allowed, limited bandwidth in edge devices can introduce significant upload delays, further increasing overall inference latency \citep{said2023bandwidth, al2017bandwidth}. These challenges underscore the importance of developing efficient foundation models tailored to the IoT context.

To improve the efficiency of foundation models for IoT tasks, four common strategies are identified in the surveyed literature: \emph{efficient prompting}, \emph{experience archive}, 
and \emph{efficient model architecture design}. Table \ref{tab:efficiency_summary} provides an overview of these methods, including their descriptions and the paradigms (prompt-based, agent-based, training-based) to which they are mostly commonly applied.  \smallskip

\input{sections/tables/efficiency_methods_summary}

\noindent\textbf{Efficient Prompting. }
Efficient prompting refers to minimizing prompt length without significantly compromising the critical information it conveys when using LLM-based methods, which rely on prompts as input. This is particularly important in IoT applications. When the LLM is hosted in the cloud, longer prompts increase token count, leading to higher network load and longer transmission times, which is an issue for IoT devices usually with limited bandwidth, ultimately slowing inference. When the LLM is deployed locally, processing longer prompts demand more memory due to the Transformer-based architecture of current models. However, edge devices typically have constrained memory and cannot efficiently process lengthy inputs. Therefore, efficient prompting is essential to ensure low-latency, resource-aware performance in IoT environments.

When LLMs are used for sensor data analysis and reasoning, which is the primary focus of most of the reviewed papers, the sensor data itself forms part of the prompt. However, as previously discussed, IoT devices can only support prompts with a limited number of tokens, which poses a major scalability challenge. For example, long-term monitoring tasks (e.g., tracking an individual’s activity over two weeks at one-minute intervals) can generate numeric values far exceeding the input capacity of most LLM-based methods. Similarly, large-scale spatiotemporal data (e.g., those from air quality monitoring networks) also surpass the manageable prompt length \citep{ouyang2024llmsense}. These limitations restrict the direct application of LLMs to many real-world IoT scenarios that involve continuous or high-dimensional sensor streams.

To address the challenge of scaling LLMs to large-scale sensor data, researchers have proposed several strategies to reduce the volume of sensor data included in the prompt while preserving inference efficiency and performance. In the following sections, we review five key approaches that address this issue. In the following sections, we review four key approaches commonly used in prompt-based methods (Section \ref{sec:prompt_based_methods}) and one approach frequently applied in training-based methods (Section \ref{sec:training_based_methods}). These efficient prompting techniques are generally not adopted in agent-based methods (Section \ref{sec:agent_based_methods}), as sensor data are typically preprocessed by external tools or lightweight models before being passed to the LLM \citep{cui2024llmind, zhong2024casit, li2024ids, cui2024drive, mao2023language, yang2025contextagent}. For strategies focused on optimizing other components of the prompt (e.g., such as instructions or chain-of-thought (CoT) demonstrations \citep{wei2022chain}), please refer to surveys such as \citet{chang2024efficient}. 

\vspace{0.1in}
\textit{Downsampling and Quantization. }
Downsampling (i.e., resampling at a lower rate) and quantization (i.e., reducing data precision, such as rounding to integers or two decimal places) are commonly used to shorten the numerical sensor data \citep{xu2024penetrative, ho2024remoni}. Quantization is effective because floating-point numbers with many decimal places are often split into multiple tokens by tokenization algorithms (e.g., Byte Pair Encoding \citep{gage1994new, sennrich2015neural}). Lowering precision reduces the number of tokens generated by such decimal expansions \citep{an2024iot}. However, quantization should be applied carefully, as discarding high-precision values may eliminate information crucial to the target task. Similarly, downsampling must preserve critical features to retain data utility \citep{ho2024remoni}. For example, R peaks in ECG data must be retained if the goal is to detect cardiovascular diseases based on this feature. \smallskip

\textit{Sliding Window. } For long sensor data sequences, a common strategy is to divide the data into smaller windows for sequential processing. However, this segmentation can lead to the loss of important contextual information found before and after each window. Preserving context is crucial for accurate interpretation of sensor data. To mitigate this issue, overlapping sliding windows are often used, allowing each segment to retain some information from adjacent windows \citep{ouyang2024llmsense, chen2024towards}.  \smallskip

\begin{figure}[h!]
    \centering
    \includegraphics[width=\columnwidth]{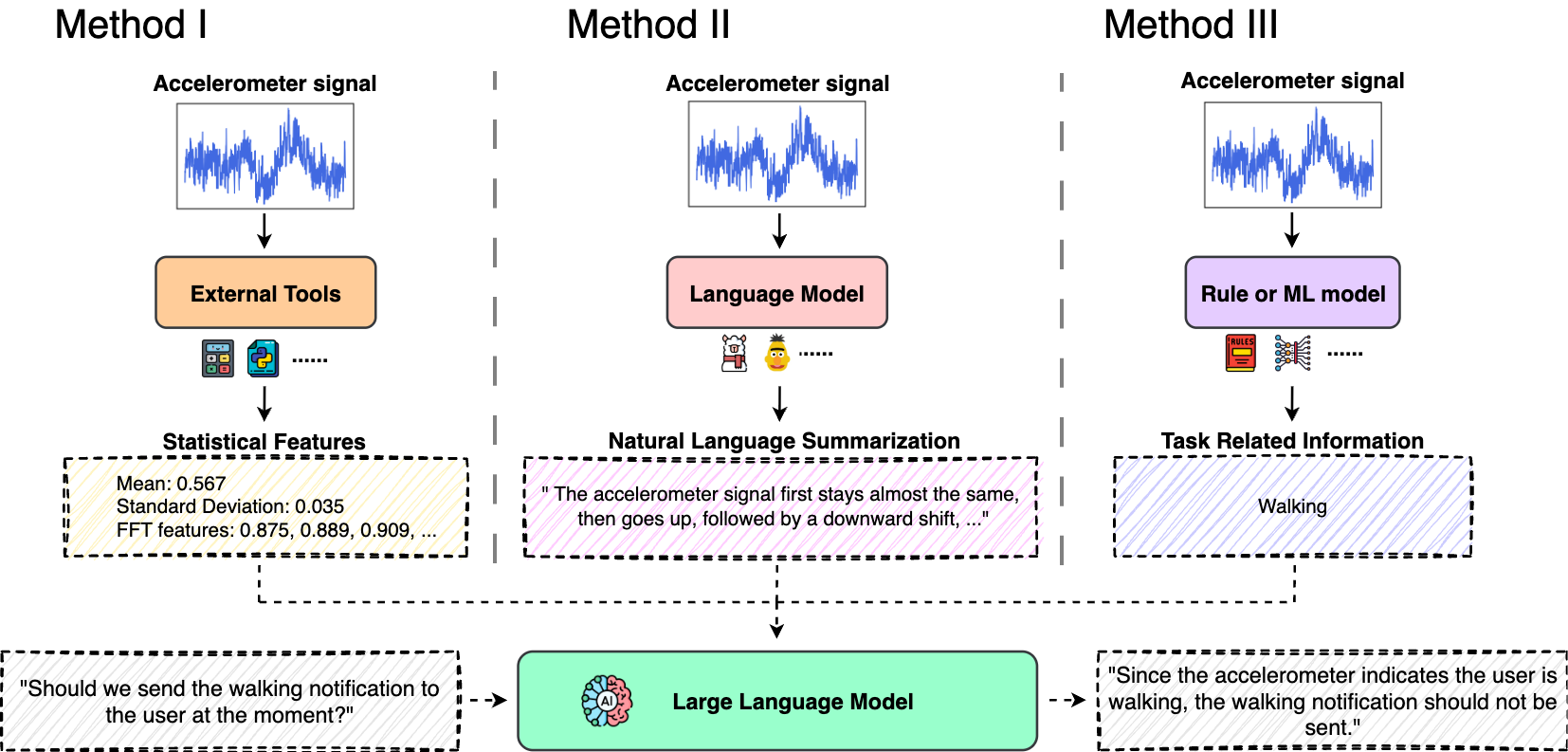}
    \caption{Illustration of data summarization methods for enhancing the efficiency of foundation models in IoT tasks, exemplified by an LLM-based Just-in-Time Adaptive Intervention (JITAI) system in mobile health \citep{nahum2016just}.}
    \label{fig:data_summarization}
\end{figure}
    
\textit{Data Summarization. } To reduce input volume, raw sensor data can be summarized before being passed to the model. Three common approaches are used for summarizing numerical sensor data. The first relies on external tools (e.g., calculators or Python scripts) to extract \emph{high-level statistical features} (e.g., mean, variance, or FFT-based metrics) \citep{an2024iot}. This method is particularly effective when these features are directly relevant to downstream tasks (e.g., an elevated average body temperature may indicate fever, suggesting a potential health concern). The second approach leverages LLMs or lightweight language models (e.g., DistillBert \citep{sanh2019distilbert}) to generate \emph{natural language summaries} of raw sensor data \citep{ouyang2024llmsense}. These models have shown strong capabilities in identifying and articulating key numerical patterns (e.g., trends and state changes) and these summaries are typically much shorter than the original data while preserving essential information. The third approach uses task specific modules (e.g., rule-based methods or lightweight ML models) to extract \emph{task related information} (e.g., identifying individuals from vibration data or recognizing activities from audio signals) \citep{yang2025socialmind, ouyang2024llmsense, sun2024ai}. Only the extracted outputs are passed to the LLM when these simpler models are insufficient for completing the task \citep{ouyang2024llmsense}. This approach reduces the LLM’s computational burden, allowing it to focus on high-level reasoning and user interaction. Figure \ref{fig:data_summarization} further illustrates aforementioned methods for enhanced clarity. \smallskip

\textit{Critical Information Incorporation. } The information contained in the sensor data is often sparse, making it possible to significantly reduce input length by including only components containing task-relevant information. For example, in human activity recognition, sensor readings frequently remain unchanged for long durations (e.g., when a person is not present in a room). To eliminate redundancy, prompts can be constructed using only state-change events (i.e., time steps where sensor values change) \citep{chen2024towards}. In multi-modal settings involving numerous sensor types, an initial summary capturing sensor types, data formats, and the first and last readings of each type can be used to prompt the LLM. The model can then perform commonsense reasoning to identify which sensors are most relevant to the downstream task, allowing only the data from those critical sensors to be included in the final prompt \citep{xiao2024efficient}.  \smallskip

\textit{Sensor Data Encoding. } When fine-tuning is permitted, large-scale sensor data can be efficiently compressed using a lightweight neural encoder (e.g., an RNN for temporal data or a GNN for spatial data). The encoder transforms the raw sensor input into one or more compact embeddings, which are treated as special tokens and concatenated with textual tokens in the prompt. These combined representations are then input into the trainable LLM, with both the encoder and the LLM jointly optimized during training \citep{li2024sensorllm, mo2024iot, chen2024sensor2text, imran2024llasa, yonekura2024generating, xue2024leveraging, wan2024meit}. This approach addresses limitations associated with treating numerical sensor data as textual input, which can lead to information loss or misinterpretation due to internal LLM constraints (e.g., inconsistent tokenization) or prompt engineering challenges (e.g., downsampling or quantization of lengthy sensor sequences). \smallskip

\noindent\textbf{Experience Archive.}
The \emph{Experience Archive} serves as a form of long-term external memory, in contrast to short-term memory used for in-context learning and internal memory encoded in model parameters \citep{abdallah2024netorchllm, cui2024llmind, li2024ids, zhong2024casit}. It stores task-specific experiences as a sequence of tuples $\mathcal{H}_{t_m:t_n} = \{(c_i, q_i, r_i) | i=t_m:t_n\}$, where $c_i$ denotes the context, $q_i$ represents the user query or instruction, and $r_i$ includes the LLM agent’s response along with the execution result produced by an external tool or executor. For example, $r_i$ may include a planned trajectory generated for an autonomous driving task based on prior environmental states and user input \citep{cui2024drive, mao2023language}. This type of memory is also referred to as \emph{episodic memory} in prior work \citep{xiong2025memory, nuxoll2007extending}.

In Section \ref{sec:context_awareness}, we introduce another form of long-term external memory, referred to as the \emph{context repository} or \emph{semantic memory}, which stores knowledge and information about the agent’s current context (e.g., environmental or personal state). The key distinction from the experience archive lies in temporal scope and structure: the experience archive captures past contextual information, interaction history, and execution results, whereas the context repository contains current factual knowledge used for reasoning and decision-making. These facts may be extracted from sensors (e.g., objects detected in a room), pre-stored by developers (e.g., maps or traffic regulations), or inferred from past experiences (e.g., personality traits derived from previous conversations). However, unlike past experiences stored in the experience archive, they do not include the original background data, the LLM agent’s responses, or the corresponding execution results from which they were derived \citep{nuxoll2007extending}.

The experience archive can significantly improve the efficiency of LLM-based agents in IoT applications. When the current task closely matches a previously encountered one, based on contextual information and user instructions, the agent can retrieve and reuse the prior response as output, avoiding the need for regeneration, which is typically more time-consuming than retrieval \citep{cui2024drive, abdallah2024netorchllm, li2024ids, zhong2024casit}. In addition to successful outcomes, storing failed experiences also contributes to efficiency: it enables self-reflection and self-improvement by helping the agent recognize and avoid repeating past mistakes, thereby conserving computational resources \citep{cui2024llmind, abdallah2024netorchllm}. Moreover, even when no exact match is found, partially similar past experiences can serve as in-context demonstrations, reducing the need for manual prompt engineering and further streamlining the decision-making process \citep{xiong2025memory}.

The experience archive can include both experiences generated during interactions between the LLM agent and the user, as well as preloaded experiences (e.g., from training or validation datasets \citep{li2024ids, xiong2025memory}). These experiences may be represented in textual form (e.g., natural language \citep{mao2023language}, programming code \citep{shi2024ehragent}) or as vector embeddings, similar to those used in retrieval-augmented generation (RAG) systems \citep{lewis2020retrieval}.

To retrieve relevant experiences, similarity metrics are applied (e.g., cosine similarity for embeddings or LLM-defined semantic similarity for text-based comparisons). This process involves interpreting the semantic content of user instructions. However, user inputs may be under-specified or embedded in diverse linguistic patterns, and the LLM may require multiple interaction rounds to fully understand the task. To address this, systems like LLMind \citep{cui2024llmind} prompt the LLM to generate concise summaries of user instructions for more accurate retrieval.

Effective use of the experience archive also depends on proper memory management \citep{xu2025mem, zhang2024survey, xiong2025memory}. First, external memory is inherently limited, making it infeasible to store all past experiences. Second, not all historical experiences are equally valuable; outdated or erroneous memories may mislead the agent. To mitigate these issues, several memory management strategies have been proposed, including selective addition and deletion \citep{xiong2025memory, zhong2024casit}, structural transformation \citep{wang2023voyager, anokhin2024arigraph, zeng2024structural}, merging \citep{zhong2024memorybank, liu2023think, hu2024hiagent}, summarization \citep{wang2024agent, zhong2024memorybank}, and reflection. Readers are referred to these works for further details. \smallskip

\noindent\textbf{Efficient Model Architecture Design. }
Efficient model architectures are essential for applying foundation models to addressing IoT tasks, since foundation models often contain a vast number of parameters to encode general knowledge across large-scale diverse datasets. This complexity results in high computational costs and prolonged inference time, making such models impractical for deployment on typical IoT devices or in scenarios requiring real-time responsiveness. In the following sections, we review several techniques aimed at improving the architectural efficiency of foundation models. \smallskip


\textit{Efficient Transformer Architecture. }
The Transformer architecture \citep{vaswani2017attention} has been widely adopted in the IoT domain to build foundation models (e.g., IoT-LM \citep{mo2024iot} and LIMU-BERT \citep{xu2021limu}). However, Transformers are known for their high computational complexity, driven by three key factors: (1) \emph{the self-attention mechanism}, which scales quadratically with input length; (2) \emph{a deep multilayer architecture} with unshared trainable parameters across different layers; and (3) \emph{fully-connected feedforward networks (FFNs)}, which contain the majority of the model’s parameters ($\geq 66\%$, typically) \citep{vaswani2017attention}. 

To address the computational challenges of self-attention, sparse attention mechanisms \cite{tay2022efficient, child2019generating, ho2019axial, chen2023learning, wei2024temporally} can be employed. Rather than allowing each token to attend to all others, sparse attention restricts attention to a limited set of positions, significantly reducing computational complexity. However, sparse attention has not been extensively validated at scale in large foundation models, leaving an open question for future research. Furthermore, in question answering systems (e.g., IoT-LM \citep{mo2024iot} and Sensor2Text \citep{chen2024sensor2text}) that use \emph{autoregressive Multi-Head Attention (MHA)} for text generation, the key-value (KV) cache is a widely adopted technique to avoid redundant attention computations during inference. To further reduce memory usage and computational overhead, several attention mechanisms have been proposed to share or compress key-value pairs, such as Multi-Query Attention (MQA) \citep{shazeer2019fast}, Grouped-Query Attention (GQA) \citep{ainslie2023gqa}, and Multi-Head Latent Attention (MLA) \citep{liu2024deepseek}. Additionally, attention mechanisms can be optimized for hardware-specific constraints, such as GPU memory and I/O throughput \cite{dao2022flashattention, dao2023flashattention}. 

To mitigate the overhead of multi-layer architectures, LIMU-BERT \citep{xu2021limu} introduces cross-layer parameter sharing, where only the first encoder layer is trained and its parameters are reused across all subsequent layers. This approach drastically reduces the total number of parameters, improving efficiency and enabling deployment on resource-constrained devices. Like sparse attention, cross-layer parameter sharing has yet to be extensively validated at scale in large foundation models, presenting an open question for future research.

Finally, to reduce the computational cost of FFN layers, Mixture-of-Experts (MoE) architectures \citep{cai2025survey} can be used. MoE layers employ sparse gating to activate only a small subset of expert sub-networks per token, rather than processing each token through all experts. This design enables models to scale to billions of parameters while maintaining efficient training and inference. Recent work such as LiteMoE \citep{zhuang2024litemoe} further enhances the efficiency of MoE-based architectures for the on-device deployment. However, MoE structures are typically not applied to other densely parameterized components of Transformer architectures, such as the projection matrices in attention layers. \smallskip

\textit{Architecture with Linear Complexity. } To further address the quadratic complexity of the self-attention mechanism and also a large amount of parameters in the Transformer model, new architecture has been proposed which has linear complexity, such as TNL \citep{qin2024lightning}, HGRN2 \citep{qin2024hgrn2}, cosformer \citep{qin2022cosformer}, and Mamba \cite{gu2023mamba}. Those new architectures  have already been applied in the IoT field. For example, MambaReID \citep{zhang2024mambareid} uses Mamba architecture to address Multi-modal object re-identification task. Their experiments demonstrates that their Mamba based model can achieve the similar accuracy on the person and vehicle re-identification tasks but consuming much less memory.\smallskip

\textit{Adapter Layers. } Adapter layers are parameter-efficient methods during the fine-tuning \citep{houlsby2019parameter}. In this method, only adapter layers, which are simple added structures to the backbone foundation model and have much smaller parameters than the backbone foundation models, are updated during the fine-tuning, which saves a lot of memory and time compared to fine-tuning the whole models \citep{kimura2024efficiency, liu2025llm4wm, li2024sensorllm, xiong2024novel}. Also low-rank adaptation (LoRA) \citep{hu2022lora}, which is the low-rank approximation architecture for the fully connected adapter, can also be used to further reduce the number of parameters of the adapter (e.g., used in LLM4WM \citep{liu2025llm4wm}, LLaSA \citep{imran2024llasa}).

%% file: sections/tables/efficiency_methods_summary.tex
\begin{table*}[t!]
\caption{\textbf{Summary of Efficiency-Enhancing Methods}, with Descriptions and Typically Applicable Paradigms (Prompt-Based, Agent-Based, Training-Based).}
\label{tab:efficiency_summary}
\resizebox{0.8\textwidth}{!}{%
\begin{tabular}{@{}llccc@{}}
\toprule
                                  &                                        & \multicolumn{3}{c}{\textbf{Typically Applicable Paradigm}}             \\ \cmidrule(l){3-5} 
\multirow{-2}{*}{\textbf{Method}} & \multirow{-2}{*}{\textbf{Description}} & \textbf{Prompt} & \textbf{Agent} & \textbf{Training} \\ \midrule
\rowcolor[HTML]{EFEFEF} 
Efficient Prompting &
  \begin{tabular}[c]{@{}l@{}}Minimizing prompt length \\ (especially sensor signals) without significantly \\ compromising the critical information.\end{tabular} &
  \ding{52} &
  &
  \ding{52} \\
\rowcolor[HTML]{FFFFFF} 
Experience Archive &
  \begin{tabular}[c]{@{}l@{}}Using long-term external memory to store \\ task-specific past experiences \\ (context, user query, LLM agent response).\end{tabular} &
   &
  \ding{52} &
  
   \\
\rowcolor[HTML]{EFEFEF} 
Efficiency Model Architecture Design &
  \begin{tabular}[c]{@{}l@{}}Designing efficient architecture \\ to reduce computational requirements.\end{tabular} &
   &
   &
  \ding{52} \\ \bottomrule
\end{tabular}%
}
\end{table*}

%% file: sections/taxonomy/7context_awareness.tex
\section{Criterion II: Context Awareness}
\label{sec:context_awareness}

Context awareness refers to a system’s ability to dynamically adapt its behavior based on situational factors relevant to a specific task \citep{perera2013context, dey2000towards}. In the context of applying FMs to IoT tasks, contextual factors typically fall into two key categories: (1) \emph{environmental context}, which includes physical conditions (e.g., time, location, room brightness, available IoT devices) \citep{guo2024sensor2scene, chen2024towards, cui2024llmind} and virtual settings (e.g., accessible software applications) \citep{liu2024chainstream}; and (2) \emph{user-specific context}, which accounts for individual preferences, schedules, backgrounds, and personalities \citep{yang2025socialmind, yang2025contextagent}.

Context awareness is critical for effectively applying FMs to IoT tasks. 
First, it enables FMs to accurately interpret raw sensor data, which typically consists of unstructured numerical values. Since these values reflect real-world physical states, understanding their meaning requires contextual metadata (e.g., sampling rate, sensor placement, and measurement units). Without this information, FMs struggle to infer the true significance of the data, impairing reasoning and decision-making \citep{an2024iot}. 
Second, since FMs are pre-trained on broad datasets to capture general patterns, they often overlook context-specific cues essential for task performance in real-world environments \cite{yuan2023distilling, prasad2023adapt}. For instance, LLMs may generate overly generic outputs when they lack grounding in environmental context (e.g., availability of physical objects or tools), which limits their ability to produce actionable, situation-appropriate responses.
Third, context-aware foundation models can dynamically adapt to diverse IoT scenarios without requiring separate models for each specific case. This adaptability is especially valuable in IoT environments, where tasks are inherently context-dependent and conditions frequently change. By enabling a single model to generalize across varying situations, context awareness enhances system flexibility and automation, while also reducing the burden of developing and maintaining multiple specialized models.
Finally, incorporating user-specific context (e.g., location, activity, or preferences) allows FM-based IoT systems to personalize services and interactions. This not only improves usability but also enhances user engagement and satisfaction \citep{yang2025socialmind}.

The collected papers propose various methods to enable context awareness in foundation models for IoT applications, which we summarize below. Table \ref{tab:context_aware_summary} also provides an overview of these methods, including their descriptions and the paradigms (prompt-based, agent-based, training-based) to which they are mostly commonly applied. \smallskip

\input{sections/tables/context_awareness_methods_summary}

\noindent\textbf{Data Enrichment. }
Data enrichment involves augmenting formatted and encoded sensor inputs with additional contextual information to help foundation models more accurately interpret the underlying physical meaning behind the data. Critical background details (e.g., units of measurement, sampling rate, and device placement) are essential, as identical sensor readings can have vastly different meanings depending on the context \citep{an2024iot, ji2024hargpt, li2024sensorllm, ouyang2024llmsense, xu2024penetrative}. For instance, sequences of the same length may reflect different durations depending on sampling rates, and identical values could represent entirely different phenomena (e.g., movement speed vs. body temperature) based on the measurement unit. Without this contextual information, foundation models may misinterpret the data and make suboptimal decisions. Therefore, such metadata should be included alongside raw sensor inputs when using prompt-based, agent-based, or training-based approaches. \smallskip

\noindent\textbf{Fine-Tuning. }
One of the most effective ways to achieve context awareness in foundation models, originally trained on general-purpose datasets, is to fine-tune them using data specific to the target environment, user, and task. This process adjusts the model’s parameters to capture context-specific patterns while preserving the general knowledge gained during pretraining \citep{guo2024sensor2scene, tian2025dailyllm}. Fine-tuning significantly improves the model’s ability to interpret and respond to contextual cues. For instance, \citet{guo2024sensor2scene} fine-tuned an LLM for an augmented reality (AR) system using user interaction and feedback data, enabling the model to generate virtual scenes more closely aligned with individual user preferences. \smallskip

\noindent\textbf{Contextual Inputs. }
While fine-tuning can significantly enhance the context awareness of foundation models, it often requires substantial computational resources and large volumes of training data.
A more lightweight alternative when applying prompt-based or agent-based methods is to incorporate environmental and personal context directly into the model inputs as textual prompts \citep{nepal2024mindscape, tian2025dailyllm, cui2024llmind, huang2022inner, song2023llm, singh2022progprompt}. 
For example, \citet{nepal2024mindscape} personalize a journaling system by embedding user context, such as current mood, stress level, and summaries of daily behavior (e.g., screen time, walking duration), into the prompt before passing it to the LLM for reasoning. Similarly, LLMind \citep{cui2024llmind} incorporates user profiles (e.g., background and experience) into the system prompt and leverages the LLM’s role-playing capabilities \citep{chen2024persona} to generate tailored responses.

However, implementing context-aware personalization may raise significant privacy concerns. Developers must be careful of these issues and avoid intrusive practices (e.g., collecting personal information from users’ social media without explicit consent) \citep{yang2025socialmind}. \smallskip

\noindent\textbf{Contextual Memory. }
Contextual information can also be stored in external memory (called \emph{contextual repository} \citep{cui2024llmind} or \emph{semantic memory} \citep{kumar2021semantic, nuxoll2007extending}) to enhance the context awareness when applying agent-based methods. This external repository should be updated in real time or at regular intervals to remain adaptive to contextual changes. 

Several studies \citep{rivkin2024aiot, cui2024llmind, saleh2025usercentrix} have proposed storing environmental or personal data in such memory structures. For example, \citet{yang2025socialmind} uses an LLM to extract persona-related information (e.g., personal interests, experiences, and background) from conversations and stores it in a persona database at the end of each interaction. For new conversational partners, the extracted persona are directly registered into the memory. For known users or previously encountered partners, they use the LLM to check for existing persona in the database. If no conflicting or redundant information is found, the extracted new persona are added. If the persona are semantically similar to existing entries, they are merged. If contradictions are detected, the old information in the database are replaced by the new ones. 
The stored personal information can then be integrated into the LLM’s reasoning, either by accessing it through external memory during inference or by incorporating relevant context into prompts, enhancing the model’s adaptability to individual users. \smallskip

\noindent\textbf{Active Sensing. }
In addition to \emph{passively} receiving environmental data (e.g., sensor data) or user information (e.g., through user interactions), agent-based methods can also \emph{actively} query physical IoT devices (e.g., sensors, robots) to gather additional information when the current context is insufficient for high-level reasoning or decision-making \citep{cui2024llmind, yang2025contextagent}. This is referred to as the \emph{active sensing} capability. For example, in the LLMind \citep{cui2024llmind} paper, an agent-based check-in and security system can prompt a robot to approach and identify a person when the system cannot recognize them from low-resolution images captured by a ceiling-mounted camera or due to occlusion. \smallskip

\noindent\textbf{Iterative Self-Refinement. }
For methods based on LLMs, due to their non-deterministic nature, incorporating contextual information into prompts does not always guarantee fully executable responses or complete satisfaction of personalization requirements. To address this, iterative self-refinement mechanism are proposed to transforms LLMs into a closed-loop system \citep{liu2024chainstream, cui2024llmind, guo2024sensor2scene, xu2024assuring, liu2024agents4plc, yang2024plug}. Specifically, the LLM’s output (e.g., a plan generated by an LLM-agent) receives feedback from the environment or the user if the response is not executable, possibly due to overlooked environmental or user-specific factors. The LLM is then reprompted with information about the execution error or user feedback, generating a revised response. This process repeats until the generated response fully meets the environmental and personalization requirements.

%% file: sections/tables/context_awareness_methods_summary.tex
\begin{table*}[t!]
\caption{\textbf{Summary of Context-Awareness-Enhancing Methods}, with Descriptions and Typically Applicable Paradigms (Prompt-Based, Agent-Based, Training-Based).}
\label{tab:context_aware_summary}
\resizebox{0.8\textwidth}{!}{%
\begin{tabular}{@{}llccc@{}}
\toprule
\rowcolor[HTML]{FFFFFF} 
\cellcolor[HTML]{FFFFFF} &
  \cellcolor[HTML]{FFFFFF} &
  \multicolumn{3}{c}{\cellcolor[HTML]{FFFFFF}\textbf{Typically Applicable Paradigm}} \\ \cmidrule(l){3-5} 
\rowcolor[HTML]{FFFFFF} 
\multirow{-2}{*}{\cellcolor[HTML]{FFFFFF}\textbf{Method}} &
  \multirow{-2}{*}{\cellcolor[HTML]{FFFFFF}\textbf{Description}} &
  \textbf{Prompt} &
  \textbf{Agent} &
  \textbf{Training} \\ \midrule
\rowcolor[HTML]{EFEFEF} 
Data Enrichment &
  \begin{tabular}[c]{@{}l@{}}Augmenting formatted and encoded sensor inputs with additional \\ contextual information to help foundation models more accurately\\  interpret the underlying physical meaning behind the data.\end{tabular} &
  \ding{52} &
  \ding{52} &
  \ding{52} \\
\rowcolor[HTML]{FFFFFF} 
Fine-Tuning &
  \begin{tabular}[c]{@{}l@{}}Fine-tuning pre-trained foundation models using data specific\\ to the target environment, user, and task.\end{tabular} &
   &
   &
  \ding{52} \\
\rowcolor[HTML]{EFEFEF} 
Contextual Inputs &
  \begin{tabular}[c]{@{}l@{}}Incorporating environmental and personal context directly \\ into the model inputs as textual prompts.\end{tabular} &
  \ding{52} &
  \ding{52} &
   \\
\rowcolor[HTML]{FFFFFF} 
Contextual Memory &
  \begin{tabular}[c]{@{}l@{}}Using long-term external memory to store kwledge \\ and information about the agent’s current context.\end{tabular} &
   &
  \ding{52} &
   \\
\rowcolor[HTML]{EFEFEF} 
Active Sensing &
  \begin{tabular}[c]{@{}l@{}}Actively querying physical IoT devices (e.g., sensors, robots) to \\ gather additional information.\end{tabular} &
   &
  \ding{52} &
   \\
\rowcolor[HTML]{FFFFFF} 
Iterative Self-Refinement &
  Refining outputs based on feedback from environments or users. &
   &
  \ding{52} &
   \\ \bottomrule
\end{tabular}%
}
\end{table*}

%% file: sections/taxonomy/8safety.tex
\section{Criterion III: Safety}
\label{sec:safety}

Safety in IoT systems refers to preventing the system and its components from causing physical harm or posing threats, and to protecting the surrounding environment from such risks \citep{atlam2020iot}.
It is worth noting that in other related surveys (e.g., \citet{ma2025safety}), 
safety may also encompass security, which refers to protecting models from external threats such as adversarial, jailbreak attacks. In this survey, we consider them \emph{differently} as \citet{shi2024large} and address security separately in Section \ref{sec:security_privacy}.

Safety is a critical concern in FM-based IoT systems, particularly when model outputs can directly control or influence physical devices (e.g., industrial robots, medical equipment, vehicles, or infrastructure). First, malfunctions, misuse, or insufficient safeguards of FMs may lead to serious consequences, including accidents, equipment failure, or environmental damage. Second, due to their probabilistic nature, FMs may generate outputs that fail to consistently adhere to safety protocols \citep{xu2024assuring, yang2024plug}. This risk increases when task goals are specified at runtime by untrained users, whose instructions may unintentionally conflict with implicit safety constraints. Third, FMs may generate overly general or context-agnostic responses that ignore critical environmental factors, potentially leading to harmful or catastrophic outcomes in the target task, even though the same responses would be safe in other environments or applications.

\begin{figure}[h!]
    \centering
    \includegraphics[width=\columnwidth]{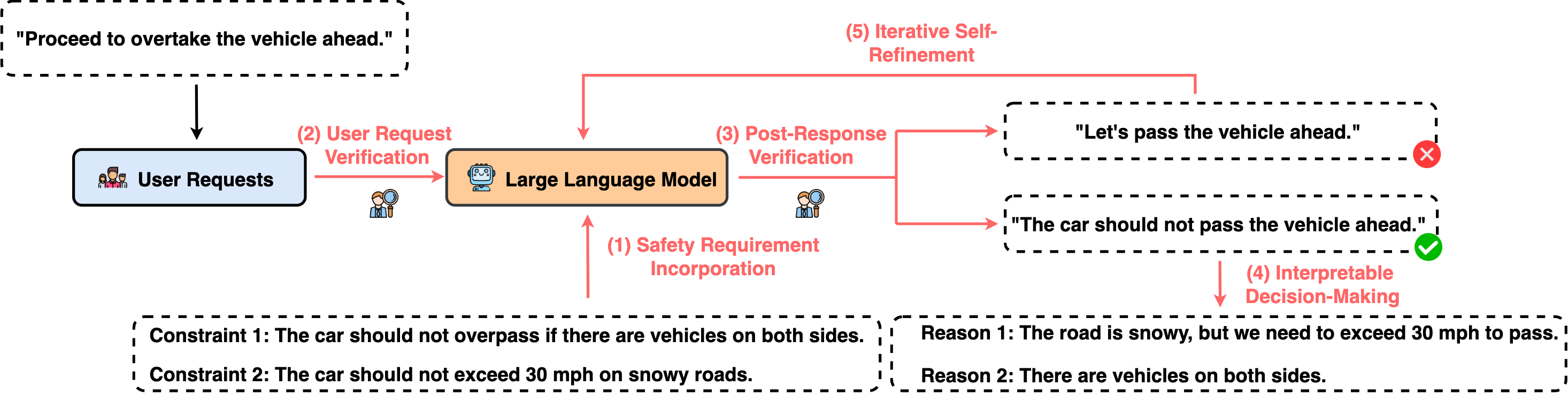}
    \caption{Illustrative example of five safety-enhancing methods introduced in this paper for applying LLMs to IoT tasks, demonstrated through an autonomous driving scenario.}
    \label{fig:safety_flowchart}
\end{figure}

All studies reviewed in this survey focus on enhancing the safety of \emph{LLMs}; therefore, we summarize the commonly used techniques for improving LLM safety in the context of IoT integration. These methods include: (1) \emph{safety requirement incorporation}, (2) \emph{user request verification}, (3) \emph{post-response verification}, (4) \emph{interpretable decision-making}, and (5) \emph{iterative self-refinement}. Table \ref{tab:safety_summary} provides an overview of these methods, including their descriptions and the paradigms (prompt-based, agent-based, training-based) to which they are mostly commonly applied. Figure \ref{fig:safety_flowchart} presents an example of an LLM-controlled autonomous driving system to illustrate these safety enhancement methods. \smallskip

\input{sections/tables/safety_methods_summary}

\noindent\textbf{Safety Requirement Incorporation. }
Embedding safety requirements into LLM inputs, typically via system prompts, is a key strategy for promoting safety awareness \citep{xu2024assuring, ravichandran2025safety, brunke2025semantically}. This approach requires developers and domain experts to proactively identify potential risks and define appropriate constraints and mitigation strategies prior to deployment. In agent-based systems, where external information can be retrieved dynamically from online sources or databases, environment- and task-specific safety requirements can also be automatically incorporated into the prompt to ensure context-aware and adaptive safety compliance. \smallskip

\noindent\textbf{User Request Verification. }
When presented with a task from user requests, an LLM should reason over both the request and the current environmental context to assess whether the task is safe and compliant with predefined safety constraints. Context-aware reasoning is essential, as high-level instructions may not be safe to execute across all situations \citep{cui2024drive, xu2024assuring, ravichandran2025safety}. For example, the model should recognize and reject unsafe user requests, such as driving at 200 miles per hour (322 km/h) on a snowy road.\smallskip

\noindent\textbf{Post-Response Verification. }
After the LLM generates the response, it is crucial to verify that the output adheres to both general and task-specific safety constraints, either manually or automatically. Manual verification, however, can be time-consuming, especially when domain expertise and complex reasoning are required. Therefore, automated verification methods are essential \citep{xu2024assuring, yang2024joint, quartey2024verifiably}. For example, \citet{xu2024assuring} utilizes the Z3 Python API to check safety constraints formulated as first-order logic (FOL) against LLM-generated plans, which are expressed in formal languages such as Linear Temporal Logic (LTL) \citep{xu2024assuring, yang2024plug}. Alternatively, responses can be validated through simulations to ensure safe and correct execution before deployment in real-world environments. \smallskip

\noindent\textbf{Interpretable Decision-Making. }
When the LLM makes the decisions or operates the physical devices, especially in the high-stake scenarios (e.g., healthcare, autonomous driving), interpretability of the final decision is very important. This means being able to provide not only the final decision, but also the reasoning procedure explicitly (e.g., reasoning chain in Chain-of-Thought). This can help humans identify if the generated decision is correct and avoid catastrophes if the LLM intermediate reasoning steps violates the safety rules \citep{cui2024drive, zhang2024large, nouri2024engineering}. \smallskip

\noindent\textbf{Iterative Self-Refinement. }
If an LLM output violates safety constraints, the model should be reprompted and required to regenerate the output iteratively, guided by feedback from either human reviewers or automated verification systems, until the generated response satisfies all predefined safety requirements \citep{xu2024assuring, yang2024plug, yang2024joint}. Alternatively, the LLM can generate initial outputs that are verified for safety, after which the model is iteratively fine-tuned on pairs of user instructions and verified outputs to improve its inherent safety awareness \citep{yang2024joint}.

%% file: sections/tables/safety_methods_summary.tex
\begin{table*}[t!]
\caption{\textbf{Summary of Safety-Enhancing Methods}, with Descriptions and Typically Applicable Paradigms (Prompt-Based, Agent-Based, Training-Based).}
\label{tab:safety_summary}
\resizebox{0.8\textwidth}{!}{%
\begin{tabular}{@{}llccc@{}}
\toprule
\rowcolor[HTML]{FFFFFF} 
\cellcolor[HTML]{FFFFFF} &
  \cellcolor[HTML]{FFFFFF} &
  \multicolumn{3}{c}{\cellcolor[HTML]{FFFFFF}\textbf{Typically Applicable Paradigm}} \\ \cmidrule(l){3-5} 
\rowcolor[HTML]{FFFFFF} 
\multirow{-2}{*}{\cellcolor[HTML]{FFFFFF}\textbf{Method}} &
  \multirow{-2}{*}{\cellcolor[HTML]{FFFFFF}\textbf{Description}} &
  \textbf{Prompt} &
  \textbf{Agent} &
  \textbf{Training} \\ \midrule
\rowcolor[HTML]{EFEFEF} 
Safety Requirement Incorporation &
  Embedding safety requirements into LLM inputs. &
  \ding{52} &
  \ding{52} &
  \ding{52} \\
\rowcolor[HTML]{FFFFFF} 
User Request Verification &
  \begin{tabular}[c]{@{}l@{}}Reasoning over both the user request and the current \\ environmental context to assess whether the task is safe \\ and compliant with predefined safety constraints.\end{tabular} &
   &
  \ding{52} &
   \\
\rowcolor[HTML]{EFEFEF} 
Post-Response Verfication &
  \begin{tabular}[c]{@{}l@{}}Verifying that the output adheres to both general and\\  task-specific safety constraints.\end{tabular} &
   &
  \ding{52} &
   \\
\rowcolor[HTML]{FFFFFF} 
Interpertable Decision-Making &
  \begin{tabular}[c]{@{}l@{}}Generating not only the final decision, \\ but also the reasoning procedure explicitly.\end{tabular} &
  \ding{52} &
  \ding{52} &
   \\
\rowcolor[HTML]{EFEFEF} 
Iterative Self-Refinement &
  Regenerating the output until satisfying the safety requirements. &
   &
  \ding{52} &
  \ding{52} \\ \bottomrule
\end{tabular}%
}
\end{table*}

%% file: sections/taxonomy/9security_and_privacy.tex
\section{Criterion IV: Security and Privacy}
\label{sec:security_privacy}

Security and privacy refer to the protection of systems and data from malicious attacks (e.g., jailbreak attacks, adversarial attacks, PAII attacks) that may lead to unauthorized access, loss of system control, or data leakage \citep{atlam2020iot, chanal2020security}. As discussed in Section \ref{sec:safety}, we distinguish security from safety, which pertains to preventing harmful outputs generated by the system. Notably, in this survey, security and privacy are considered from the perspective of \emph{external} attackers, whereas safety concerns arise from the foundation model’s \emph{internal} probabilistic behavior. Therefore, \emph{even if an FM-based IoT system is secure and private, it may still be unsafe}.

Security and privacy are critical for IoT systems for several reasons. First, the number of IoT devices is growing exponentially, and attacks targeting these systems are becoming increasingly frequent and sophisticated \citep{li2024ids}. Second, IoT devices and networks often collect and transmit sensitive data (e.g., personal health information, location details, and business-critical metrics). Unauthorized access to or manipulation of this data can result in serious consequences, including identity theft, financial loss, and operational disruptions \citep{almodovar2022can, aghaei2022securebert, diaf2024bartpredict, wang2024privacyoracle}. Finally, because IoT devices are typically networked, compromising a single device can jeopardize the security of the entire system.

In this section, we summarize approaches to enhancing IoT security and privacy with foundation models in two contexts: (1) using foundation models to protect the security and privacy of IoT systems without directly participating in downstream tasks (e.g., human activity recognition, robot control), and (2) securing FM-based IoT systems where the foundation model plays a central role in downstream tasks. Table \ref{tab:security_privacy_summary} provides also an overview of these methods, including their descriptions and the paradigms (prompt-based, agent-based, training-based) to which they are mostly commonly applied.\smallskip

\input{sections/tables/security_privacy_methods_summary}

\noindent\textbf{FMs for Anomaly Detection. }
To enhance security and privacy, LLMs have been employed to detect potential anomalies, such as unusual activities in system or network logs and malicious sensor data-sharing requests.
LLMs are commonly used for anomaly detection due to their ability to estimate the likelihood of a sentence or paragraph based on their training data. Specifically, an LLM can be trained on text-based datasets containing normal activity logs or benign user requests. When the model encounters abnormal behavior or malicious requests, it assigns a low probability to these inputs, indicating potential anomalies \citep{almodovar2022can, aghaei2022securebert, diaf2024bartpredict}.

However, defining normal activities or requests in IoT systems based on pre-collected datasets is challenging, particularly in dynamic and heterogeneous environments where “normal” behavior can evolve over time. As a result, LLMs trained on static datasets may misclassify newly emerged benign activities as malicious. Such misclassifications can deny service to legitimate users, reducing system usability and potentially leading to financial losses.

To address this challenge, LLM agents are used to safeguard IoT systems in dynamic environments due to their reasoning and planning capabilities, as well as their ability to search and retrieve relevant information from external sources (e.g., external memory or the internet). These updatable resources allow the system to adapt to previously unseen intrusion patterns, such as zero-day attacks \citep{li2024ids, wang2024privacyoracle}. \smallskip

\noindent\textbf{FMs for Data Obfuscation. }
In addition to detecting and identifying abnormal behavior, FMs can also protect security and privacy of IoT systems through \emph{data obfuscation}, particularly for protecting sensitive personal information.

Data obfuscation is a technique that transforms sensitive data into a less meaningful or recognizable form to prevent unauthorized access and ensure compliance with data protection regulations \citep{bakken2004data}. FMs can support data obfuscation in several ways: (1) identifying and masking sensitive information, such as personally identifiable information (PII) \citep{ouyang2024llmsense}; (2) summarizing raw sensor data into natural language descriptions \citep{ouyang2024llmsense}; (3) synthesizing realistic but non-identifiable sensor data \citep{yang2023privacy}; and (4) enabling LLM agents to generate privacy-preserving data transformation pipelines (e.g., face blurring in video) by selecting tools from a library and planning multi-step workflows \citep{wang2024privacyoracle}.

When leveraging FMs for data obfuscation, two key considerations must be addressed: (1) \emph{Privacy–utility trade-off}: Excessive obfuscation can make transformed data significantly different from the original, making it difficult to recover critical information. This degradation can negatively impact the performance of downstream tasks \citep{yang2023privacy, wang2024privacyoracle}. (2) \emph{Personalization}: Rather than applying uniform rules (e.g., masking biometric identifiers or common activity types), obfuscation should account for individual user preferences regarding what information can be shared. This user-controlled privacy level, known as \emph{privacy preference}, guides the filtering of data based on specific needs \citep{benazir2024maximizing, yang2023privacy, wang2024privacyoracle}. Supporting personalized privacy settings enhances system usability and can be achieved through techniques discussed in Section \ref{sec:context_awareness}. \smallskip

\noindent\textbf{Privacy Considerations for FM-based IoT system. }
Two techniques are commonly used by the literature to protect privacy in FM-based IoT systems:
(1) \emph{Using less interpretable data modalities}: Selecting data types that are difficult for humans to interpret, such as vibration data instead of video for tasks like human activity monitoring, can improve privacy \citep{yang2025socialmind}. However, this approach requires foundation models capable of understanding such non-human-interpretable data, which is more challenging due to the limited availability of labeled training datasets.
(2) \emph{Data obfuscation in edge–cloud architectures}: When the edge-cloud collaboration system \citep{yao2022edge, tian2024edge} is leveraged, data should be obfuscated on \emph{local devices}, using rule-based methods, machine learning models, or lightweight foundation models, before being transmitted to the cloud \citep{ouyang2024llmsense, benazir2024maximizing, yuan2024wip}. The cloud hosts more powerful FMs for task-specific reasoning, but transmitting raw data increases exposure to privacy risks during transit and in cloud environments, where broader access makes data more vulnerable to attacks.

%% file: sections/tables/security_privacy_methods_summary.tex
\begin{table*}[t!]
\caption{\textbf{Summary of Security\&Privacy-Enhancing Methods}, with Descriptions and Typically Applicable Paradigms (Prompt-Based, Agent-Based, Training-Based).}
\label{tab:security_privacy_summary}
\resizebox{0.8\textwidth}{!}{%
\begin{tabular}{@{}llccc@{}}
\toprule
\rowcolor[HTML]{FFFFFF} 
\cellcolor[HTML]{FFFFFF} &
  \cellcolor[HTML]{FFFFFF} &
  \multicolumn{3}{c}{\cellcolor[HTML]{FFFFFF}\textbf{Typically Applicable Paradigm}} \\ \cmidrule(l){3-5} 
\rowcolor[HTML]{FFFFFF} 
\multirow{-2}{*}{\cellcolor[HTML]{FFFFFF}\textbf{Method}} &
  \multirow{-2}{*}{\cellcolor[HTML]{FFFFFF}\textbf{Description}} &
  \multicolumn{1}{c}{\cellcolor[HTML]{FFFFFF}\textbf{Prompt}} &
  \textbf{Agent} &
  \textbf{Training} \\ \midrule
\rowcolor[HTML]{EFEFEF} 
FMs for Anomaly Detection &
  \begin{tabular}[c]{@{}l@{}}Detecting potential anomalies, such as unusual \\ activities in system or network logs and malicious \\ sensor data-sharing requests.\end{tabular} &
   &
  \ding{52} &
  \ding{52} \\
\rowcolor[HTML]{FFFFFF} 
FMs for Data Obfuscation &
  \begin{tabular}[c]{@{}l@{}}Transforming sensitive data into a less meaningful \\ or recognizable form to prevent unauthorized access \\ and ensure compliance with data protection regulations.\end{tabular} &
  \ding{52} &
  \ding{52} &
   \\ \bottomrule
\end{tabular}%
}
\end{table*}

%% file: sections/taxonomy/10evaluation.tex
\section{Evaluation}
\label{sec:evaluation}

In this section, we examine how foundation models are evaluated for IoT tasks. We review the metrics used for both downstream task performance and the four key performance criteria discussed earlier. Additionally, we summarize common evaluation strategies, outlining their applicable scenarios, relevant criteria, and associated advantages and limitations.

\noindent\textbf{Metrics. }
Most evaluation metrics depend on the specific downstream task. We categorize them into four groups and summarize the commonly used metrics for each:
(1) \emph{Classification tasks} (e.g., beam prediction \citep{alikhani2024large, liu2025llm4wm, sheng2025beam}, human activity recognition \citep{chen2024towards, ji2024hargpt, ouyang2025mmbind, xiong2024novel, yan2024language, weng2024large, aboulfotouh2024building, cleland2024leveraging, sun2024ai}): Accuracy, F1 score, precision, recall, and specificity.
(2)	\emph{Regression tasks} (e.g., channel prediction \citep{liu2025llm4wm, liu2024llm4cp, liu2025wifo, fan2025csi}, air quality forecasting \citep{fan2024llmair}): Mean Squared Error (MSE), Mean Absolute Error (MAE), Root Mean Squared Error (RMSE), and Normalized MSE (NMSE).
(3) \emph{Complex IoT tasks involving LLM agents} (e.g., wireless network management): Success rate, which measures the proportion of tasks successfully completed, indicating whether the final objective was achieved \citep{abdallah2024netorchllm, liu2024chainstream, yang2024plug}.
(4) \emph{Sensor QA tasks} (e.g., sensor summarization, ECG report generation) \citep{wan2024meit, li2024sensorllm, mo2024iot, chen2024sensor2text, imran2024llasa, yonekura2024generating, xue2024leveraging}: Natural language processing (NLP) metrics such as BLEU \citep{papineni2002bleu}, ROUGE \citep{lin2004rouge}, and METEOR \citep{banerjee2005meteor}.

We also summarize specific evaluation metrics from the reviewed papers for each performance criterion discussed in this work:
(1)	\emph{Efficiency}:
Convergence speed (e.g., number of epochs to convergence) assesses the training efficiency of fine-tuning-based methods \citep{kimura2024vibrofm, kimura2024efficiency}.
Inference latency measures the time required for the FM-based IoT system to complete a downstream task during inference \citep{yang2023edgefm}.
Memory usage measures the amount of memory consumed during inference \citep{worae2024unified}.
(2)	\emph{Context-Awareness}:
Executability rate evaluates how likely the model’s response can be executed in a specific environment \citep{liu2024chainstream}.
Personalization score assesses how well the response is tailored to an individual’s unique characteristics (e.g., persona, background, schedule) \citep{yang2025socialmind}.
(3)	\emph{Safety}:
Safety rate quantifies the likelihood that a model’s response violates predefined safety constraints \citep{ma2025safety}.
(4) \emph{Security and Privacy}:
For security, the common task is anomaly detection and the corresponding metrics are classification based metrics (e.g., F1, precision, recall) \citep{li2024ids, houssel2024towards, diaf2024bartpredict, almodovar2022can, aghaei2022securebert}.
For privacy, evaluation typically involves the attack accuracy, which measures how accurately an adversarial model can recover sensitive information that is intended to be protected \citep{yang2023privacy, wang2024privacyoracle}. \smallskip

\noindent\textbf{Strategies. }
Common evaluation strategies for foundation models in IoT tasks include: (1) ground-truth comparison, (2) human evaluation, (3) LLM-as-a-judge, and (4) real-world evaluation. We describe each approach in detail below.

\textit{Ground-Truth Comparison.}
The most common approach to evaluating foundation models for IoT tasks is to compare their outputs against ground-truth labels using explicitly annotated datasets for specific downstream tasks. Evaluation tasks typically fall into two categories:
(1) \emph{Closed-ended tasks}: These tasks have unique ground-truth labels. Common examples include classification (e.g., human activity recognition \citep{chen2024towards, ji2024hargpt, ouyang2025mmbind, xiong2024novel, yan2024language, weng2024large, aboulfotouh2024building, cleland2024leveraging, sun2024ai}, beam prediction \citep{alikhani2024large, liu2025llm4wm, sheng2025beam}, disease detection \citep{ho2024remoni}) and regression tasks (e.g., PM2.5 prediction \citep{fan2024llmair}). Performance is measured using metrics such as accuracy for classification, or MAE and RMSE for regression, which quantify the difference between the model’s predictions and the true labels.
(2) \emph{Open-ended tasks}: These are typically found in question-answering scenarios (e.g., “Please summarize the trend of the sensor signal” or "Please generate a report for the ECG."), where multiple valid responses may exist due to the inherent variability in natural language \citep{wan2024meit, li2024sensorllm, mo2024iot, chen2024sensor2text, imran2024llasa, yonekura2024generating, xue2024leveraging}. In such cases, natural language generation metrics like BLEU \citep{papineni2002bleu} and ROUGE \citep{lin2004rouge} are used to assess the semantic similarity between the model’s output and a reference response, which serves as the ground-truth. 

Among the performance criteria discussed in this survey, ground-truth comparison is most commonly used in evaluating \emph{security and privacy}. This is because their tasks often involve classification, such as determining whether a behavior is malicious or benign using foundation models. In contrast, ground-truth comparison is less frequently used for evaluating context awareness and safety, as these tasks are more subjective and influenced by individual preferences and real-world variability, making it difficult to define universally applicable labels.

\textit{Human Evaluation. }
Human evaluation is typically employed in the following scenarios when addressing IoT tasks:
(1) \emph{Open-ended generation tasks}: As noted earlier, open-ended tasks can have multiple valid responses, while datasets usually provide only a single reference answer. Relying solely on automated comparison with a reference may result in incomplete or biased evaluation. Human evaluation serves as the gold standard in these cases, allowing assessment across multiple dimensions (e.g., fluency, coherence, and relevance). It also enables qualitative judgments in situations where evaluation criteria cannot be easily formalized mathematically. Humans may either compare different outputs or assign scalar satisfaction scores to individual responses \citep{imran2024llasa, nie2024llm}.
(2) \emph{Real-world deployment evaluation}: In practical IoT applications, FM-based systems are often validated in real-world environments where ground-truth labels are unavailable \citep{yang2024drhouse, englhardt2024classification, nie2024llm}. To benchmark performance in such settings, human judgment is essential.
(3) \emph{Usability evaluation}: Human evaluation is also critical for assessing the usability of FM-based IoT systems, particularly in personalized applications (see Section \ref{sec:context_awareness}) \citep{guo2024sensor2scene, nie2024llm}. User feedback and experience play a key role in iterative development and refinement of these systems.

Among the performance criteria discussed in this survey, human evaluation is most commonly used to assess \emph{context-awareness} and \emph{safety}. These aspects are inherently subjective and heavily influenced by human experience, as well as situational nuances. For example, in an urban setting, a “safe” action might involve abrupt braking to avoid hitting a pedestrian, while on a highway, such behavior could increase the risk of a collision and be considered unsafe. The appropriate response depends on factors like speed, traffic density, and road conditions. Due to this complexity, single-label quantitative evaluations are often insufficient, making automated assessment challenging and highlighting the need for human judgment in evaluating these criteria.

However, human evaluation is generally slower and more costly than automated methods, especially when complex reasoning or expert knowledge is required (e.g., in disease diagnosis). Furthermore, to ensure meaningful and reliable results, evaluators must establish clear annotation guidelines, evaluate inter-rater reliability, ensure demographic diversity among annotators, and monitor consistency throughout the evaluation process \citep{hallgren2012computing, gwet2014handbook}. 

\textit{LLM-as-a-Judge. }
To address the high cost and slow turnaround of human annotation and evaluation, LLMs have gained popularity as automated judges for open-ended tasks and real-world scenarios. Trained on vast and diverse real-world data, LLMs possess broad general and domain-specific knowledge, allowing them to perform evaluations across a range of domains without task-specific fine-tuning. This makes LLM-based evaluation significantly more scalable and cost-effective \citep{li2024generation, gu2024survey}. Moreover, LLM agents, augmented with reasoning and planning capabilities, have recently emerged as even more powerful evaluators \citep{zhuge2024agent}. In addition, many state-of-the-art commercial LLMs have been aligned with human preferences through reinforcement learning (e.g., RLHF \citep{ouyang2022training}) or alternative methods such as Direct Preference Optimization (DPO) \citep{rafailov2023direct}, making them more representative of human judgment. Thus, \emph{context-awareness} and \emph{safety} can also be assessed by LLM judges.

Despite these advantages, using LLMs as judges comes with inherent limitations and biases, including position bias, length bias, and output instability (e.g., flipping between responses) \citep{zheng2023judging, ye2024justice, wei2024systematic}. These issues must be carefully measured, mitigated, and monitored to ensure fair and reliable evaluation outcomes. Addressing these concerns is crucial for producing trustworthy evaluations that can guide the development and deployment of foundation models in real-world applications. \smallskip

\textit{Real-World Evaluation.}
Real-world evaluation is essential for training and deploying foundation models in IoT applications, as these systems are ultimately intended for use in practical settings such as smart homes, cities, transportation, and healthcare. Testing in real-world environments allows researchers and developers to identify system limitations across diverse scenarios and iteratively improve performance.

During real-world evaluation, it is critical to assess not only the performance criteria outlined in this survey, but also key properties such as generalization (scalability), robustness, and usability in varying levels of complexity. Moreover, \emph{such evaluations must be conducted in a carefully controlled manner to ensure safety for both humans and the environment.}

%% file: sections/5discussion.tex
\section{Discussion}\label{sec:discussion}

In this section, we discuss the limitations of current approaches and propose future directions for more effective application and evaluation of foundation models in IoT tasks. \smallskip

\noindent\textbf{Insufficient Evaluation. }
Current FM-based approaches for IoT tasks often suffer from insufficient evaluation. Below, we highlight this issue from both the \emph{general} perspective as well as the specific perspective from \emph{security and privacy}. \smallskip

\textit{Lack of Cross-Domain Comparison. }
As discussed in the Introduction (section \ref{sec:introduction}), cross-domain comparisons are largely lacking in current research. Many studies evaluate their proposed methods only against simple baselines, neglecting comparisons with more advanced techniques from other IoT domains. This limits our ability to assess the relative strengths and weaknesses of different approaches and hampers practitioners in choosing the most suitable methods for new tasks. Additionally, no existing work evaluates \emph{all} the performance criteria identified in this survey, which are essential for deploying foundation models in real-world IoT applications. To address these gaps, we propose the following: 
(1)\emph{Cross-Domain Comparisons}: Researchers should benchmark their methods not only against basic baselines but also against advanced techniques from other IoT subfields, particularly those highlighted in this survey.
(2) \emph{Generalization Evaluation and Leaderboard Creation}:
A standardized evaluation framework should be established, incorporating the performance criteria, methodologies, datasets, and sensor types commonly used across IoT domains. Developing a shared leaderboard would promote transparent, comprehensive comparisons and help identify the most effective methods for specific criteria, tasks, and sensor modalities.  \smallskip

\textit{Lack of Fine-Grained Evaluation. }
Many components in FM-based IoT systems lack fine-grained evaluation. For example, agent-based methods often involve multiple modules (i.e., tool use, memory, and plan generation) but evaluations typically focus only on the overall system performance. This overlooks critical questions (e.g., whether the agent selects the most appropriate tools, generates accurate plans, or retrieves relevant knowledge for specific tasks and user queries). Without fine-grained evaluation, it is difficult to identify system bottlenecks or guide meaningful improvements. Moving forward, component-level assessments using dedicated metrics are essential to understand each module’s contribution to overall performance and to inform more effective system design. \smallskip

\textit{Lack of Real-World Evaluation. } 
Many studies fail to evaluate their methods in real-world environments using the full set of performance criteria outlined in this survey. This is a significant limitation, as IoT systems operate in complex and variable deployment settings that cannot be fully captured through simulation. Without real-world validation, it is difficult to assess the generalizability, reliability, and robustness of FM-based IoT solutions. 
In the future, we strongly recommend evaluating complete FM-based IoT systems in real-world settings using the full set of criteria defined in this survey. This is essential for identifying practical limitations, informing future improvements, and guiding the development of robust, deployable IoT solutions. \smallskip

\textit{Lack of Security and Privacy Evaluation for FMs. }
While the methods discussed in Section \ref{sec:security_privacy} leverage foundation models (e.g., LLMs, diffusion models \citep{yang2023diffusion}) to defend against security and privacy threats (e.g., personal information inference attacks), the models themselves are inherently vulnerable \citep{ma2025safety, abdali2024securing}. Consequently, using FMs in IoT security applications introduces additional risks. However, the real-world impact of these vulnerabilities in IoT settings remains largely unexplored, and current research offers limited guidance on mitigating such risks. Future work should include more comprehensive evaluations and the development of benchmark datasets to assess security and privacy threats arising from both pretrained and newly proposed FMs in IoT contexts.  \smallskip

\noindent\textbf{Advanced FM Techniques for IoT. }
Advanced techniques in foundation models remain underexplored in the context of IoT. In this section, we highlight four such techniques related to LLMs and LLM agents that have not been widely applied to IoT tasks: \emph{large reasoning models}, \emph{multi-agent system}, \emph{human preference alignment}, and \emph{new model architecture and training objective}. \smallskip

\textit{Large Reasoning Model (LRM).} LRMs (e.g., OpenAI’s o1 \citep{jaech2024openai}, o3 \citep{openai2025o3}, and DeepSeek-R1 \citep{guo2025deepseek}) are a class of LLMs designed specifically for complex reasoning and planning, going beyond the instruction-following and question-answering capabilities of traditional LLMs. Many IoT applications—particularly those involving complex environments or high-stakes scenarios (e.g., healthcare or autonomous driving) require strong contextual reasoning and decision-making. LRMs, which emphasize logical inference over surface-level statistical patterns, hold significant promise for addressing these challenges more effectively. \smallskip

\textit{Multi-Agent System.} A multi-agent system based on LLMs \citep{guo2024large, li2024survey} consists of multiple LLM agents, each equipped with distinct capabilities defined by their roles, tools, and memory modules. 

Compared to single-agent systems (as discussed in Section \ref{sec:agent_based_methods}), LLM-based multi-agent systems are better suited for complex and dynamic IoT environments due to their enhanced scalability, adaptability, fault tolerance, and collaborative capacity.
First, multi-agent systems integrate diverse skills and perspectives, making them well-suited for interdependent IoT scenarios that require coordination and negotiation (e.g., smart cities or supply chain management).
Second, they can dynamically reallocate resources, adapt to new devices, and respond in real time to changing conditions. These capabilities that are difficult to achieve with a single agent make them more effective in unpredictable IoT settings.
Third, agents can cross-validate each other’s outputs, reducing the risk of errors and hallucinations, which is an especially valuable feature for safety-critical applications.
Finally, multi-agent systems can distribute complex or lengthy tasks across agents, preserving coherence over time and across devices, thus overcoming the context window limitations of single LLMs.

However, effective multi-agent deployment requires careful task decomposition. Specifically, it is essential to determine how to partition the overall task into non-overlapping subtasks that align with each agent’s strengths. Additionally, designing mechanisms for inter-agent communication and collaboration is crucial for achieving coherent and coordinated outcomes. These communication strategies should be compatible with the distributed and hierarchical nature of IoT networks, where each node may host a distinct agent, and agents communicate over the network via node-to-node connections. \smallskip

\textit{Human Preference Alignment. } Advanced human preference alignment techniques (e.g., Reinforcement Learning from Human Feedback (RLHF) \citep{ouyang2022training, kaufmann2023survey} and Direct Preference Optimization (DPO) \citep{rafailov2023direct}) have not yet been widely applied to personalize LLMs for IoT tasks. These alignment methods not only enhance personalization, enabling the model to better reflect individual user preferences, but also help embed socially and ethically appropriate behaviors. For example, teaching a model to prioritize waiting for an elderly person to cross the street rather than proceeding immediately can significantly improve the safety and trustworthiness of LLM-based IoT systems operating in real-world environments. \\

\textit{New Model Architecture and Training Objective. } Current FMs used in training-based methods for IoT tasks are primarily based on the Transformer architecture \citep{vaswani2017attention}. However, Transformers may not be the most effective or efficient choice for processing sensor data, even for time-series data \citep{zeng2023transformers, das2023long}, despite their popularity in this domain. This highlights the need for novel architectural designs and training strategies tailored specifically to the characteristics of IoT data. Below, we outline several key considerations for developing such models.

Designing FMs for IoT applications requires careful consideration of the unique characteristics of sensor data. Key considerations include: 
(1) \emph{Architecture choice}: model architecture should be tailored to the unique properties or structures of sensor data. For example, in geo-spatial applications like air quality monitoring \citep{hettige2024airphynet, du2025graph}, graph neural networks instead of Transformer may better capture spatial dependencies. 
(2) \emph{Heterogeneous sensor types}: IoT networks often involve diverse sensor modalities. Designing models that can extract and integrate multi-modal information remains an open challenge. 
(3) \emph{Sparse but salient signals}: Sensor data frequently contain sparse but crucial patterns. Incorporating selective attention or filtering mechanisms may help models focus on high-value segments.
(4) \emph{Metadata utilization}: Contextual information (e.g., sampling rates and device placements) should be explicitly encoded and included alongside raw sensor inputs to improve model understanding of the data collection process. How to encode them effectively into the model architecture is an open problem. 
(5) \emph{Efficiency and robustness}: Models should be computationally lightweight for deployment on edge devices and robust to common issues like missing or noisy data.
(6) \emph{IoT-specific Training Objective}: Many existing pretraining tasks are directly adapted from NLP (e.g., masked token prediction \citep{devlin2019bert}), but these may not suit the continuous, multi-modal nature of sensor data. There is a urgent need to develop IoT-specific unsupervised learning objectives.

%% file: sections/6conclusion.tex
\vspace{-0.05in}
\section{Conclusion}
\label{sec:conclusion}
In this survey, we provide a comprehensive overview of research leveraging foundation models for IoT tasks. We identify four shared performance criteria across diverse IoT applications, outline three key paradigms for applying foundation models, and examine current evaluation strategies for both general performance and individual criteria. Based on this analysis, we highlight several open research challenges and propose future directions. Our work offers guidance for more systematic evaluation, enables cross-domain comparisons, and provides valuable insights for applying and assessing foundation models in emerging IoT scenarios.

%% file: sections/7acknowledgements.tex
\begin{acks}
This research is supported by a UC Merced Spring 2023 Climate Action Seed Competition Grant, CAHSI-Google Institutional Research Program Award, and F3 R\&D GSR Award funded by the US Department of Commerce, Economic Development Administration Build Back Better Regional Challenge.
\end{acks}

%% file: sections/8appendices.tex
\input{sections/appendix/prompt_example}
\input{sections/appendix/method_selection}

%% file: sections/appendix/prompt_example.tex
\section{Example Prompt for Human Activity Recognition Based on IMU Data}
\label{sec:appendix_example_prompt}

\begin{tcolorbox}[colback=gray!5, colframe=gray!40!black, title=System Prompt]
You are an IoT domain expert specializing in human activity recognition (HAR) using inertial measurement unit (IMU) sensor data. Your goal is to accurately determine whether the input IMU data corresponds to a specific human activity, based on provided data and context.\\

The task involves:
\begin{itemize}[noitemsep]
    \item Understanding the sensor data (acceleration and gyroscope) collected from wearable IMU devices.
    \item Using domain knowledge and context (e.g., units, sampling rate, placement) to interpret sensor patterns.
    \item Applying expert reasoning to infer human activity based on motion patterns.
\end{itemize}

You will follow these \textbf{decomposed steps}:
\begin{enumerate}[noitemsep]
    \item Analyze the time series patterns from the accelerometer and gyroscope.
    \item Use context information to interpret the movement (e.g., high acceleration with oscillation = running).
    \item Match the interpreted signal to the target activity class.
    \item Provide the answer in the requested format.
\end{enumerate}

You will be given a prompt that includes:
\begin{itemize}[noitemsep]
    \item A task description
    \item IMU data input
    \item Data collection context
    \item In-context demonstrations
    \item Output format constraints
\end{itemize}

You must \textbf{strictly follow the output format} provided in the user prompt. Do not include any explanation, justification, or additional content in your output.
\end{tcolorbox}


\begin{tcolorbox}[colback=gray!5, colframe=gray!40!black, title=User Prompt]
\textbf{Task:} Determine whether the person is \emph{running} based on the provided IMU data segment. \\

\textbf{Data Collection Context:}
\begin{itemize}[noitemsep]
    \item IMU Device: Worn on the right ankle
    \item Sampling Rate: 50 Hz (i.e., 50 samples per second)
    \item Duration: 10 seconds
    \item Sensors:
    \begin{itemize}[noitemsep]
        \item Accelerometer (X, Y, Z) in m/s²
        \item Gyroscope (X, Y, Z) in deg/s
    \end{itemize}
    \item Units: All values are floating point with 2 decimal places
    \item Format: Each row corresponds to one timestamp
\end{itemize}

\textbf{In-Context Demonstration \#1}
\begin{Verbatim}[fontsize=\small]
Task: Is the person running?  
IMU Segment:
Time | Acc_X | Acc_Y | Acc_Z | Gyro_X | Gyro_Y | Gyro_Z  
-----|-------|-------|-------|--------|--------|--------
0.00 | 0.20  | 9.81  | 0.10  | 0.01   | 0.02   | 0.00  
...  | ...   | ...   | ...   | ...    | ...    | ...    
[brief burst of high-frequency, high-magnitude motion]  
Answer: Yes
\end{Verbatim}

\textbf{In-Context Demonstration \#2}
\begin{Verbatim}[fontsize=\small]
Task: Is the person running?  
IMU Segment:
Time | Acc_X | Acc_Y | Acc_Z | Gyro_X | Gyro_Y | Gyro_Z  
-----|-------|-------|-------|--------|--------|--------
0.00 | 0.01  | 9.80  | 0.00  | 0.00   | 0.01   | 0.01  
...  | ...   | ...   | ...   | ...    | ...    | ...    
[mostly flat and stable readings over time]  
Answer: No
\end{Verbatim}

\textbf{Your Task}
\begin{Verbatim}[fontsize=\small]
Task: Is the person running?  
IMU Segment:
Time | Acc_X | Acc_Y | Acc_Z | Gyro_X | Gyro_Y | Gyro_Z  
-----|-------|-------|-------|--------|--------|--------
0.00 | 0.03  | 9.79  | 0.01  | 0.01   | 0.00   | 0.01  
0.02 | 0.05  | 9.82  | 0.03  | 0.01   | 0.01   | 0.00  
0.04 | 0.04  | 9.78  | 0.02  | 0.00   | 0.02   | 0.01  
...  | ...   | ...   | ...   | ...    | ...    | ...    
\end{Verbatim}

\textbf{Output Format:} \\
Please \textbf{only respond with} \texttt{Yes} or \texttt{No}. \textbf{Do not} include any explanation or additional content.
\end{tcolorbox}

%% file: sections/appendix/method_selection.tex
\section{Method Selection}
\label{sec:appendix_method_selection}

\begin{figure}[H]
    \centering
    \includegraphics[width=\linewidth]{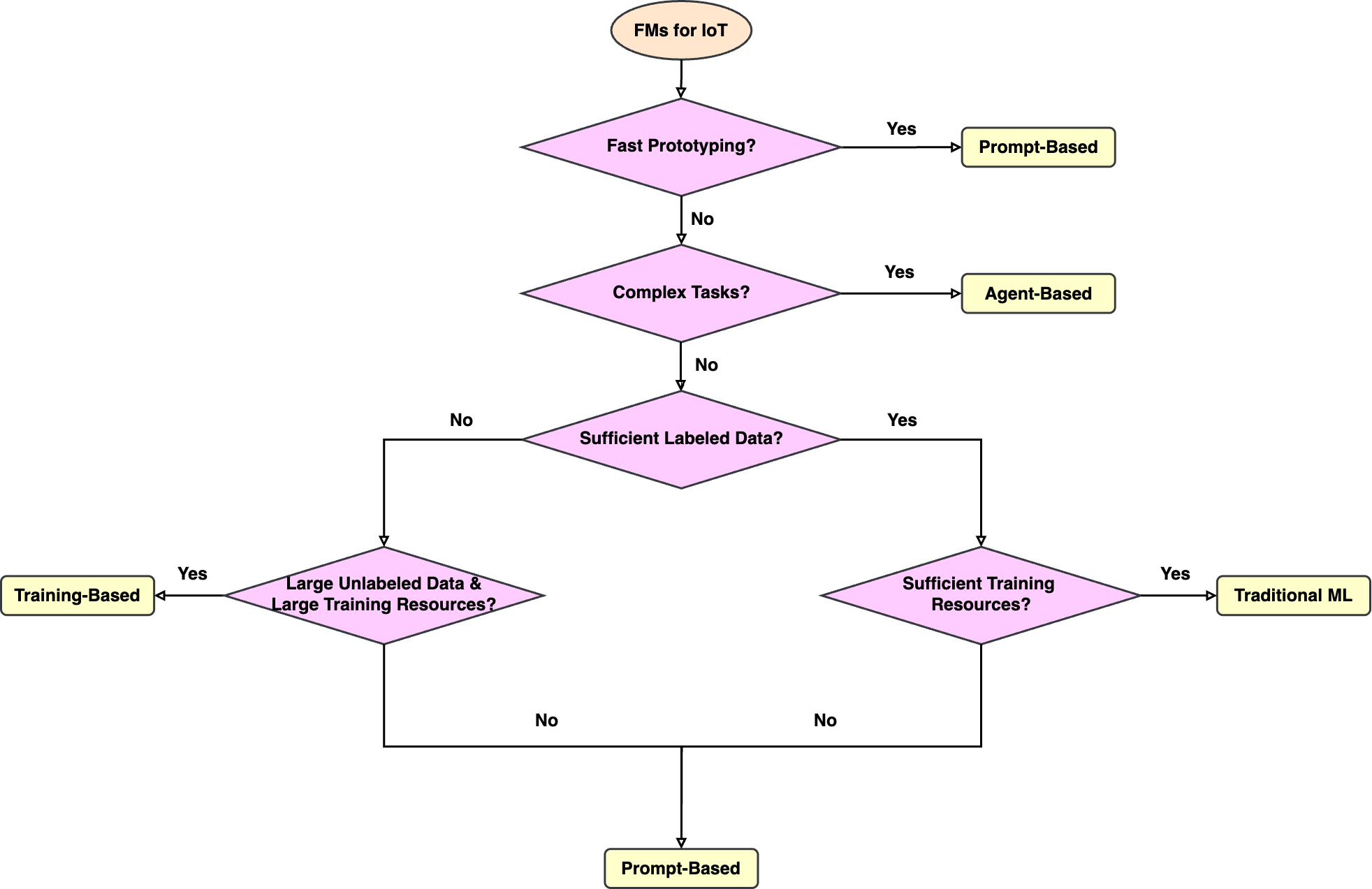}
    \caption{Method selection decision tree. The decision tree considers the three most commonly used frameworks for applying foundation models to IoT tasks: \emph{prompt-based methods}, \emph{agent-based methods}, and \emph{training-based methods}, as well as \emph{traditional supervised learning methods}. The selection is guided by six key dimensions introduced in Section \ref{sec:method_comparison} computation requirement (CR), error rate on specific tasks (ER), task specificity (TS), development time (DT), labeled data requirement (LDR), and unlabeled data requirement (UDR). While the decision tree provides general guidance, practitioners and researchers should adapt it to specific scenarios, particularly when additional constraints such as security and privacy are involved.}
    \label{fig:method_selection}
\end{figure}

%% file: main.bbl

\begin{thebibliography}{229}


\ifx \showCODEN    \undefined \def \showCODEN     #1{\unskip}     \fi
\ifx \showISBNx    \undefined \def \showISBNx     #1{\unskip}     \fi
\ifx \showISBNxiii \undefined \def \showISBNxiii  #1{\unskip}     \fi
\ifx \showISSN     \undefined \def \showISSN      #1{\unskip}     \fi
\ifx \showLCCN     \undefined \def \showLCCN      #1{\unskip}     \fi
\ifx \shownote     \undefined \def \shownote      #1{#1}          \fi
\ifx \showarticletitle \undefined \def \showarticletitle #1{#1}   \fi
\ifx \showURL      \undefined \def \showURL       {\relax}        \fi
\providecommand\bibfield[2]{#2}
\providecommand\bibinfo[2]{#2}
\providecommand\natexlab[1]{#1}
\providecommand\showeprint[2][]{arXiv:#2}

\bibitem[Abdali et~al\mbox{.}(2024)]%
        {abdali2024securing}
\bibfield{author}{\bibinfo{person}{Sara Abdali}, \bibinfo{person}{Richard Anarfi}, \bibinfo{person}{CJ Barberan}, {and} \bibinfo{person}{Jia He}.} \bibinfo{year}{2024}\natexlab{}.
\newblock \showarticletitle{Securing large language models: Threats, vulnerabilities and responsible practices}.
\newblock \bibinfo{journal}{\emph{arXiv preprint arXiv:2403.12503}} (\bibinfo{year}{2024}).
\newblock


\bibitem[Abdallah et~al\mbox{.}(2024)]%
        {abdallah2024netorchllm}
\bibfield{author}{\bibinfo{person}{Asmaa Abdallah}, \bibinfo{person}{Abdullatif Albaseer}, \bibinfo{person}{Abdulkadir Celik}, \bibinfo{person}{Mohamed Abdallah}, {and} \bibinfo{person}{Ahmed~M Eltawil}.} \bibinfo{year}{2024}\natexlab{}.
\newblock \showarticletitle{NetOrchLLM: Mastering Wireless Network Orchestration with Large Language Models}.
\newblock \bibinfo{journal}{\emph{arXiv preprint arXiv:2412.10107}} (\bibinfo{year}{2024}).
\newblock


\bibitem[Aboulfotouh et~al\mbox{.}(2024)]%
        {aboulfotouh2024building}
\bibfield{author}{\bibinfo{person}{Ahmed Aboulfotouh}, \bibinfo{person}{Ashkan Eshaghbeigi}, {and} \bibinfo{person}{Hatem Abou-Zeid}.} \bibinfo{year}{2024}\natexlab{}.
\newblock \showarticletitle{Building 6G Radio Foundation Models with Transformer Architectures}.
\newblock \bibinfo{journal}{\emph{arXiv preprint arXiv:2411.09996}} (\bibinfo{year}{2024}).
\newblock


\bibitem[Achiam et~al\mbox{.}(2023)]%
        {achiam2023gpt}
\bibfield{author}{\bibinfo{person}{Josh Achiam}, \bibinfo{person}{Steven Adler}, \bibinfo{person}{Sandhini Agarwal}, \bibinfo{person}{Lama Ahmad}, \bibinfo{person}{Ilge Akkaya}, \bibinfo{person}{Florencia~Leoni Aleman}, \bibinfo{person}{Diogo Almeida}, \bibinfo{person}{Janko Altenschmidt}, \bibinfo{person}{Sam Altman}, \bibinfo{person}{Shyamal Anadkat}, {et~al\mbox{.}}} \bibinfo{year}{2023}\natexlab{}.
\newblock \showarticletitle{Gpt-4 technical report}.
\newblock \bibinfo{journal}{\emph{arXiv preprint arXiv:2303.08774}} (\bibinfo{year}{2023}).
\newblock


\bibitem[Agarwal et~al\mbox{.}(2024)]%
        {agarwal2024medhalu}
\bibfield{author}{\bibinfo{person}{Vibhor Agarwal}, \bibinfo{person}{Yiqiao Jin}, \bibinfo{person}{Mohit Chandra}, \bibinfo{person}{Munmun De~Choudhury}, \bibinfo{person}{Srijan Kumar}, {and} \bibinfo{person}{Nishanth Sastry}.} \bibinfo{year}{2024}\natexlab{}.
\newblock \showarticletitle{MedHalu: Hallucinations in Responses to Healthcare Queries by Large Language Models}.
\newblock \bibinfo{journal}{\emph{arXiv preprint arXiv:2409.19492}} (\bibinfo{year}{2024}).
\newblock


\bibitem[Aghaei et~al\mbox{.}(2022)]%
        {aghaei2022securebert}
\bibfield{author}{\bibinfo{person}{Ehsan Aghaei}, \bibinfo{person}{Xi Niu}, \bibinfo{person}{Waseem Shadid}, {and} \bibinfo{person}{Ehab Al-Shaer}.} \bibinfo{year}{2022}\natexlab{}.
\newblock \showarticletitle{Securebert: A domain-specific language model for cybersecurity}. In \bibinfo{booktitle}{\emph{International Conference on Security and Privacy in Communication Systems}}. Springer, \bibinfo{pages}{39--56}.
\newblock


\bibitem[Aghzal et~al\mbox{.}(2025)]%
        {aghzal2025survey}
\bibfield{author}{\bibinfo{person}{Mohamed Aghzal}, \bibinfo{person}{Erion Plaku}, \bibinfo{person}{Gregory~J Stein}, {and} \bibinfo{person}{Ziyu Yao}.} \bibinfo{year}{2025}\natexlab{}.
\newblock \showarticletitle{A survey on large language models for automated planning}.
\newblock \bibinfo{journal}{\emph{arXiv preprint arXiv:2502.12435}} (\bibinfo{year}{2025}).
\newblock


\bibitem[Ahmad et~al\mbox{.}(2023)]%
        {ahmad2023creating}
\bibfield{author}{\bibinfo{person}{Muhammad~Aurangzeb Ahmad}, \bibinfo{person}{Ilker Yaramis}, {and} \bibinfo{person}{Taposh~Dutta Roy}.} \bibinfo{year}{2023}\natexlab{}.
\newblock \showarticletitle{Creating trustworthy llms: Dealing with hallucinations in healthcare ai}.
\newblock \bibinfo{journal}{\emph{arXiv preprint arXiv:2311.01463}} (\bibinfo{year}{2023}).
\newblock


\bibitem[Ainslie et~al\mbox{.}(2023)]%
        {ainslie2023gqa}
\bibfield{author}{\bibinfo{person}{Joshua Ainslie}, \bibinfo{person}{James Lee-Thorp}, \bibinfo{person}{Michiel De~Jong}, \bibinfo{person}{Yury Zemlyanskiy}, \bibinfo{person}{Federico Lebr{\'o}n}, {and} \bibinfo{person}{Sumit Sanghai}.} \bibinfo{year}{2023}\natexlab{}.
\newblock \showarticletitle{Gqa: Training generalized multi-query transformer models from multi-head checkpoints}.
\newblock \bibinfo{journal}{\emph{arXiv preprint arXiv:2305.13245}} (\bibinfo{year}{2023}).
\newblock


\bibitem[Al-Zihad et~al\mbox{.}(2017)]%
        {al2017bandwidth}
\bibfield{author}{\bibinfo{person}{Md Al-Zihad}, \bibinfo{person}{Saad~Ahmed Akash}, \bibinfo{person}{Tamal Adhikary}, {and} \bibinfo{person}{Md~Abdur Razzaque}.} \bibinfo{year}{2017}\natexlab{}.
\newblock \showarticletitle{Bandwidth allocation and computation offloading for service specific IoT edge devices}. In \bibinfo{booktitle}{\emph{2017 IEEE Region 10 Humanitarian Technology Conference (R10-HTC)}}. IEEE, \bibinfo{pages}{516--519}.
\newblock


\bibitem[Alikhani et~al\mbox{.}(2024)]%
        {alikhani2024large}
\bibfield{author}{\bibinfo{person}{Sadjad Alikhani}, \bibinfo{person}{Gouranga Charan}, {and} \bibinfo{person}{Ahmed Alkhateeb}.} \bibinfo{year}{2024}\natexlab{}.
\newblock \showarticletitle{Large wireless model (LWM): A foundation model for wireless channels}.
\newblock \bibinfo{journal}{\emph{arXiv preprint arXiv:2411.08872}} (\bibinfo{year}{2024}).
\newblock


\bibitem[Almodovar et~al\mbox{.}(2022)]%
        {almodovar2022can}
\bibfield{author}{\bibinfo{person}{Crispin Almodovar}, \bibinfo{person}{Fariza Sabrina}, \bibinfo{person}{Sarvnaz Karimi}, {and} \bibinfo{person}{Salahuddin Azad}.} \bibinfo{year}{2022}\natexlab{}.
\newblock \showarticletitle{Can language models help in system security? investigating log anomaly detection using BERT}. In \bibinfo{booktitle}{\emph{Proceedings of the 20th Annual Workshop of the Australasian Language Technology Association}}. \bibinfo{pages}{139--147}.
\newblock


\bibitem[An et~al\mbox{.}(2024)]%
        {an2024iot}
\bibfield{author}{\bibinfo{person}{Tuo An}, \bibinfo{person}{Yunjiao Zhou}, \bibinfo{person}{Han Zou}, {and} \bibinfo{person}{Jianfei Yang}.} \bibinfo{year}{2024}\natexlab{}.
\newblock \showarticletitle{Iot-llm: Enhancing real-world iot task reasoning with large language models}.
\newblock \bibinfo{journal}{\emph{arXiv preprint arXiv:2410.02429}} (\bibinfo{year}{2024}).
\newblock


\bibitem[Anokhin et~al\mbox{.}(2024)]%
        {anokhin2024arigraph}
\bibfield{author}{\bibinfo{person}{Petr Anokhin}, \bibinfo{person}{Nikita Semenov}, \bibinfo{person}{Artyom Sorokin}, \bibinfo{person}{Dmitry Evseev}, \bibinfo{person}{Andrey Kravchenko}, \bibinfo{person}{Mikhail Burtsev}, {and} \bibinfo{person}{Evgeny Burnaev}.} \bibinfo{year}{2024}\natexlab{}.
\newblock \showarticletitle{Arigraph: Learning knowledge graph world models with episodic memory for llm agents}.
\newblock \bibinfo{journal}{\emph{arXiv preprint arXiv:2407.04363}} (\bibinfo{year}{2024}).
\newblock


\bibitem[Arrotta et~al\mbox{.}(2024)]%
        {arrotta2024contextgpt}
\bibfield{author}{\bibinfo{person}{Luca Arrotta}, \bibinfo{person}{Claudio Bettini}, \bibinfo{person}{Gabriele Civitarese}, {and} \bibinfo{person}{Michele Fiori}.} \bibinfo{year}{2024}\natexlab{}.
\newblock \showarticletitle{Contextgpt: Infusing llms knowledge into neuro-symbolic activity recognition models}. In \bibinfo{booktitle}{\emph{2024 IEEE International Conference on Smart Computing (SMARTCOMP)}}. IEEE, \bibinfo{pages}{55--62}.
\newblock


\bibitem[Atlam and Wills(2020)]%
        {atlam2020iot}
\bibfield{author}{\bibinfo{person}{Hany~F Atlam} {and} \bibinfo{person}{Gary~B Wills}.} \bibinfo{year}{2020}\natexlab{}.
\newblock \showarticletitle{IoT security, privacy, safety and ethics}.
\newblock \bibinfo{journal}{\emph{Digital twin technologies and smart cities}} (\bibinfo{year}{2020}), \bibinfo{pages}{123--149}.
\newblock


\bibitem[Bachute and Subhedar(2021)]%
        {bachute2021autonomous}
\bibfield{author}{\bibinfo{person}{Mrinal~R Bachute} {and} \bibinfo{person}{Javed~M Subhedar}.} \bibinfo{year}{2021}\natexlab{}.
\newblock \showarticletitle{Autonomous driving architectures: insights of machine learning and deep learning algorithms}.
\newblock \bibinfo{journal}{\emph{Machine Learning with Applications}}  \bibinfo{volume}{6} (\bibinfo{year}{2021}), \bibinfo{pages}{100164}.
\newblock


\bibitem[Bakken et~al\mbox{.}(2004)]%
        {bakken2004data}
\bibfield{author}{\bibinfo{person}{David~E Bakken}, \bibinfo{person}{R Rarameswaran}, \bibinfo{person}{Douglas~M Blough}, \bibinfo{person}{Andy~A Franz}, {and} \bibinfo{person}{Ty~J Palmer}.} \bibinfo{year}{2004}\natexlab{}.
\newblock \showarticletitle{Data obfuscation: Anonymity and desensitization of usable data sets}.
\newblock \bibinfo{journal}{\emph{IEEE Security \& Privacy}} \bibinfo{volume}{2}, \bibinfo{number}{6} (\bibinfo{year}{2004}), \bibinfo{pages}{34--41}.
\newblock


\bibitem[Banerjee and Lavie(2005)]%
        {banerjee2005meteor}
\bibfield{author}{\bibinfo{person}{Satanjeev Banerjee} {and} \bibinfo{person}{Alon Lavie}.} \bibinfo{year}{2005}\natexlab{}.
\newblock \showarticletitle{METEOR: An automatic metric for MT evaluation with improved correlation with human judgments}. In \bibinfo{booktitle}{\emph{Proceedings of the acl workshop on intrinsic and extrinsic evaluation measures for machine translation and/or summarization}}. \bibinfo{pages}{65--72}.
\newblock


\bibitem[Baris et~al\mbox{.}(2025)]%
        {baris2025foundation}
\bibfield{author}{\bibinfo{person}{Ozan Baris}, \bibinfo{person}{Yizhuo Chen}, \bibinfo{person}{Gaofeng Dong}, \bibinfo{person}{Liying Han}, \bibinfo{person}{Tomoyoshi Kimura}, \bibinfo{person}{Pengrui Quan}, \bibinfo{person}{Ruijie Wang}, \bibinfo{person}{Tianchen Wang}, \bibinfo{person}{Tarek Abdelzaher}, \bibinfo{person}{Mario Berg{\'e}s}, {et~al\mbox{.}}} \bibinfo{year}{2025}\natexlab{}.
\newblock \showarticletitle{Foundation Models for CPS-IoT: Opportunities and Challenges}.
\newblock \bibinfo{journal}{\emph{arXiv preprint arXiv:2501.16368}} (\bibinfo{year}{2025}).
\newblock


\bibitem[Benazir and Lin(2024)]%
        {benazir2024maximizing}
\bibfield{author}{\bibinfo{person}{Afsara Benazir} {and} \bibinfo{person}{Felix~Xiaozhu Lin}.} \bibinfo{year}{2024}\natexlab{}.
\newblock \showarticletitle{Maximizing the Capabilities of Tiny Speech Foundation Models in a Privacy Preserving Manner}. In \bibinfo{booktitle}{\emph{Proceedings of the 30th Annual International Conference on Mobile Computing and Networking}}. \bibinfo{pages}{1677--1679}.
\newblock


\bibitem[Benos et~al\mbox{.}(2021)]%
        {benos2021machine}
\bibfield{author}{\bibinfo{person}{Lefteris Benos}, \bibinfo{person}{Aristotelis~C Tagarakis}, \bibinfo{person}{Georgios Dolias}, \bibinfo{person}{Remigio Berruto}, \bibinfo{person}{Dimitrios Kateris}, {and} \bibinfo{person}{Dionysis Bochtis}.} \bibinfo{year}{2021}\natexlab{}.
\newblock \showarticletitle{Machine learning in agriculture: A comprehensive updated review}.
\newblock \bibinfo{journal}{\emph{Sensors}} \bibinfo{volume}{21}, \bibinfo{number}{11} (\bibinfo{year}{2021}), \bibinfo{pages}{3758}.
\newblock


\bibitem[Bhat et~al\mbox{.}(2025)]%
        {bhat2025llm}
\bibfield{author}{\bibinfo{person}{Akshat Bhat}, \bibinfo{person}{Aishee Mondal}, {and} \bibinfo{person}{Aniket Tripathy}.} \bibinfo{year}{2025}\natexlab{}.
\newblock \showarticletitle{LLM Agents for Internet of Things (IoT) Applications}.
\newblock \bibinfo{journal}{\emph{https://openreview.net/forum?id=BikB3f8ByV}} (\bibinfo{year}{2025}).
\newblock


\bibitem[Bian et~al\mbox{.}(2022)]%
        {bian2022machine}
\bibfield{author}{\bibinfo{person}{Jiang Bian}, \bibinfo{person}{Abdullah Al~Arafat}, \bibinfo{person}{Haoyi Xiong}, \bibinfo{person}{Jing Li}, \bibinfo{person}{Li Li}, \bibinfo{person}{Hongyang Chen}, \bibinfo{person}{Jun Wang}, \bibinfo{person}{Dejing Dou}, {and} \bibinfo{person}{Zhishan Guo}.} \bibinfo{year}{2022}\natexlab{}.
\newblock \showarticletitle{Machine learning in real-time Internet of Things (IoT) systems: A survey}.
\newblock \bibinfo{journal}{\emph{IEEE Internet of Things Journal}} \bibinfo{volume}{9}, \bibinfo{number}{11} (\bibinfo{year}{2022}), \bibinfo{pages}{8364--8386}.
\newblock


\bibitem[Bommasani et~al\mbox{.}(2021)]%
        {bommasani2021opportunities}
\bibfield{author}{\bibinfo{person}{Rishi Bommasani}, \bibinfo{person}{Drew~A Hudson}, \bibinfo{person}{Ehsan Adeli}, \bibinfo{person}{Russ Altman}, \bibinfo{person}{Simran Arora}, \bibinfo{person}{Sydney von Arx}, \bibinfo{person}{Michael~S Bernstein}, \bibinfo{person}{Jeannette Bohg}, \bibinfo{person}{Antoine Bosselut}, \bibinfo{person}{Emma Brunskill}, {et~al\mbox{.}}} \bibinfo{year}{2021}\natexlab{}.
\newblock \showarticletitle{On the opportunities and risks of foundation models}.
\newblock \bibinfo{journal}{\emph{arXiv preprint arXiv:2108.07258}} (\bibinfo{year}{2021}).
\newblock


\bibitem[Brown et~al\mbox{.}(2020)]%
        {brown2020language}
\bibfield{author}{\bibinfo{person}{Tom Brown}, \bibinfo{person}{Benjamin Mann}, \bibinfo{person}{Nick Ryder}, \bibinfo{person}{Melanie Subbiah}, \bibinfo{person}{Jared~D Kaplan}, \bibinfo{person}{Prafulla Dhariwal}, \bibinfo{person}{Arvind Neelakantan}, \bibinfo{person}{Pranav Shyam}, \bibinfo{person}{Girish Sastry}, \bibinfo{person}{Amanda Askell}, {et~al\mbox{.}}} \bibinfo{year}{2020}\natexlab{}.
\newblock \showarticletitle{Language models are few-shot learners}.
\newblock \bibinfo{journal}{\emph{Advances in neural information processing systems}}  \bibinfo{volume}{33} (\bibinfo{year}{2020}), \bibinfo{pages}{1877--1901}.
\newblock


\bibitem[Brunke et~al\mbox{.}(2025)]%
        {brunke2025semantically}
\bibfield{author}{\bibinfo{person}{Lukas Brunke}, \bibinfo{person}{Yanni Zhang}, \bibinfo{person}{Ralf R{\"o}mer}, \bibinfo{person}{Jack Naimer}, \bibinfo{person}{Nikola Staykov}, \bibinfo{person}{Siqi Zhou}, {and} \bibinfo{person}{Angela~P Schoellig}.} \bibinfo{year}{2025}\natexlab{}.
\newblock \showarticletitle{Semantically safe robot manipulation: From semantic scene understanding to motion safeguards}.
\newblock \bibinfo{journal}{\emph{IEEE Robotics and Automation Letters}} (\bibinfo{year}{2025}).
\newblock


\bibitem[Bzai et~al\mbox{.}(2022)]%
        {bzai2022machine}
\bibfield{author}{\bibinfo{person}{Jamal Bzai}, \bibinfo{person}{Furqan Alam}, \bibinfo{person}{Arwa Dhafer}, \bibinfo{person}{Miroslav Bojovi{\'c}}, \bibinfo{person}{Saleh~M Altowaijri}, \bibinfo{person}{Imran~Khan Niazi}, {and} \bibinfo{person}{Rashid Mehmood}.} \bibinfo{year}{2022}\natexlab{}.
\newblock \showarticletitle{Machine learning-enabled internet of things (iot): Data, applications, and industry perspective}.
\newblock \bibinfo{journal}{\emph{Electronics}} \bibinfo{volume}{11}, \bibinfo{number}{17} (\bibinfo{year}{2022}), \bibinfo{pages}{2676}.
\newblock


\bibitem[Cai et~al\mbox{.}(2025)]%
        {cai2025survey}
\bibfield{author}{\bibinfo{person}{Weilin Cai}, \bibinfo{person}{Juyong Jiang}, \bibinfo{person}{Fan Wang}, \bibinfo{person}{Jing Tang}, \bibinfo{person}{Sunghun Kim}, {and} \bibinfo{person}{Jiayi Huang}.} \bibinfo{year}{2025}\natexlab{}.
\newblock \showarticletitle{A Survey on Mixture of Experts in Large Language Models}.
\newblock \bibinfo{journal}{\emph{IEEE Transactions on Knowledge and Data Engineering}} (\bibinfo{year}{2025}).
\newblock


\bibitem[Chanal and Kakkasageri(2020)]%
        {chanal2020security}
\bibfield{author}{\bibinfo{person}{Poornima~M Chanal} {and} \bibinfo{person}{Mahabaleshwar~S Kakkasageri}.} \bibinfo{year}{2020}\natexlab{}.
\newblock \showarticletitle{Security and privacy in IoT: a survey}.
\newblock \bibinfo{journal}{\emph{Wireless Personal Communications}} \bibinfo{volume}{115}, \bibinfo{number}{2} (\bibinfo{year}{2020}), \bibinfo{pages}{1667--1693}.
\newblock


\bibitem[Chang et~al\mbox{.}(2024)]%
        {chang2024efficient}
\bibfield{author}{\bibinfo{person}{Kaiyan Chang}, \bibinfo{person}{Songcheng Xu}, \bibinfo{person}{Chenglong Wang}, \bibinfo{person}{Yingfeng Luo}, \bibinfo{person}{Xiaoqian Liu}, \bibinfo{person}{Tong Xiao}, {and} \bibinfo{person}{Jingbo Zhu}.} \bibinfo{year}{2024}\natexlab{}.
\newblock \showarticletitle{Efficient prompting methods for large language models: A survey}.
\newblock \bibinfo{journal}{\emph{arXiv preprint arXiv:2404.01077}} (\bibinfo{year}{2024}).
\newblock


\bibitem[Chen et~al\mbox{.}(2024c)]%
        {chen2024persona}
\bibfield{author}{\bibinfo{person}{Jiangjie Chen}, \bibinfo{person}{Xintao Wang}, \bibinfo{person}{Rui Xu}, \bibinfo{person}{Siyu Yuan}, \bibinfo{person}{Yikai Zhang}, \bibinfo{person}{Wei Shi}, \bibinfo{person}{Jian Xie}, \bibinfo{person}{Shuang Li}, \bibinfo{person}{Ruihan Yang}, \bibinfo{person}{Tinghui Zhu}, {et~al\mbox{.}}} \bibinfo{year}{2024}\natexlab{c}.
\newblock \showarticletitle{From persona to personalization: A survey on role-playing language agents}.
\newblock \bibinfo{journal}{\emph{arXiv preprint arXiv:2404.18231}} (\bibinfo{year}{2024}).
\newblock


\bibitem[Chen et~al\mbox{.}(2024a)]%
        {chen2024sensor2text}
\bibfield{author}{\bibinfo{person}{Wenqiang Chen}, \bibinfo{person}{Jiaxuan Cheng}, \bibinfo{person}{Leyao Wang}, \bibinfo{person}{Wei Zhao}, {and} \bibinfo{person}{Wojciech Matusik}.} \bibinfo{year}{2024}\natexlab{a}.
\newblock \showarticletitle{Sensor2Text: Enabling Natural Language Interactions for Daily Activity Tracking Using Wearable Sensors}.
\newblock \bibinfo{journal}{\emph{Proceedings of the ACM on Interactive, Mobile, Wearable and Ubiquitous Technologies}} \bibinfo{volume}{8}, \bibinfo{number}{4} (\bibinfo{year}{2024}), \bibinfo{pages}{1--26}.
\newblock


\bibitem[Chen et~al\mbox{.}(2022)]%
        {chen2022program}
\bibfield{author}{\bibinfo{person}{Wenhu Chen}, \bibinfo{person}{Xueguang Ma}, \bibinfo{person}{Xinyi Wang}, {and} \bibinfo{person}{William~W Cohen}.} \bibinfo{year}{2022}\natexlab{}.
\newblock \showarticletitle{Program of thoughts prompting: Disentangling computation from reasoning for numerical reasoning tasks}.
\newblock \bibinfo{journal}{\emph{arXiv preprint arXiv:2211.12588}} (\bibinfo{year}{2022}).
\newblock


\bibitem[Chen et~al\mbox{.}(2024b)]%
        {chen2024towards}
\bibfield{author}{\bibinfo{person}{Xi Chen}, \bibinfo{person}{Julien Cumin}, \bibinfo{person}{Fano Ramparany}, {and} \bibinfo{person}{Dominique Vaufreydaz}.} \bibinfo{year}{2024}\natexlab{b}.
\newblock \showarticletitle{Towards llm-powered ambient sensor based multi-person human activity recognition}. In \bibinfo{booktitle}{\emph{2024 IEEE 30th International Conference on Parallel and Distributed Systems (ICPADS)}}. IEEE, \bibinfo{pages}{609--616}.
\newblock


\bibitem[Chen et~al\mbox{.}(2023)]%
        {chen2023learning}
\bibfield{author}{\bibinfo{person}{Xiang Chen}, \bibinfo{person}{Hao Li}, \bibinfo{person}{Mingqiang Li}, {and} \bibinfo{person}{Jinshan Pan}.} \bibinfo{year}{2023}\natexlab{}.
\newblock \showarticletitle{Learning a sparse transformer network for effective image deraining}. In \bibinfo{booktitle}{\emph{Proceedings of the IEEE/CVF conference on computer vision and pattern recognition}}. \bibinfo{pages}{5896--5905}.
\newblock


\bibitem[Child et~al\mbox{.}(2019)]%
        {child2019generating}
\bibfield{author}{\bibinfo{person}{Rewon Child}, \bibinfo{person}{Scott Gray}, \bibinfo{person}{Alec Radford}, {and} \bibinfo{person}{Ilya Sutskever}.} \bibinfo{year}{2019}\natexlab{}.
\newblock \showarticletitle{Generating long sequences with sparse transformers}.
\newblock \bibinfo{journal}{\emph{arXiv preprint arXiv:1904.10509}} (\bibinfo{year}{2019}).
\newblock


\bibitem[Choi et~al\mbox{.}(2019)]%
        {choi2019gaussian}
\bibfield{author}{\bibinfo{person}{Jiwoong Choi}, \bibinfo{person}{Dayoung Chun}, \bibinfo{person}{Hyun Kim}, {and} \bibinfo{person}{Hyuk-Jae Lee}.} \bibinfo{year}{2019}\natexlab{}.
\newblock \showarticletitle{Gaussian yolov3: An accurate and fast object detector using localization uncertainty for autonomous driving}. In \bibinfo{booktitle}{\emph{Proceedings of the IEEE/CVF International conference on computer vision}}. \bibinfo{pages}{502--511}.
\newblock


\bibitem[Cleland et~al\mbox{.}(2024)]%
        {cleland2024leveraging}
\bibfield{author}{\bibinfo{person}{Ian Cleland}, \bibinfo{person}{Luke Nugent}, \bibinfo{person}{Federico Cruciani}, {and} \bibinfo{person}{Chris Nugent}.} \bibinfo{year}{2024}\natexlab{}.
\newblock \showarticletitle{Leveraging large language models for activity recognition in smart environments}. In \bibinfo{booktitle}{\emph{2024 International Conference on Activity and Behavior Computing (ABC)}}. IEEE, \bibinfo{pages}{1--8}.
\newblock


\bibitem[Cui et~al\mbox{.}(2024b)]%
        {cui2024drive}
\bibfield{author}{\bibinfo{person}{Can Cui}, \bibinfo{person}{Yunsheng Ma}, \bibinfo{person}{Xu Cao}, \bibinfo{person}{Wenqian Ye}, {and} \bibinfo{person}{Ziran Wang}.} \bibinfo{year}{2024}\natexlab{b}.
\newblock \showarticletitle{Drive as you speak: Enabling human-like interaction with large language models in autonomous vehicles}. In \bibinfo{booktitle}{\emph{Proceedings of the IEEE/CVF Winter Conference on Applications of Computer Vision}}. \bibinfo{pages}{902--909}.
\newblock


\bibitem[Cui et~al\mbox{.}(2024a)]%
        {cui2024llmind}
\bibfield{author}{\bibinfo{person}{Hongwei Cui}, \bibinfo{person}{Yuyang Du}, \bibinfo{person}{Qun Yang}, \bibinfo{person}{Yulin Shao}, {and} \bibinfo{person}{Soung~Chang Liew}.} \bibinfo{year}{2024}\natexlab{a}.
\newblock \showarticletitle{Llmind: Orchestrating ai and iot with llm for complex task execution}.
\newblock \bibinfo{journal}{\emph{IEEE Communications Magazine}} (\bibinfo{year}{2024}).
\newblock


\bibitem[Da et~al\mbox{.}(2024)]%
        {da2024prompt}
\bibfield{author}{\bibinfo{person}{Longchao Da}, \bibinfo{person}{Minquan Gao}, \bibinfo{person}{Hao Mei}, {and} \bibinfo{person}{Hua Wei}.} \bibinfo{year}{2024}\natexlab{}.
\newblock \showarticletitle{Prompt to transfer: Sim-to-real transfer for traffic signal control with prompt learning}. In \bibinfo{booktitle}{\emph{Proceedings of the AAAI Conference on Artificial Intelligence}}, Vol.~\bibinfo{volume}{38}. \bibinfo{pages}{82--90}.
\newblock


\bibitem[Dai(2019)]%
        {dai2019hybridnet}
\bibfield{author}{\bibinfo{person}{Xuerui Dai}.} \bibinfo{year}{2019}\natexlab{}.
\newblock \showarticletitle{HybridNet: A fast vehicle detection system for autonomous driving}.
\newblock \bibinfo{journal}{\emph{Signal Processing: Image Communication}}  \bibinfo{volume}{70} (\bibinfo{year}{2019}), \bibinfo{pages}{79--88}.
\newblock


\bibitem[Dao(2023)]%
        {dao2023flashattention}
\bibfield{author}{\bibinfo{person}{Tri Dao}.} \bibinfo{year}{2023}\natexlab{}.
\newblock \showarticletitle{Flashattention-2: Faster attention with better parallelism and work partitioning}.
\newblock \bibinfo{journal}{\emph{arXiv preprint arXiv:2307.08691}} (\bibinfo{year}{2023}).
\newblock


\bibitem[Dao et~al\mbox{.}(2022)]%
        {dao2022flashattention}
\bibfield{author}{\bibinfo{person}{Tri Dao}, \bibinfo{person}{Dan Fu}, \bibinfo{person}{Stefano Ermon}, \bibinfo{person}{Atri Rudra}, {and} \bibinfo{person}{Christopher R{\'e}}.} \bibinfo{year}{2022}\natexlab{}.
\newblock \showarticletitle{Flashattention: Fast and memory-efficient exact attention with io-awareness}.
\newblock \bibinfo{journal}{\emph{Advances in neural information processing systems}}  \bibinfo{volume}{35} (\bibinfo{year}{2022}), \bibinfo{pages}{16344--16359}.
\newblock


\bibitem[Das et~al\mbox{.}(2023)]%
        {das2023long}
\bibfield{author}{\bibinfo{person}{Abhimanyu Das}, \bibinfo{person}{Weihao Kong}, \bibinfo{person}{Andrew Leach}, \bibinfo{person}{Shaan Mathur}, \bibinfo{person}{Rajat Sen}, {and} \bibinfo{person}{Rose Yu}.} \bibinfo{year}{2023}\natexlab{}.
\newblock \showarticletitle{Long-term forecasting with tide: Time-series dense encoder}.
\newblock \bibinfo{journal}{\emph{arXiv preprint arXiv:2304.08424}} (\bibinfo{year}{2023}).
\newblock


\bibitem[Devlin et~al\mbox{.}(2019)]%
        {devlin2019bert}
\bibfield{author}{\bibinfo{person}{Jacob Devlin}, \bibinfo{person}{Ming-Wei Chang}, \bibinfo{person}{Kenton Lee}, {and} \bibinfo{person}{Kristina Toutanova}.} \bibinfo{year}{2019}\natexlab{}.
\newblock \showarticletitle{Bert: Pre-training of deep bidirectional transformers for language understanding}. In \bibinfo{booktitle}{\emph{Proceedings of the 2019 conference of the North American chapter of the association for computational linguistics: human language technologies, volume 1 (long and short papers)}}. \bibinfo{pages}{4171--4186}.
\newblock


\bibitem[Dey et~al\mbox{.}(2000)]%
        {dey2000towards}
\bibfield{author}{\bibinfo{person}{Anind~K Dey}, \bibinfo{person}{Gregory~D Abowd}, {et~al\mbox{.}}} \bibinfo{year}{2000}\natexlab{}.
\newblock \showarticletitle{Towards a better understanding of context and context-awareness}. In \bibinfo{booktitle}{\emph{CHI 2000 workshop on the what, who, where, when, and how of context-awareness}}, Vol.~\bibinfo{volume}{4}. \bibinfo{pages}{1--6}.
\newblock


\bibitem[Diaf et~al\mbox{.}(2024)]%
        {diaf2024bartpredict}
\bibfield{author}{\bibinfo{person}{Alaeddine Diaf}, \bibinfo{person}{Abdelaziz~Amara Korba}, \bibinfo{person}{Nour~Elislem Karabadji}, {and} \bibinfo{person}{Yacine Ghamri-Doudane}.} \bibinfo{year}{2024}\natexlab{}.
\newblock \showarticletitle{BARTPredict: Empowering IoT Security with LLM-Driven Cyber Threat Prediction}. In \bibinfo{booktitle}{\emph{GLOBECOM 2024-2024 IEEE Global Communications Conference}}. IEEE, \bibinfo{pages}{1239--1244}.
\newblock


\bibitem[Dogan et~al\mbox{.}(2011)]%
        {dogan2011autonomous}
\bibfield{author}{\bibinfo{person}{{\"U}r{\"u}n Dogan}, \bibinfo{person}{Johann Edelbrunner}, {and} \bibinfo{person}{Ioannis Iossifidis}.} \bibinfo{year}{2011}\natexlab{}.
\newblock \showarticletitle{Autonomous driving: A comparison of machine learning techniques by means of the prediction of lane change behavior}. In \bibinfo{booktitle}{\emph{2011 ieee international conference on robotics and biomimetics}}. IEEE, \bibinfo{pages}{1837--1843}.
\newblock


\bibitem[Dona et~al\mbox{.}(2024)]%
        {dona2024llms}
\bibfield{author}{\bibinfo{person}{Malsha Ashani~Mahawatta Dona}, \bibinfo{person}{Beatriz Cabrero-Daniel}, \bibinfo{person}{Yinan Yu}, {and} \bibinfo{person}{Christian Berger}.} \bibinfo{year}{2024}\natexlab{}.
\newblock \showarticletitle{LLMs Can Check Their Own Results to Mitigate Hallucinations in Traffic Understanding Tasks}. In \bibinfo{booktitle}{\emph{IFIP International Conference on Testing Software and Systems}}. Springer, \bibinfo{pages}{114--130}.
\newblock


\bibitem[Dou et~al\mbox{.}(2023)]%
        {dou2023towards}
\bibfield{author}{\bibinfo{person}{Fei Dou}, \bibinfo{person}{Jin Ye}, \bibinfo{person}{Geng Yuan}, \bibinfo{person}{Qin Lu}, \bibinfo{person}{Wei Niu}, \bibinfo{person}{Haijian Sun}, \bibinfo{person}{Le Guan}, \bibinfo{person}{Guoyu Lu}, \bibinfo{person}{Gengchen Mai}, \bibinfo{person}{Ninghao Liu}, {et~al\mbox{.}}} \bibinfo{year}{2023}\natexlab{}.
\newblock \showarticletitle{Towards artificial general intelligence (agi) in the internet of things (iot): Opportunities and challenges}.
\newblock \bibinfo{journal}{\emph{arXiv preprint arXiv:2309.07438}} (\bibinfo{year}{2023}).
\newblock


\bibitem[Du et~al\mbox{.}(2025)]%
        {du2025graph}
\bibfield{author}{\bibinfo{person}{Shangjie Du}, \bibinfo{person}{Hui Wei}, \bibinfo{person}{Dong~Yoon Lee}, \bibinfo{person}{Zhizhang Hu}, {and} \bibinfo{person}{Shijia Pan}.} \bibinfo{year}{2025}\natexlab{}.
\newblock \showarticletitle{Graph-Based Physics-Guided Urban PM2. 5 Air Quality Imputation with Constrained Monitoring Data}.
\newblock \bibinfo{journal}{\emph{ACM Transactions on Sensor Networks}} (\bibinfo{year}{2025}).
\newblock


\bibitem[Englhardt et~al\mbox{.}(2024)]%
        {englhardt2024classification}
\bibfield{author}{\bibinfo{person}{Zachary Englhardt}, \bibinfo{person}{Chengqian Ma}, \bibinfo{person}{Margaret~E Morris}, \bibinfo{person}{Chun-Cheng Chang}, \bibinfo{person}{Xuhai"~Orson" Xu}, \bibinfo{person}{Lianhui Qin}, \bibinfo{person}{Daniel McDuff}, \bibinfo{person}{Xin Liu}, \bibinfo{person}{Shwetak Patel}, {and} \bibinfo{person}{Vikram Iyer}.} \bibinfo{year}{2024}\natexlab{}.
\newblock \showarticletitle{From classification to clinical insights: Towards analyzing and reasoning about mobile and behavioral health data with large language models}.
\newblock \bibinfo{journal}{\emph{Proceedings of the ACM on Interactive, Mobile, Wearable and Ubiquitous Technologies}} \bibinfo{volume}{8}, \bibinfo{number}{2} (\bibinfo{year}{2024}), \bibinfo{pages}{1--25}.
\newblock


\bibitem[Fan et~al\mbox{.}(2024)]%
        {fan2024llmair}
\bibfield{author}{\bibinfo{person}{Jinxiao Fan}, \bibinfo{person}{Haolin Chu}, \bibinfo{person}{Liang Liu}, {and} \bibinfo{person}{Huadong Ma}.} \bibinfo{year}{2024}\natexlab{}.
\newblock \showarticletitle{LLMAir: Adaptive Reprogramming Large Language Model for Air Quality Prediction}. In \bibinfo{booktitle}{\emph{2024 IEEE 30th International Conference on Parallel and Distributed Systems (ICPADS)}}. IEEE, \bibinfo{pages}{423--430}.
\newblock


\bibitem[Fan et~al\mbox{.}(2025)]%
        {fan2025csi}
\bibfield{author}{\bibinfo{person}{Shilong Fan}, \bibinfo{person}{Zhenyu Liu}, \bibinfo{person}{Xinyu Gu}, {and} \bibinfo{person}{Haozhen Li}.} \bibinfo{year}{2025}\natexlab{}.
\newblock \showarticletitle{Csi-LLM: A novel downlink channel prediction method aligned with LLM pre-training}. In \bibinfo{booktitle}{\emph{2025 IEEE Wireless Communications and Networking Conference (WCNC)}}. IEEE, \bibinfo{pages}{1--6}.
\newblock


\bibitem[Ferrag et~al\mbox{.}(2025)]%
        {ferrag2025llm}
\bibfield{author}{\bibinfo{person}{Mohamed~Amine Ferrag}, \bibinfo{person}{Norbert Tihanyi}, {and} \bibinfo{person}{Merouane Debbah}.} \bibinfo{year}{2025}\natexlab{}.
\newblock \showarticletitle{From LLM Reasoning to Autonomous AI Agents: A Comprehensive Review}.
\newblock \bibinfo{journal}{\emph{arXiv preprint arXiv:2504.19678}} (\bibinfo{year}{2025}).
\newblock


\bibitem[Gage(1994)]%
        {gage1994new}
\bibfield{author}{\bibinfo{person}{Philip Gage}.} \bibinfo{year}{1994}\natexlab{}.
\newblock \showarticletitle{A new algorithm for data compression}.
\newblock \bibinfo{journal}{\emph{The C Users Journal}} \bibinfo{volume}{12}, \bibinfo{number}{2} (\bibinfo{year}{1994}), \bibinfo{pages}{23--38}.
\newblock


\bibitem[Georgiou et~al\mbox{.}(2017)]%
        {georgiou2017iot}
\bibfield{author}{\bibinfo{person}{Kyriakos Georgiou}, \bibinfo{person}{Samuel Xavier-de Souza}, {and} \bibinfo{person}{Kerstin Eder}.} \bibinfo{year}{2017}\natexlab{}.
\newblock \showarticletitle{The IoT energy challenge: A software perspective}.
\newblock \bibinfo{journal}{\emph{IEEE Embedded Systems Letters}} \bibinfo{volume}{10}, \bibinfo{number}{3} (\bibinfo{year}{2017}), \bibinfo{pages}{53--56}.
\newblock


\bibitem[Ghassemi et~al\mbox{.}(2024)]%
        {ghassemi2024multi}
\bibfield{author}{\bibinfo{person}{Mohammad Ghassemi}, \bibinfo{person}{Han Zhang}, \bibinfo{person}{Ali Afana}, \bibinfo{person}{Akram~Bin Sediq}, {and} \bibinfo{person}{Melike Erol-Kantarci}.} \bibinfo{year}{2024}\natexlab{}.
\newblock \showarticletitle{Multi-modal transformer and reinforcement learning-based beam management}.
\newblock \bibinfo{journal}{\emph{IEEE Networking Letters}} (\bibinfo{year}{2024}).
\newblock


\bibitem[Grigorescu et~al\mbox{.}(2020)]%
        {grigorescu2020survey}
\bibfield{author}{\bibinfo{person}{Sorin Grigorescu}, \bibinfo{person}{Bogdan Trasnea}, \bibinfo{person}{Tiberiu Cocias}, {and} \bibinfo{person}{Gigel Macesanu}.} \bibinfo{year}{2020}\natexlab{}.
\newblock \showarticletitle{A survey of deep learning techniques for autonomous driving}.
\newblock \bibinfo{journal}{\emph{Journal of field robotics}} \bibinfo{volume}{37}, \bibinfo{number}{3} (\bibinfo{year}{2020}), \bibinfo{pages}{362--386}.
\newblock


\bibitem[Gu and Dao(2023)]%
        {gu2023mamba}
\bibfield{author}{\bibinfo{person}{Albert Gu} {and} \bibinfo{person}{Tri Dao}.} \bibinfo{year}{2023}\natexlab{}.
\newblock \showarticletitle{Mamba: Linear-time sequence modeling with selective state spaces}.
\newblock \bibinfo{journal}{\emph{arXiv preprint arXiv:2312.00752}} (\bibinfo{year}{2023}).
\newblock


\bibitem[Gu et~al\mbox{.}(2024)]%
        {gu2024survey}
\bibfield{author}{\bibinfo{person}{Jiawei Gu}, \bibinfo{person}{Xuhui Jiang}, \bibinfo{person}{Zhichao Shi}, \bibinfo{person}{Hexiang Tan}, \bibinfo{person}{Xuehao Zhai}, \bibinfo{person}{Chengjin Xu}, \bibinfo{person}{Wei Li}, \bibinfo{person}{Yinghan Shen}, \bibinfo{person}{Shengjie Ma}, \bibinfo{person}{Honghao Liu}, {et~al\mbox{.}}} \bibinfo{year}{2024}\natexlab{}.
\newblock \showarticletitle{A survey on llm-as-a-judge}.
\newblock \bibinfo{journal}{\emph{arXiv preprint arXiv:2411.15594}} (\bibinfo{year}{2024}).
\newblock


\bibitem[Guo et~al\mbox{.}(2025)]%
        {guo2025deepseek}
\bibfield{author}{\bibinfo{person}{Daya Guo}, \bibinfo{person}{Dejian Yang}, \bibinfo{person}{Haowei Zhang}, \bibinfo{person}{Junxiao Song}, \bibinfo{person}{Ruoyu Zhang}, \bibinfo{person}{Runxin Xu}, \bibinfo{person}{Qihao Zhu}, \bibinfo{person}{Shirong Ma}, \bibinfo{person}{Peiyi Wang}, \bibinfo{person}{Xiao Bi}, {et~al\mbox{.}}} \bibinfo{year}{2025}\natexlab{}.
\newblock \showarticletitle{Deepseek-r1: Incentivizing reasoning capability in llms via reinforcement learning}.
\newblock \bibinfo{journal}{\emph{arXiv preprint arXiv:2501.12948}} (\bibinfo{year}{2025}).
\newblock


\bibitem[Guo et~al\mbox{.}(2024a)]%
        {guo2024large}
\bibfield{author}{\bibinfo{person}{Taicheng Guo}, \bibinfo{person}{Xiuying Chen}, \bibinfo{person}{Yaqi Wang}, \bibinfo{person}{Ruidi Chang}, \bibinfo{person}{Shichao Pei}, \bibinfo{person}{Nitesh~V Chawla}, \bibinfo{person}{Olaf Wiest}, {and} \bibinfo{person}{Xiangliang Zhang}.} \bibinfo{year}{2024}\natexlab{a}.
\newblock \showarticletitle{Large language model based multi-agents: A survey of progress and challenges}.
\newblock \bibinfo{journal}{\emph{arXiv preprint arXiv:2402.01680}} (\bibinfo{year}{2024}).
\newblock


\bibitem[Guo et~al\mbox{.}(2024b)]%
        {guo2024sensor2scene}
\bibfield{author}{\bibinfo{person}{Yunqi Guo}, \bibinfo{person}{Kaiyuan Hou}, \bibinfo{person}{Zhenyu Yan}, \bibinfo{person}{Hongkai Chen}, \bibinfo{person}{Guoliang Xing}, {and} \bibinfo{person}{Xiaofan Jiang}.} \bibinfo{year}{2024}\natexlab{b}.
\newblock \showarticletitle{Sensor2Scene: Foundation Model-Driven Interactive Realities}. In \bibinfo{booktitle}{\emph{2024 IEEE International Workshop on Foundation Models for Cyber-Physical Systems \& Internet of Things (FMSys)}}. IEEE, \bibinfo{pages}{13--19}.
\newblock


\bibitem[Gwet(2014)]%
        {gwet2014handbook}
\bibfield{author}{\bibinfo{person}{Kilem~L Gwet}.} \bibinfo{year}{2014}\natexlab{}.
\newblock \bibinfo{booktitle}{\emph{Handbook of inter-rater reliability: The definitive guide to measuring the extent of agreement among raters}}.
\newblock \bibinfo{publisher}{Advanced Analytics, LLC}.
\newblock


\bibitem[Habehh and Gohel(2021)]%
        {habehh2021machine}
\bibfield{author}{\bibinfo{person}{Hafsa Habehh} {and} \bibinfo{person}{Suril Gohel}.} \bibinfo{year}{2021}\natexlab{}.
\newblock \showarticletitle{Machine learning in healthcare}.
\newblock \bibinfo{journal}{\emph{Current genomics}} \bibinfo{volume}{22}, \bibinfo{number}{4} (\bibinfo{year}{2021}), \bibinfo{pages}{291--300}.
\newblock


\bibitem[Hallgren(2012)]%
        {hallgren2012computing}
\bibfield{author}{\bibinfo{person}{Kevin~A Hallgren}.} \bibinfo{year}{2012}\natexlab{}.
\newblock \showarticletitle{Computing inter-rater reliability for observational data: an overview and tutorial}.
\newblock \bibinfo{journal}{\emph{Tutorials in quantitative methods for psychology}} \bibinfo{volume}{8}, \bibinfo{number}{1} (\bibinfo{year}{2012}), \bibinfo{pages}{23}.
\newblock


\bibitem[Hesham et~al\mbox{.}(2024)]%
        {hesham2024localingua}
\bibfield{author}{\bibinfo{person}{Ahmed Hesham}, \bibinfo{person}{Eman Samir}, \bibinfo{person}{Hamada Rizk}, {and} \bibinfo{person}{Moustafa Youssef}.} \bibinfo{year}{2024}\natexlab{}.
\newblock \showarticletitle{LocaLingua: Leveraging Language Models for Cross-Building WiFi Mapping}. In \bibinfo{booktitle}{\emph{Proceedings of the 32nd ACM International Conference on Advances in Geographic Information Systems}}. \bibinfo{pages}{723--724}.
\newblock


\bibitem[Hettige et~al\mbox{.}(2024)]%
        {hettige2024airphynet}
\bibfield{author}{\bibinfo{person}{Kethmi~Hirushini Hettige}, \bibinfo{person}{Jiahao Ji}, \bibinfo{person}{Shili Xiang}, \bibinfo{person}{Cheng Long}, \bibinfo{person}{Gao Cong}, {and} \bibinfo{person}{Jingyuan Wang}.} \bibinfo{year}{2024}\natexlab{}.
\newblock \showarticletitle{Airphynet: Harnessing physics-guided neural networks for air quality prediction}.
\newblock \bibinfo{journal}{\emph{arXiv preprint arXiv:2402.03784}} (\bibinfo{year}{2024}).
\newblock


\bibitem[Ho et~al\mbox{.}(2019)]%
        {ho2019axial}
\bibfield{author}{\bibinfo{person}{Jonathan Ho}, \bibinfo{person}{Nal Kalchbrenner}, \bibinfo{person}{Dirk Weissenborn}, {and} \bibinfo{person}{Tim Salimans}.} \bibinfo{year}{2019}\natexlab{}.
\newblock \showarticletitle{Axial attention in multidimensional transformers}.
\newblock \bibinfo{journal}{\emph{arXiv preprint arXiv:1912.12180}} (\bibinfo{year}{2019}).
\newblock


\bibitem[Ho et~al\mbox{.}(2024)]%
        {ho2024remoni}
\bibfield{author}{\bibinfo{person}{Thanh~Cong Ho}, \bibinfo{person}{Farah Kharrat}, \bibinfo{person}{Abderrazek Abid}, \bibinfo{person}{Fakhri Karray}, {and} \bibinfo{person}{Anis Koubaa}.} \bibinfo{year}{2024}\natexlab{}.
\newblock \showarticletitle{Remoni: An autonomous system integrating wearables and multimodal large language models for enhanced remote health monitoring}. In \bibinfo{booktitle}{\emph{2024 IEEE International Symposium on Medical Measurements and Applications (MeMeA)}}. IEEE, \bibinfo{pages}{1--6}.
\newblock


\bibitem[Houlsby et~al\mbox{.}(2019)]%
        {houlsby2019parameter}
\bibfield{author}{\bibinfo{person}{Neil Houlsby}, \bibinfo{person}{Andrei Giurgiu}, \bibinfo{person}{Stanislaw Jastrzebski}, \bibinfo{person}{Bruna Morrone}, \bibinfo{person}{Quentin De~Laroussilhe}, \bibinfo{person}{Andrea Gesmundo}, \bibinfo{person}{Mona Attariyan}, {and} \bibinfo{person}{Sylvain Gelly}.} \bibinfo{year}{2019}\natexlab{}.
\newblock \showarticletitle{Parameter-efficient transfer learning for NLP}. In \bibinfo{booktitle}{\emph{International conference on machine learning}}. PMLR, \bibinfo{pages}{2790--2799}.
\newblock


\bibitem[Houssel et~al\mbox{.}(2024)]%
        {houssel2024towards}
\bibfield{author}{\bibinfo{person}{Paul~RB Houssel}, \bibinfo{person}{Priyanka Singh}, \bibinfo{person}{Siamak Layeghy}, {and} \bibinfo{person}{Marius Portmann}.} \bibinfo{year}{2024}\natexlab{}.
\newblock \showarticletitle{Towards explainable network intrusion detection using large language models}.
\newblock \bibinfo{journal}{\emph{arXiv preprint arXiv:2408.04342}} (\bibinfo{year}{2024}).
\newblock


\bibitem[Hu et~al\mbox{.}(2022)]%
        {hu2022lora}
\bibfield{author}{\bibinfo{person}{Edward~J Hu}, \bibinfo{person}{Yelong Shen}, \bibinfo{person}{Phillip Wallis}, \bibinfo{person}{Zeyuan Allen-Zhu}, \bibinfo{person}{Yuanzhi Li}, \bibinfo{person}{Shean Wang}, \bibinfo{person}{Lu Wang}, \bibinfo{person}{Weizhu Chen}, {et~al\mbox{.}}} \bibinfo{year}{2022}\natexlab{}.
\newblock \showarticletitle{Lora: Low-rank adaptation of large language models.}
\newblock \bibinfo{journal}{\emph{ICLR}} \bibinfo{volume}{1}, \bibinfo{number}{2} (\bibinfo{year}{2022}), \bibinfo{pages}{3}.
\newblock


\bibitem[Hu et~al\mbox{.}(2024)]%
        {hu2024hiagent}
\bibfield{author}{\bibinfo{person}{Mengkang Hu}, \bibinfo{person}{Tianxing Chen}, \bibinfo{person}{Qiguang Chen}, \bibinfo{person}{Yao Mu}, \bibinfo{person}{Wenqi Shao}, {and} \bibinfo{person}{Ping Luo}.} \bibinfo{year}{2024}\natexlab{}.
\newblock \showarticletitle{Hiagent: Hierarchical working memory management for solving long-horizon agent tasks with large language model}.
\newblock \bibinfo{journal}{\emph{arXiv preprint arXiv:2408.09559}} (\bibinfo{year}{2024}).
\newblock


\bibitem[Huang et~al\mbox{.}(2025)]%
        {huang2025survey}
\bibfield{author}{\bibinfo{person}{Lei Huang}, \bibinfo{person}{Weijiang Yu}, \bibinfo{person}{Weitao Ma}, \bibinfo{person}{Weihong Zhong}, \bibinfo{person}{Zhangyin Feng}, \bibinfo{person}{Haotian Wang}, \bibinfo{person}{Qianglong Chen}, \bibinfo{person}{Weihua Peng}, \bibinfo{person}{Xiaocheng Feng}, \bibinfo{person}{Bing Qin}, {et~al\mbox{.}}} \bibinfo{year}{2025}\natexlab{}.
\newblock \showarticletitle{A survey on hallucination in large language models: Principles, taxonomy, challenges, and open questions}.
\newblock \bibinfo{journal}{\emph{ACM Transactions on Information Systems}} \bibinfo{volume}{43}, \bibinfo{number}{2} (\bibinfo{year}{2025}), \bibinfo{pages}{1--55}.
\newblock


\bibitem[Huang et~al\mbox{.}(2022)]%
        {huang2022inner}
\bibfield{author}{\bibinfo{person}{Wenlong Huang}, \bibinfo{person}{Fei Xia}, \bibinfo{person}{Ted Xiao}, \bibinfo{person}{Harris Chan}, \bibinfo{person}{Jacky Liang}, \bibinfo{person}{Pete Florence}, \bibinfo{person}{Andy Zeng}, \bibinfo{person}{Jonathan Tompson}, \bibinfo{person}{Igor Mordatch}, \bibinfo{person}{Yevgen Chebotar}, {et~al\mbox{.}}} \bibinfo{year}{2022}\natexlab{}.
\newblock \showarticletitle{Inner monologue: Embodied reasoning through planning with language models}.
\newblock \bibinfo{journal}{\emph{arXiv preprint arXiv:2207.05608}} (\bibinfo{year}{2022}).
\newblock


\bibitem[Huang et~al\mbox{.}(2024a)]%
        {huang2024understanding}
\bibfield{author}{\bibinfo{person}{Xu Huang}, \bibinfo{person}{Weiwen Liu}, \bibinfo{person}{Xiaolong Chen}, \bibinfo{person}{Xingmei Wang}, \bibinfo{person}{Hao Wang}, \bibinfo{person}{Defu Lian}, \bibinfo{person}{Yasheng Wang}, \bibinfo{person}{Ruiming Tang}, {and} \bibinfo{person}{Enhong Chen}.} \bibinfo{year}{2024}\natexlab{a}.
\newblock \showarticletitle{Understanding the planning of LLM agents: A survey}.
\newblock \bibinfo{journal}{\emph{arXiv preprint arXiv:2402.02716}} (\bibinfo{year}{2024}).
\newblock


\bibitem[Huang et~al\mbox{.}(2024b)]%
        {huang2024drivlme}
\bibfield{author}{\bibinfo{person}{Yidong Huang}, \bibinfo{person}{Jacob Sansom}, \bibinfo{person}{Ziqiao Ma}, \bibinfo{person}{Felix Gervits}, {and} \bibinfo{person}{Joyce Chai}.} \bibinfo{year}{2024}\natexlab{b}.
\newblock \showarticletitle{Drivlme: Enhancing llm-based autonomous driving agents with embodied and social experiences}. In \bibinfo{booktitle}{\emph{2024 IEEE/RSJ International Conference on Intelligent Robots and Systems (IROS)}}. IEEE, \bibinfo{pages}{3153--3160}.
\newblock


\bibitem[Imran et~al\mbox{.}(2024)]%
        {imran2024llasa}
\bibfield{author}{\bibinfo{person}{Sheikh~Asif Imran}, \bibinfo{person}{Mohammad Nur~Hossain Khan}, \bibinfo{person}{Subrata Biswas}, {and} \bibinfo{person}{Bashima Islam}.} \bibinfo{year}{2024}\natexlab{}.
\newblock \showarticletitle{LLaSA: A Multimodal LLM for Human Activity Analysis Through Wearable and Smartphone Sensors}.
\newblock \bibinfo{journal}{\emph{arXiv preprint arXiv:2406.14498}} (\bibinfo{year}{2024}).
\newblock


\bibitem[Jaech et~al\mbox{.}(2024)]%
        {jaech2024openai}
\bibfield{author}{\bibinfo{person}{Aaron Jaech}, \bibinfo{person}{Adam Kalai}, \bibinfo{person}{Adam Lerer}, \bibinfo{person}{Adam Richardson}, \bibinfo{person}{Ahmed El-Kishky}, \bibinfo{person}{Aiden Low}, \bibinfo{person}{Alec Helyar}, \bibinfo{person}{Aleksander Madry}, \bibinfo{person}{Alex Beutel}, \bibinfo{person}{Alex Carney}, {et~al\mbox{.}}} \bibinfo{year}{2024}\natexlab{}.
\newblock \showarticletitle{Openai o1 system card}.
\newblock \bibinfo{journal}{\emph{arXiv preprint arXiv:2412.16720}} (\bibinfo{year}{2024}).
\newblock


\bibitem[Javaid et~al\mbox{.}(2022)]%
        {javaid2022significance}
\bibfield{author}{\bibinfo{person}{Mohd Javaid}, \bibinfo{person}{Abid Haleem}, \bibinfo{person}{Ravi~Pratap Singh}, \bibinfo{person}{Rajiv Suman}, {and} \bibinfo{person}{Shanay Rab}.} \bibinfo{year}{2022}\natexlab{}.
\newblock \showarticletitle{Significance of machine learning in healthcare: Features, pillars and applications}.
\newblock \bibinfo{journal}{\emph{International Journal of Intelligent Networks}}  \bibinfo{volume}{3} (\bibinfo{year}{2022}), \bibinfo{pages}{58--73}.
\newblock


\bibitem[Ji et~al\mbox{.}(2024)]%
        {ji2024hargpt}
\bibfield{author}{\bibinfo{person}{Sijie Ji}, \bibinfo{person}{Xinzhe Zheng}, {and} \bibinfo{person}{Chenshu Wu}.} \bibinfo{year}{2024}\natexlab{}.
\newblock \showarticletitle{Hargpt: Are llms zero-shot human activity recognizers?}. In \bibinfo{booktitle}{\emph{2024 IEEE International Workshop on Foundation Models for Cyber-Physical Systems \& Internet of Things (FMSys)}}. IEEE, \bibinfo{pages}{38--43}.
\newblock


\bibitem[Kara et~al\mbox{.}(2024)]%
        {kara2024phymask}
\bibfield{author}{\bibinfo{person}{Denizhan Kara}, \bibinfo{person}{Tomoyoshi Kimura}, \bibinfo{person}{Yatong Chen}, \bibinfo{person}{Jinyang Li}, \bibinfo{person}{Ruijie Wang}, \bibinfo{person}{Yizhuo Chen}, \bibinfo{person}{Tianshi Wang}, \bibinfo{person}{Shengzhong Liu}, {and} \bibinfo{person}{Tarek Abdelzaher}.} \bibinfo{year}{2024}\natexlab{}.
\newblock \showarticletitle{PhyMask: An Adaptive Masking Paradigm for Efficient Self-Supervised Learning in IoT}. In \bibinfo{booktitle}{\emph{Proceedings of the 22nd ACM Conference on Embedded Networked Sensor Systems}}. \bibinfo{pages}{97--111}.
\newblock


\bibitem[Karar et~al\mbox{.}(2022)]%
        {karar2022survey}
\bibfield{author}{\bibinfo{person}{Mohamed~Esmail Karar}, \bibinfo{person}{Hazem~Ibrahim Shehata}, {and} \bibinfo{person}{Omar Reyad}.} \bibinfo{year}{2022}\natexlab{}.
\newblock \showarticletitle{A survey of IoT-based fall detection for aiding elderly care: Sensors, methods, challenges and future trends}.
\newblock \bibinfo{journal}{\emph{Applied Sciences}} \bibinfo{volume}{12}, \bibinfo{number}{7} (\bibinfo{year}{2022}), \bibinfo{pages}{3276}.
\newblock


\bibitem[Karine and Marlin(2025)]%
        {karine2025using}
\bibfield{author}{\bibinfo{person}{Karine Karine} {and} \bibinfo{person}{Benjamin Marlin}.} \bibinfo{year}{2025}\natexlab{}.
\newblock \showarticletitle{Using LLMs to improve RL policies in personalized health adaptive interventions}. In \bibinfo{booktitle}{\emph{Proceedings of the Second Workshop on Patient-Oriented Language Processing (CL4Health)}}. \bibinfo{pages}{137--147}.
\newblock


\bibitem[Kaufmann et~al\mbox{.}(2023)]%
        {kaufmann2023survey}
\bibfield{author}{\bibinfo{person}{Timo Kaufmann}, \bibinfo{person}{Paul Weng}, \bibinfo{person}{Viktor Bengs}, {and} \bibinfo{person}{Eyke H{\"u}llermeier}.} \bibinfo{year}{2023}\natexlab{}.
\newblock \showarticletitle{A survey of reinforcement learning from human feedback}.
\newblock \bibinfo{journal}{\emph{arXiv preprint arXiv:2312.14925}}  \bibinfo{volume}{10} (\bibinfo{year}{2023}).
\newblock


\bibitem[Khatiwada et~al\mbox{.}(2025)]%
        {khatiwada2025large}
\bibfield{author}{\bibinfo{person}{Kushal Khatiwada}, \bibinfo{person}{Jayden Hopper}, \bibinfo{person}{Joseph Cheatham}, \bibinfo{person}{Ayan Joshi}, {and} \bibinfo{person}{Sabur Baidya}.} \bibinfo{year}{2025}\natexlab{}.
\newblock \showarticletitle{Large Language Models in the IoT Ecosystem--A Survey on Security Challenges and Applications}.
\newblock \bibinfo{journal}{\emph{arXiv preprint arXiv:2505.17586}} (\bibinfo{year}{2025}).
\newblock


\bibitem[Kim et~al\mbox{.}(2025)]%
        {kim2025medical}
\bibfield{author}{\bibinfo{person}{Yubin Kim}, \bibinfo{person}{Hyewon Jeong}, \bibinfo{person}{Shan Chen}, \bibinfo{person}{Shuyue~Stella Li}, \bibinfo{person}{Mingyu Lu}, \bibinfo{person}{Kumail Alhamoud}, \bibinfo{person}{Jimin Mun}, \bibinfo{person}{Cristina Grau}, \bibinfo{person}{Minseok Jung}, \bibinfo{person}{Rodrigo Gameiro}, {et~al\mbox{.}}} \bibinfo{year}{2025}\natexlab{}.
\newblock \showarticletitle{Medical hallucinations in foundation models and their impact on healthcare}.
\newblock \bibinfo{journal}{\emph{arXiv preprint arXiv:2503.05777}} (\bibinfo{year}{2025}).
\newblock


\bibitem[Kimura et~al\mbox{.}(2024a)]%
        {kimura2024vibrofm}
\bibfield{author}{\bibinfo{person}{Tomoyoshi Kimura}, \bibinfo{person}{Jinyang Li}, \bibinfo{person}{Tianshi Wang}, \bibinfo{person}{Yizhuo Chen}, \bibinfo{person}{Ruijie Wang}, \bibinfo{person}{Denizhan Kara}, \bibinfo{person}{Maggie Wigness}, \bibinfo{person}{Joydeep Bhattacharyya}, \bibinfo{person}{Mudhakar Srivatsa}, \bibinfo{person}{Shengzhong Liu}, {et~al\mbox{.}}} \bibinfo{year}{2024}\natexlab{a}.
\newblock \showarticletitle{Vibrofm: Towards micro foundation models for robust multimodal iot sensing}. In \bibinfo{booktitle}{\emph{2024 IEEE 21st International Conference on Mobile Ad-Hoc and Smart Systems (MASS)}}. IEEE, \bibinfo{pages}{10--18}.
\newblock


\bibitem[Kimura et~al\mbox{.}(2024b)]%
        {kimura2024efficiency}
\bibfield{author}{\bibinfo{person}{Tomoyoshi Kimura}, \bibinfo{person}{Jinyang Li}, \bibinfo{person}{Tianshi Wang}, \bibinfo{person}{Denizhan Kara}, \bibinfo{person}{Yizhuo Chen}, \bibinfo{person}{Yigong Hu}, \bibinfo{person}{Ruijie Wang}, \bibinfo{person}{Maggie Wigness}, \bibinfo{person}{Shengzhong Liu}, \bibinfo{person}{Mani Srivastava}, {et~al\mbox{.}}} \bibinfo{year}{2024}\natexlab{b}.
\newblock \showarticletitle{On the efficiency and robustness of vibration-based foundation models for iot sensing: A case study}. In \bibinfo{booktitle}{\emph{2024 IEEE International Workshop on Foundation Models for Cyber-Physical Systems \& Internet of Things (FMSys)}}. IEEE, \bibinfo{pages}{7--12}.
\newblock


\bibitem[K{\"o}k et~al\mbox{.}(2024)]%
        {kok2024iot}
\bibfield{author}{\bibinfo{person}{{\.I}brahim K{\"o}k}, \bibinfo{person}{Orhan Demirci}, {and} \bibinfo{person}{Suat {\"O}zdemir}.} \bibinfo{year}{2024}\natexlab{}.
\newblock \showarticletitle{When IoT Meet LLMs: Applications and Challenges}. In \bibinfo{booktitle}{\emph{2024 IEEE International Conference on Big Data (BigData)}}. IEEE, \bibinfo{pages}{7075--7084}.
\newblock


\bibitem[Kumar(2021)]%
        {kumar2021semantic}
\bibfield{author}{\bibinfo{person}{Abhilasha~A Kumar}.} \bibinfo{year}{2021}\natexlab{}.
\newblock \showarticletitle{Semantic memory: A review of methods, models, and current challenges}.
\newblock \bibinfo{journal}{\emph{Psychonomic bulletin \& review}} \bibinfo{volume}{28}, \bibinfo{number}{1} (\bibinfo{year}{2021}), \bibinfo{pages}{40--80}.
\newblock


\bibitem[Lewis et~al\mbox{.}(2020)]%
        {lewis2020retrieval}
\bibfield{author}{\bibinfo{person}{Patrick Lewis}, \bibinfo{person}{Ethan Perez}, \bibinfo{person}{Aleksandra Piktus}, \bibinfo{person}{Fabio Petroni}, \bibinfo{person}{Vladimir Karpukhin}, \bibinfo{person}{Naman Goyal}, \bibinfo{person}{Heinrich K{\"u}ttler}, \bibinfo{person}{Mike Lewis}, \bibinfo{person}{Wen-tau Yih}, \bibinfo{person}{Tim Rockt{\"a}schel}, {et~al\mbox{.}}} \bibinfo{year}{2020}\natexlab{}.
\newblock \showarticletitle{Retrieval-augmented generation for knowledge-intensive nlp tasks}.
\newblock \bibinfo{journal}{\emph{Advances in neural information processing systems}}  \bibinfo{volume}{33} (\bibinfo{year}{2020}), \bibinfo{pages}{9459--9474}.
\newblock


\bibitem[Li et~al\mbox{.}(2024c)]%
        {li2024llmcount}
\bibfield{author}{\bibinfo{person}{Boyan Li}, \bibinfo{person}{Shengyi Ding}, \bibinfo{person}{Deen Ma}, \bibinfo{person}{Yixuan Wu}, \bibinfo{person}{Hongjie Liao}, {and} \bibinfo{person}{Kaiyuan Hu}.} \bibinfo{year}{2024}\natexlab{c}.
\newblock \showarticletitle{LLMCount: Enhancing Stationary mmWave Detection with Multimodal-LLM}.
\newblock \bibinfo{journal}{\emph{arXiv preprint arXiv:2409.16209}} (\bibinfo{year}{2024}).
\newblock


\bibitem[Li et~al\mbox{.}(2024d)]%
        {li2024generation}
\bibfield{author}{\bibinfo{person}{Dawei Li}, \bibinfo{person}{Bohan Jiang}, \bibinfo{person}{Liangjie Huang}, \bibinfo{person}{Alimohammad Beigi}, \bibinfo{person}{Chengshuai Zhao}, \bibinfo{person}{Zhen Tan}, \bibinfo{person}{Amrita Bhattacharjee}, \bibinfo{person}{Yuxuan Jiang}, \bibinfo{person}{Canyu Chen}, \bibinfo{person}{Tianhao Wu}, {et~al\mbox{.}}} \bibinfo{year}{2024}\natexlab{d}.
\newblock \showarticletitle{From generation to judgment: Opportunities and challenges of llm-as-a-judge}.
\newblock \bibinfo{journal}{\emph{arXiv preprint arXiv:2411.16594}} (\bibinfo{year}{2024}).
\newblock


\bibitem[Li et~al\mbox{.}(2024a)]%
        {li2024lasp}
\bibfield{author}{\bibinfo{person}{Haoming Li}, \bibinfo{person}{Zhaoliang Chen}, \bibinfo{person}{Jonathan Zhang}, {and} \bibinfo{person}{Fei Liu}.} \bibinfo{year}{2024}\natexlab{a}.
\newblock \showarticletitle{LASP: Surveying the State-of-the-Art in Large Language Model-Assisted AI Planning}.
\newblock \bibinfo{journal}{\emph{arXiv preprint arXiv:2409.01806}} (\bibinfo{year}{2024}).
\newblock


\bibitem[Li et~al\mbox{.}(2024e)]%
        {li2024survey}
\bibfield{author}{\bibinfo{person}{Xinyi Li}, \bibinfo{person}{Sai Wang}, \bibinfo{person}{Siqi Zeng}, \bibinfo{person}{Yu Wu}, {and} \bibinfo{person}{Yi Yang}.} \bibinfo{year}{2024}\natexlab{e}.
\newblock \showarticletitle{A survey on LLM-based multi-agent systems: workflow, infrastructure, and challenges}.
\newblock \bibinfo{journal}{\emph{Vicinagearth}} \bibinfo{volume}{1}, \bibinfo{number}{1} (\bibinfo{year}{2024}), \bibinfo{pages}{9}.
\newblock


\bibitem[Li et~al\mbox{.}(2024f)]%
        {li2024ids}
\bibfield{author}{\bibinfo{person}{Yanjie Li}, \bibinfo{person}{Zhen Xiang}, \bibinfo{person}{Nathaniel~D Bastian}, \bibinfo{person}{Dawn Song}, {and} \bibinfo{person}{Bo Li}.} \bibinfo{year}{2024}\natexlab{f}.
\newblock \showarticletitle{IDS-Agent: An LLM Agent for Explainable Intrusion Detection in IoT Networks}. In \bibinfo{booktitle}{\emph{NeurIPS 2024 Workshop on Open-World Agents}}.
\newblock


\bibitem[Li et~al\mbox{.}(2024b)]%
        {li2024sensorllm}
\bibfield{author}{\bibinfo{person}{Zechen Li}, \bibinfo{person}{Shohreh Deldari}, \bibinfo{person}{Linyao Chen}, \bibinfo{person}{Hao Xue}, {and} \bibinfo{person}{Flora~D Salim}.} \bibinfo{year}{2024}\natexlab{b}.
\newblock \showarticletitle{Sensorllm: Aligning large language models with motion sensors for human activity recognition}.
\newblock \bibinfo{journal}{\emph{arXiv preprint arXiv:2410.10624}} (\bibinfo{year}{2024}).
\newblock


\bibitem[Liakos et~al\mbox{.}(2018)]%
        {liakos2018machine}
\bibfield{author}{\bibinfo{person}{Konstantinos~G Liakos}, \bibinfo{person}{Patrizia Busato}, \bibinfo{person}{Dimitrios Moshou}, \bibinfo{person}{Simon Pearson}, {and} \bibinfo{person}{Dionysis Bochtis}.} \bibinfo{year}{2018}\natexlab{}.
\newblock \showarticletitle{Machine learning in agriculture: A review}.
\newblock \bibinfo{journal}{\emph{Sensors}} \bibinfo{volume}{18}, \bibinfo{number}{8} (\bibinfo{year}{2018}), \bibinfo{pages}{2674}.
\newblock


\bibitem[Liang and Tong(2025)]%
        {liang2025llm}
\bibfield{author}{\bibinfo{person}{Guannan Liang} {and} \bibinfo{person}{Qianqian Tong}.} \bibinfo{year}{2025}\natexlab{}.
\newblock \showarticletitle{LLM-Powered AI Agent Systems and Their Applications in Industry}.
\newblock \bibinfo{journal}{\emph{arXiv preprint arXiv:2505.16120}} (\bibinfo{year}{2025}).
\newblock


\bibitem[Lin(2004)]%
        {lin2004rouge}
\bibfield{author}{\bibinfo{person}{Chin-Yew Lin}.} \bibinfo{year}{2004}\natexlab{}.
\newblock \showarticletitle{Rouge: A package for automatic evaluation of summaries}. In \bibinfo{booktitle}{\emph{Text summarization branches out}}. \bibinfo{pages}{74--81}.
\newblock


\bibitem[Liu et~al\mbox{.}(2024a)]%
        {liu2024deepseek}
\bibfield{author}{\bibinfo{person}{Aixin Liu}, \bibinfo{person}{Bei Feng}, \bibinfo{person}{Bin Wang}, \bibinfo{person}{Bingxuan Wang}, \bibinfo{person}{Bo Liu}, \bibinfo{person}{Chenggang Zhao}, \bibinfo{person}{Chengqi Dengr}, \bibinfo{person}{Chong Ruan}, \bibinfo{person}{Damai Dai}, \bibinfo{person}{Daya Guo}, {et~al\mbox{.}}} \bibinfo{year}{2024}\natexlab{a}.
\newblock \showarticletitle{Deepseek-v2: A strong, economical, and efficient mixture-of-experts language model}.
\newblock \bibinfo{journal}{\emph{arXiv preprint arXiv:2405.04434}} (\bibinfo{year}{2024}).
\newblock


\bibitem[Liu et~al\mbox{.}(2025a)]%
        {liu2025wifo}
\bibfield{author}{\bibinfo{person}{Boxun Liu}, \bibinfo{person}{Shijian Gao}, \bibinfo{person}{Xuanyu Liu}, \bibinfo{person}{Xiang Cheng}, {and} \bibinfo{person}{Liuqing Yang}.} \bibinfo{year}{2025}\natexlab{a}.
\newblock \showarticletitle{WiFo: Wireless foundation model for channel prediction}.
\newblock \bibinfo{journal}{\emph{Science China Information Sciences}} \bibinfo{volume}{68}, \bibinfo{number}{6} (\bibinfo{year}{2025}), \bibinfo{pages}{1--13}.
\newblock


\bibitem[Liu et~al\mbox{.}(2024c)]%
        {liu2024llm4cp}
\bibfield{author}{\bibinfo{person}{Boxun Liu}, \bibinfo{person}{Xuanyu Liu}, \bibinfo{person}{Shijian Gao}, \bibinfo{person}{Xiang Cheng}, {and} \bibinfo{person}{Liuqing Yang}.} \bibinfo{year}{2024}\natexlab{c}.
\newblock \showarticletitle{LLM4CP: Adapting large language models for channel prediction}.
\newblock \bibinfo{journal}{\emph{Journal of Communications and Information Networks}} \bibinfo{volume}{9}, \bibinfo{number}{2} (\bibinfo{year}{2024}), \bibinfo{pages}{113--125}.
\newblock


\bibitem[Liu et~al\mbox{.}(2024b)]%
        {liu2024chainstream}
\bibfield{author}{\bibinfo{person}{Jiacheng Liu}, \bibinfo{person}{Yuanchun Li}, \bibinfo{person}{Liangyan Li}, \bibinfo{person}{Yi Sun}, \bibinfo{person}{Hao Wen}, \bibinfo{person}{Xiangyu Li}, \bibinfo{person}{Yao Guo}, {and} \bibinfo{person}{Yunxin Liu}.} \bibinfo{year}{2024}\natexlab{b}.
\newblock \showarticletitle{ChainStream: An LLM-based Framework for Unified Synthetic Sensing}.
\newblock \bibinfo{journal}{\emph{arXiv preprint arXiv:2412.15240}} (\bibinfo{year}{2024}).
\newblock


\bibitem[Liu et~al\mbox{.}(2024d)]%
        {liu2024tasking}
\bibfield{author}{\bibinfo{person}{Kaiwei Liu}, \bibinfo{person}{Bufang Yang}, \bibinfo{person}{Lilin Xu}, \bibinfo{person}{Yunqi Guo}, \bibinfo{person}{Neiwen Ling}, \bibinfo{person}{Zhihe Zhao}, \bibinfo{person}{Guoliang Xing}, \bibinfo{person}{Xian Shuai}, \bibinfo{person}{Xiaozhe Ren}, \bibinfo{person}{Xin Jiang}, {et~al\mbox{.}}} \bibinfo{year}{2024}\natexlab{d}.
\newblock \showarticletitle{Tasking Heterogeneous Sensor Systems with LLMs}. In \bibinfo{booktitle}{\emph{Proceedings of the 22nd ACM Conference on Embedded Networked Sensor Systems}}. \bibinfo{pages}{901--902}.
\newblock


\bibitem[Liu et~al\mbox{.}(2023a)]%
        {liu2023think}
\bibfield{author}{\bibinfo{person}{Lei Liu}, \bibinfo{person}{Xiaoyan Yang}, \bibinfo{person}{Yue Shen}, \bibinfo{person}{Binbin Hu}, \bibinfo{person}{Zhiqiang Zhang}, \bibinfo{person}{Jinjie Gu}, {and} \bibinfo{person}{Guannan Zhang}.} \bibinfo{year}{2023}\natexlab{a}.
\newblock \showarticletitle{Think-in-memory: Recalling and post-thinking enable llms with long-term memory}.
\newblock \bibinfo{journal}{\emph{arXiv preprint arXiv:2311.08719}} (\bibinfo{year}{2023}).
\newblock


\bibitem[Liu et~al\mbox{.}(2023b)]%
        {liu2023pre}
\bibfield{author}{\bibinfo{person}{Pengfei Liu}, \bibinfo{person}{Weizhe Yuan}, \bibinfo{person}{Jinlan Fu}, \bibinfo{person}{Zhengbao Jiang}, \bibinfo{person}{Hiroaki Hayashi}, {and} \bibinfo{person}{Graham Neubig}.} \bibinfo{year}{2023}\natexlab{b}.
\newblock \showarticletitle{Pre-train, prompt, and predict: A systematic survey of prompting methods in natural language processing}.
\newblock \bibinfo{journal}{\emph{ACM computing surveys}} \bibinfo{volume}{55}, \bibinfo{number}{9} (\bibinfo{year}{2023}), \bibinfo{pages}{1--35}.
\newblock


\bibitem[Liu et~al\mbox{.}(2025b)]%
        {liu2025llm4wm}
\bibfield{author}{\bibinfo{person}{Xuanyu Liu}, \bibinfo{person}{Shijian Gao}, \bibinfo{person}{Boxun Liu}, \bibinfo{person}{Xiang Cheng}, {and} \bibinfo{person}{Liuqing Yang}.} \bibinfo{year}{2025}\natexlab{b}.
\newblock \showarticletitle{LLM4WM: Adapting LLM for Wireless Multi-Tasking}.
\newblock \bibinfo{journal}{\emph{arXiv preprint arXiv:2501.12983}} (\bibinfo{year}{2025}).
\newblock


\bibitem[Liu et~al\mbox{.}(2024e)]%
        {liu2024agents4plc}
\bibfield{author}{\bibinfo{person}{Zihan Liu}, \bibinfo{person}{Ruinan Zeng}, \bibinfo{person}{Dongxia Wang}, \bibinfo{person}{Gengyun Peng}, \bibinfo{person}{Jingyi Wang}, \bibinfo{person}{Qiang Liu}, \bibinfo{person}{Peiyu Liu}, {and} \bibinfo{person}{Wenhai Wang}.} \bibinfo{year}{2024}\natexlab{e}.
\newblock \showarticletitle{Agents4PLC: Automating Closed-loop PLC Code Generation and Verification in Industrial Control Systems using LLM-based Agents}.
\newblock \bibinfo{journal}{\emph{arXiv preprint arXiv:2410.14209}} (\bibinfo{year}{2024}).
\newblock


\bibitem[Luo et~al\mbox{.}(2025)]%
        {luo2025large}
\bibfield{author}{\bibinfo{person}{Junyu Luo}, \bibinfo{person}{Weizhi Zhang}, \bibinfo{person}{Ye Yuan}, \bibinfo{person}{Yusheng Zhao}, \bibinfo{person}{Junwei Yang}, \bibinfo{person}{Yiyang Gu}, \bibinfo{person}{Bohan Wu}, \bibinfo{person}{Binqi Chen}, \bibinfo{person}{Ziyue Qiao}, \bibinfo{person}{Qingqing Long}, {et~al\mbox{.}}} \bibinfo{year}{2025}\natexlab{}.
\newblock \showarticletitle{Large Language Model Agent: A Survey on Methodology, Applications and Challenges}.
\newblock \bibinfo{journal}{\emph{arXiv preprint arXiv:2503.21460}} (\bibinfo{year}{2025}).
\newblock


\bibitem[Ma et~al\mbox{.}(2025)]%
        {ma2025safety}
\bibfield{author}{\bibinfo{person}{Xingjun Ma}, \bibinfo{person}{Yifeng Gao}, \bibinfo{person}{Yixu Wang}, \bibinfo{person}{Ruofan Wang}, \bibinfo{person}{Xin Wang}, \bibinfo{person}{Ye Sun}, \bibinfo{person}{Yifan Ding}, \bibinfo{person}{Hengyuan Xu}, \bibinfo{person}{Yunhao Chen}, \bibinfo{person}{Yunhan Zhao}, {et~al\mbox{.}}} \bibinfo{year}{2025}\natexlab{}.
\newblock \showarticletitle{Safety at scale: A comprehensive survey of large model safety}.
\newblock \bibinfo{journal}{\emph{arXiv preprint arXiv:2502.05206}} (\bibinfo{year}{2025}).
\newblock


\bibitem[Mao et~al\mbox{.}(2023)]%
        {mao2023language}
\bibfield{author}{\bibinfo{person}{Jiageng Mao}, \bibinfo{person}{Junjie Ye}, \bibinfo{person}{Yuxi Qian}, \bibinfo{person}{Marco Pavone}, {and} \bibinfo{person}{Yue Wang}.} \bibinfo{year}{2023}\natexlab{}.
\newblock \showarticletitle{A language agent for autonomous driving}.
\newblock \bibinfo{journal}{\emph{arXiv preprint arXiv:2311.10813}} (\bibinfo{year}{2023}).
\newblock


\bibitem[Mehta et~al\mbox{.}(2022)]%
        {mehta2022machine}
\bibfield{author}{\bibinfo{person}{Sonam Mehta}, \bibinfo{person}{Bharat Bhushan}, {and} \bibinfo{person}{Raghvendra Kumar}.} \bibinfo{year}{2022}\natexlab{}.
\newblock \showarticletitle{Machine learning approaches for smart city applications: Emergence, challenges and opportunities}.
\newblock \bibinfo{journal}{\emph{Recent advances in internet of things and machine learning: Real-world applications}} (\bibinfo{year}{2022}), \bibinfo{pages}{147--163}.
\newblock


\bibitem[Mo et~al\mbox{.}(2024)]%
        {mo2024iot}
\bibfield{author}{\bibinfo{person}{Shentong Mo}, \bibinfo{person}{Russ Salakhutdinov}, \bibinfo{person}{Louis-Philippe Morency}, {and} \bibinfo{person}{Paul~Pu Liang}.} \bibinfo{year}{2024}\natexlab{}.
\newblock \showarticletitle{Iot-lm: Large multisensory language models for the internet of things}.
\newblock \bibinfo{journal}{\emph{arXiv preprint arXiv:2407.09801}} (\bibinfo{year}{2024}).
\newblock


\bibitem[Mozaffari et~al\mbox{.}(2019)]%
        {mozaffari2019practical}
\bibfield{author}{\bibinfo{person}{Nassim Mozaffari}, \bibinfo{person}{Javad Rezazadeh}, \bibinfo{person}{Reza Farahbakhsh}, \bibinfo{person}{Samaneh Yazdani}, {and} \bibinfo{person}{Kumbesan Sandrasegaran}.} \bibinfo{year}{2019}\natexlab{}.
\newblock \showarticletitle{Practical fall detection based on IoT technologies: A survey}.
\newblock \bibinfo{journal}{\emph{Internet of things}}  \bibinfo{volume}{8} (\bibinfo{year}{2019}), \bibinfo{pages}{100124}.
\newblock


\bibitem[Nahum-Shani et~al\mbox{.}(2016)]%
        {nahum2016just}
\bibfield{author}{\bibinfo{person}{Inbal Nahum-Shani}, \bibinfo{person}{Shawna~N Smith}, \bibinfo{person}{Bonnie~J Spring}, \bibinfo{person}{Linda~M Collins}, \bibinfo{person}{Katie Witkiewitz}, \bibinfo{person}{Ambuj Tewari}, {and} \bibinfo{person}{Susan~A Murphy}.} \bibinfo{year}{2016}\natexlab{}.
\newblock \showarticletitle{Just-in-time adaptive interventions (JITAIs) in mobile health: key components and design principles for ongoing health behavior support}.
\newblock \bibinfo{journal}{\emph{Annals of behavioral medicine}} (\bibinfo{year}{2016}), \bibinfo{pages}{1--17}.
\newblock


\bibitem[Nepal et~al\mbox{.}(2024)]%
        {nepal2024mindscape}
\bibfield{author}{\bibinfo{person}{Subigya Nepal}, \bibinfo{person}{Arvind Pillai}, \bibinfo{person}{William Campbell}, \bibinfo{person}{Talie Massachi}, \bibinfo{person}{Michael~V Heinz}, \bibinfo{person}{Ashmita Kunwar}, \bibinfo{person}{Eunsol~Soul Choi}, \bibinfo{person}{Xuhai Xu}, \bibinfo{person}{Joanna Kuc}, \bibinfo{person}{Jeremy~F Huckins}, {et~al\mbox{.}}} \bibinfo{year}{2024}\natexlab{}.
\newblock \showarticletitle{MindScape Study: Integrating LLM and Behavioral Sensing for Personalized AI-Driven Journaling Experiences}.
\newblock \bibinfo{journal}{\emph{Proceedings of the ACM on interactive, mobile, wearable and ubiquitous technologies}} \bibinfo{volume}{8}, \bibinfo{number}{4} (\bibinfo{year}{2024}), \bibinfo{pages}{1--44}.
\newblock


\bibitem[Neven et~al\mbox{.}(2017)]%
        {neven2017fast}
\bibfield{author}{\bibinfo{person}{Davy Neven}, \bibinfo{person}{Bert De~Brabandere}, \bibinfo{person}{Stamatios Georgoulis}, \bibinfo{person}{Marc Proesmans}, {and} \bibinfo{person}{Luc Van~Gool}.} \bibinfo{year}{2017}\natexlab{}.
\newblock \showarticletitle{Fast scene understanding for autonomous driving}.
\newblock \bibinfo{journal}{\emph{arXiv preprint arXiv:1708.02550}} (\bibinfo{year}{2017}).
\newblock


\bibitem[Nie et~al\mbox{.}(2024)]%
        {nie2024llm}
\bibfield{author}{\bibinfo{person}{Jingping Nie}, \bibinfo{person}{Hanya Shao}, \bibinfo{person}{Yuang Fan}, \bibinfo{person}{Qijia Shao}, \bibinfo{person}{Haoxuan You}, \bibinfo{person}{Matthias Preindl}, {and} \bibinfo{person}{Xiaofan Jiang}.} \bibinfo{year}{2024}\natexlab{}.
\newblock \showarticletitle{LLM-based conversational AI therapist for daily functioning screening and psychotherapeutic intervention via everyday smart devices}.
\newblock \bibinfo{journal}{\emph{arXiv preprint arXiv:2403.10779}} (\bibinfo{year}{2024}).
\newblock


\bibitem[Nouri et~al\mbox{.}(2024)]%
        {nouri2024engineering}
\bibfield{author}{\bibinfo{person}{Ali Nouri}, \bibinfo{person}{Beatriz Cabrero-Daniel}, \bibinfo{person}{Fredrik T{\"o}rner}, \bibinfo{person}{H{\aa}kan Sivencrona}, {and} \bibinfo{person}{Christian Berger}.} \bibinfo{year}{2024}\natexlab{}.
\newblock \showarticletitle{Engineering safety requirements for autonomous driving with large language models}. In \bibinfo{booktitle}{\emph{2024 IEEE 32nd International Requirements Engineering Conference (RE)}}. IEEE, \bibinfo{pages}{218--228}.
\newblock


\bibitem[Nuxoll and Laird(2007)]%
        {nuxoll2007extending}
\bibfield{author}{\bibinfo{person}{Andrew~M Nuxoll} {and} \bibinfo{person}{John~E Laird}.} \bibinfo{year}{2007}\natexlab{}.
\newblock \showarticletitle{Extending cognitive architecture with episodic memory}. In \bibinfo{booktitle}{\emph{AAAI}}. \bibinfo{pages}{1560--1564}.
\newblock


\bibitem[OpenAI(2025)]%
        {openai2025o3}
\bibfield{author}{\bibinfo{person}{OpenAI}.} \bibinfo{year}{2025}\natexlab{}.
\newblock \showarticletitle{OpenAI o3 and o4-mini System Card}.
\newblock \bibinfo{journal}{\emph{https://cdn.openai.com/pdf/2221c875-02dc-4789-800b-e7758f3722c1/o3-and-o4-mini-system-card.pdf}} (\bibinfo{year}{2025}).
\newblock


\bibitem[Ouyang et~al\mbox{.}(2022)]%
        {ouyang2022training}
\bibfield{author}{\bibinfo{person}{Long Ouyang}, \bibinfo{person}{Jeffrey Wu}, \bibinfo{person}{Xu Jiang}, \bibinfo{person}{Diogo Almeida}, \bibinfo{person}{Carroll Wainwright}, \bibinfo{person}{Pamela Mishkin}, \bibinfo{person}{Chong Zhang}, \bibinfo{person}{Sandhini Agarwal}, \bibinfo{person}{Katarina Slama}, \bibinfo{person}{Alex Ray}, {et~al\mbox{.}}} \bibinfo{year}{2022}\natexlab{}.
\newblock \showarticletitle{Training language models to follow instructions with human feedback}.
\newblock \bibinfo{journal}{\emph{Advances in neural information processing systems}}  \bibinfo{volume}{35} (\bibinfo{year}{2022}), \bibinfo{pages}{27730--27744}.
\newblock


\bibitem[Ouyang and Srivastava(2024)]%
        {ouyang2024llmsense}
\bibfield{author}{\bibinfo{person}{Xiaomin Ouyang} {and} \bibinfo{person}{Mani Srivastava}.} \bibinfo{year}{2024}\natexlab{}.
\newblock \showarticletitle{LLMSense: Harnessing LLMs for high-level reasoning over spatiotemporal sensor traces}. In \bibinfo{booktitle}{\emph{2024 IEEE 3rd Workshop on Machine Learning on Edge in Sensor Systems (SenSys-ML)}}. IEEE, \bibinfo{pages}{9--14}.
\newblock


\bibitem[Ouyang et~al\mbox{.}(2025)]%
        {ouyang2025mmbind}
\bibfield{author}{\bibinfo{person}{Xiaomin Ouyang}, \bibinfo{person}{Jason Wu}, \bibinfo{person}{Tomoyoshi Kimura}, \bibinfo{person}{Yihan Lin}, \bibinfo{person}{Gunjan Verma}, \bibinfo{person}{Tarek Abdelzaher}, {and} \bibinfo{person}{Mani Srivastava}.} \bibinfo{year}{2025}\natexlab{}.
\newblock \showarticletitle{MMBind: Unleashing the Potential of Distributed and Heterogeneous Data for Multimodal Learning in IoT}. In \bibinfo{booktitle}{\emph{Proceedings of the 23rd ACM Conference on Embedded Networked Sensor Systems}}. \bibinfo{pages}{491--503}.
\newblock


\bibitem[Papineni et~al\mbox{.}(2002)]%
        {papineni2002bleu}
\bibfield{author}{\bibinfo{person}{Kishore Papineni}, \bibinfo{person}{Salim Roukos}, \bibinfo{person}{Todd Ward}, {and} \bibinfo{person}{Wei-Jing Zhu}.} \bibinfo{year}{2002}\natexlab{}.
\newblock \showarticletitle{Bleu: a method for automatic evaluation of machine translation}. In \bibinfo{booktitle}{\emph{Proceedings of the 40th annual meeting of the Association for Computational Linguistics}}. \bibinfo{pages}{311--318}.
\newblock


\bibitem[Pereira et~al\mbox{.}(2020)]%
        {pereira2020challenges}
\bibfield{author}{\bibinfo{person}{Felisberto Pereira}, \bibinfo{person}{Ricardo Correia}, \bibinfo{person}{Pedro Pinho}, \bibinfo{person}{S{\'e}rgio~I Lopes}, {and} \bibinfo{person}{Nuno~Borges Carvalho}.} \bibinfo{year}{2020}\natexlab{}.
\newblock \showarticletitle{Challenges in resource-constrained IoT devices: Energy and communication as critical success factors for future IoT deployment}.
\newblock \bibinfo{journal}{\emph{Sensors}} \bibinfo{volume}{20}, \bibinfo{number}{22} (\bibinfo{year}{2020}), \bibinfo{pages}{6420}.
\newblock


\bibitem[Perera et~al\mbox{.}(2013)]%
        {perera2013context}
\bibfield{author}{\bibinfo{person}{Charith Perera}, \bibinfo{person}{Arkady Zaslavsky}, \bibinfo{person}{Peter Christen}, {and} \bibinfo{person}{Dimitrios Georgakopoulos}.} \bibinfo{year}{2013}\natexlab{}.
\newblock \showarticletitle{Context aware computing for the internet of things: A survey}.
\newblock \bibinfo{journal}{\emph{IEEE communications surveys \& tutorials}} \bibinfo{volume}{16}, \bibinfo{number}{1} (\bibinfo{year}{2013}), \bibinfo{pages}{414--454}.
\newblock


\bibitem[Plaat et~al\mbox{.}(2024)]%
        {plaat2024reasoning}
\bibfield{author}{\bibinfo{person}{Aske Plaat}, \bibinfo{person}{Annie Wong}, \bibinfo{person}{Suzan Verberne}, \bibinfo{person}{Joost Broekens}, \bibinfo{person}{Niki van Stein}, {and} \bibinfo{person}{Thomas Back}.} \bibinfo{year}{2024}\natexlab{}.
\newblock \showarticletitle{Reasoning with large language models, a survey}.
\newblock \bibinfo{journal}{\emph{arXiv preprint arXiv:2407.11511}} (\bibinfo{year}{2024}).
\newblock


\bibitem[Prasad et~al\mbox{.}(2023)]%
        {prasad2023adapt}
\bibfield{author}{\bibinfo{person}{Archiki Prasad}, \bibinfo{person}{Alexander Koller}, \bibinfo{person}{Mareike Hartmann}, \bibinfo{person}{Peter Clark}, \bibinfo{person}{Ashish Sabharwal}, \bibinfo{person}{Mohit Bansal}, {and} \bibinfo{person}{Tushar Khot}.} \bibinfo{year}{2023}\natexlab{}.
\newblock \showarticletitle{Adapt: As-needed decomposition and planning with language models}.
\newblock \bibinfo{journal}{\emph{arXiv preprint arXiv:2311.05772}} (\bibinfo{year}{2023}).
\newblock


\bibitem[Qin et~al\mbox{.}(2022)]%
        {qin2022cosformer}
\bibfield{author}{\bibinfo{person}{Zhen Qin}, \bibinfo{person}{Weixuan Sun}, \bibinfo{person}{Hui Deng}, \bibinfo{person}{Dongxu Li}, \bibinfo{person}{Yunshen Wei}, \bibinfo{person}{Baohong Lv}, \bibinfo{person}{Junjie Yan}, \bibinfo{person}{Lingpeng Kong}, {and} \bibinfo{person}{Yiran Zhong}.} \bibinfo{year}{2022}\natexlab{}.
\newblock \showarticletitle{cosformer: Rethinking softmax in attention}.
\newblock \bibinfo{journal}{\emph{arXiv preprint arXiv:2202.08791}} (\bibinfo{year}{2022}).
\newblock


\bibitem[Qin et~al\mbox{.}(2024a)]%
        {qin2024lightning}
\bibfield{author}{\bibinfo{person}{Zhen Qin}, \bibinfo{person}{Weigao Sun}, \bibinfo{person}{Dong Li}, \bibinfo{person}{Xuyang Shen}, \bibinfo{person}{Weixuan Sun}, {and} \bibinfo{person}{Yiran Zhong}.} \bibinfo{year}{2024}\natexlab{a}.
\newblock \showarticletitle{Lightning attention-2: A free lunch for handling unlimited sequence lengths in large language models}.
\newblock \bibinfo{journal}{\emph{arXiv preprint arXiv:2401.04658}} (\bibinfo{year}{2024}).
\newblock


\bibitem[Qin et~al\mbox{.}(2024b)]%
        {qin2024hgrn2}
\bibfield{author}{\bibinfo{person}{Zhen Qin}, \bibinfo{person}{Songlin Yang}, \bibinfo{person}{Weixuan Sun}, \bibinfo{person}{Xuyang Shen}, \bibinfo{person}{Dong Li}, \bibinfo{person}{Weigao Sun}, {and} \bibinfo{person}{Yiran Zhong}.} \bibinfo{year}{2024}\natexlab{b}.
\newblock \showarticletitle{Hgrn2: Gated linear rnns with state expansion}.
\newblock \bibinfo{journal}{\emph{arXiv preprint arXiv:2404.07904}} (\bibinfo{year}{2024}).
\newblock


\bibitem[Qu et~al\mbox{.}(2025)]%
        {qu2025tool}
\bibfield{author}{\bibinfo{person}{Changle Qu}, \bibinfo{person}{Sunhao Dai}, \bibinfo{person}{Xiaochi Wei}, \bibinfo{person}{Hengyi Cai}, \bibinfo{person}{Shuaiqiang Wang}, \bibinfo{person}{Dawei Yin}, \bibinfo{person}{Jun Xu}, {and} \bibinfo{person}{Ji-Rong Wen}.} \bibinfo{year}{2025}\natexlab{}.
\newblock \showarticletitle{Tool learning with large language models: A survey}.
\newblock \bibinfo{journal}{\emph{Frontiers of Computer Science}} \bibinfo{volume}{19}, \bibinfo{number}{8} (\bibinfo{year}{2025}), \bibinfo{pages}{198343}.
\newblock


\bibitem[Quartey et~al\mbox{.}(2024)]%
        {quartey2024verifiably}
\bibfield{author}{\bibinfo{person}{Benedict Quartey}, \bibinfo{person}{Eric Rosen}, \bibinfo{person}{Stefanie Tellex}, {and} \bibinfo{person}{George Konidaris}.} \bibinfo{year}{2024}\natexlab{}.
\newblock \showarticletitle{Verifiably following complex robot instructions with foundation models}.
\newblock \bibinfo{journal}{\emph{arXiv preprint arXiv:2402.11498}} (\bibinfo{year}{2024}).
\newblock


\bibitem[Rafailov et~al\mbox{.}(2023)]%
        {rafailov2023direct}
\bibfield{author}{\bibinfo{person}{Rafael Rafailov}, \bibinfo{person}{Archit Sharma}, \bibinfo{person}{Eric Mitchell}, \bibinfo{person}{Christopher~D Manning}, \bibinfo{person}{Stefano Ermon}, {and} \bibinfo{person}{Chelsea Finn}.} \bibinfo{year}{2023}\natexlab{}.
\newblock \showarticletitle{Direct preference optimization: Your language model is secretly a reward model}.
\newblock \bibinfo{journal}{\emph{Advances in Neural Information Processing Systems}}  \bibinfo{volume}{36} (\bibinfo{year}{2023}), \bibinfo{pages}{53728--53741}.
\newblock


\bibitem[Ravichandran et~al\mbox{.}(2025)]%
        {ravichandran2025safety}
\bibfield{author}{\bibinfo{person}{Zachary Ravichandran}, \bibinfo{person}{Alexander Robey}, \bibinfo{person}{Vijay Kumar}, \bibinfo{person}{George~J Pappas}, {and} \bibinfo{person}{Hamed Hassani}.} \bibinfo{year}{2025}\natexlab{}.
\newblock \showarticletitle{Safety Guardrails for LLM-Enabled Robots}.
\newblock \bibinfo{journal}{\emph{arXiv preprint arXiv:2503.07885}} (\bibinfo{year}{2025}).
\newblock


\bibitem[Rivkin et~al\mbox{.}(2024)]%
        {rivkin2024aiot}
\bibfield{author}{\bibinfo{person}{Dmitriy Rivkin}, \bibinfo{person}{Francois Hogan}, \bibinfo{person}{Amal Feriani}, \bibinfo{person}{Abhisek Konar}, \bibinfo{person}{Adam Sigal}, \bibinfo{person}{Xue Liu}, {and} \bibinfo{person}{Gregory Dudek}.} \bibinfo{year}{2024}\natexlab{}.
\newblock \showarticletitle{AIoT Smart Home via Autonomous LLM Agents}.
\newblock \bibinfo{journal}{\emph{IEEE Internet of Things Journal}} (\bibinfo{year}{2024}).
\newblock


\bibitem[Rong and Rutagemwa(2024)]%
        {rong2024leveraging}
\bibfield{author}{\bibinfo{person}{Bo Rong} {and} \bibinfo{person}{Humphrey Rutagemwa}.} \bibinfo{year}{2024}\natexlab{}.
\newblock \showarticletitle{Leveraging large language models for intelligent control of 6g integrated tn-ntn with iot service}.
\newblock \bibinfo{journal}{\emph{IEEE Network}} (\bibinfo{year}{2024}).
\newblock


\bibitem[Sahoo et~al\mbox{.}(2024)]%
        {sahoo2024systematic}
\bibfield{author}{\bibinfo{person}{Pranab Sahoo}, \bibinfo{person}{Ayush~Kumar Singh}, \bibinfo{person}{Sriparna Saha}, \bibinfo{person}{Vinija Jain}, \bibinfo{person}{Samrat Mondal}, {and} \bibinfo{person}{Aman Chadha}.} \bibinfo{year}{2024}\natexlab{}.
\newblock \showarticletitle{A systematic survey of prompt engineering in large language models: Techniques and applications}.
\newblock \bibinfo{journal}{\emph{arXiv preprint arXiv:2402.07927}} (\bibinfo{year}{2024}).
\newblock


\bibitem[Said(2023)]%
        {said2023bandwidth}
\bibfield{author}{\bibinfo{person}{Omar Said}.} \bibinfo{year}{2023}\natexlab{}.
\newblock \showarticletitle{A bandwidth control scheme for reducing the negative impact of bottlenecks in IoT environments: simulation and performance evaluation}.
\newblock \bibinfo{journal}{\emph{Internet of Things}}  \bibinfo{volume}{21} (\bibinfo{year}{2023}), \bibinfo{pages}{100682}.
\newblock


\bibitem[Saleh et~al\mbox{.}(2025)]%
        {saleh2025usercentrix}
\bibfield{author}{\bibinfo{person}{Alaa Saleh}, \bibinfo{person}{Sasu Tarkoma}, \bibinfo{person}{Praveen~Kumar Donta}, \bibinfo{person}{Naser~Hossein Motlagh}, \bibinfo{person}{Schahram Dustdar}, \bibinfo{person}{Susanna Pirttikangas}, {and} \bibinfo{person}{Lauri Lov{\'e}n}.} \bibinfo{year}{2025}\natexlab{}.
\newblock \showarticletitle{Usercentrix: An agentic memory-augmented ai framework for smart spaces}.
\newblock \bibinfo{journal}{\emph{arXiv preprint arXiv:2505.00472}} (\bibinfo{year}{2025}).
\newblock


\bibitem[Sanh et~al\mbox{.}(2019)]%
        {sanh2019distilbert}
\bibfield{author}{\bibinfo{person}{Victor Sanh}, \bibinfo{person}{Lysandre Debut}, \bibinfo{person}{Julien Chaumond}, {and} \bibinfo{person}{Thomas Wolf}.} \bibinfo{year}{2019}\natexlab{}.
\newblock \showarticletitle{DistilBERT, a distilled version of BERT: smaller, faster, cheaper and lighter}.
\newblock \bibinfo{journal}{\emph{arXiv preprint arXiv:1910.01108}} (\bibinfo{year}{2019}).
\newblock


\bibitem[Sarhaddi et~al\mbox{.}(2025)]%
        {sarhaddi2025llms}
\bibfield{author}{\bibinfo{person}{Fatemeh Sarhaddi}, \bibinfo{person}{Ngoc~Thi Nguyen}, \bibinfo{person}{Agustin Zuniga}, \bibinfo{person}{Pan Hui}, \bibinfo{person}{Sasu Tarkoma}, \bibinfo{person}{Huber Flores}, {and} \bibinfo{person}{Petteri Nurmi}.} \bibinfo{year}{2025}\natexlab{}.
\newblock \showarticletitle{Llms and iot: A comprehensive survey on large language models and the internet of things}.
\newblock \bibinfo{journal}{\emph{Authorea Preprints}} (\bibinfo{year}{2025}).
\newblock


\bibitem[Seng et~al\mbox{.}(2022)]%
        {seng2022artificial}
\bibfield{author}{\bibinfo{person}{Kah~Phooi Seng}, \bibinfo{person}{Li~Minn Ang}, {and} \bibinfo{person}{Ericmoore Ngharamike}.} \bibinfo{year}{2022}\natexlab{}.
\newblock \showarticletitle{Artificial intelligence Internet of Things: A new paradigm of distributed sensor networks}.
\newblock \bibinfo{journal}{\emph{International Journal of Distributed Sensor Networks}} \bibinfo{volume}{18}, \bibinfo{number}{3} (\bibinfo{year}{2022}), \bibinfo{pages}{15501477211062835}.
\newblock


\bibitem[Sennrich et~al\mbox{.}(2015)]%
        {sennrich2015neural}
\bibfield{author}{\bibinfo{person}{Rico Sennrich}, \bibinfo{person}{Barry Haddow}, {and} \bibinfo{person}{Alexandra Birch}.} \bibinfo{year}{2015}\natexlab{}.
\newblock \showarticletitle{Neural machine translation of rare words with subword units}.
\newblock \bibinfo{journal}{\emph{arXiv preprint arXiv:1508.07909}} (\bibinfo{year}{2015}).
\newblock


\bibitem[Shailaja et~al\mbox{.}(2018)]%
        {shailaja2018machine}
\bibfield{author}{\bibinfo{person}{K Shailaja}, \bibinfo{person}{Banoth Seetharamulu}, {and} \bibinfo{person}{MA Jabbar}.} \bibinfo{year}{2018}\natexlab{}.
\newblock \showarticletitle{Machine learning in healthcare: A review}. In \bibinfo{booktitle}{\emph{2018 Second international conference on electronics, communication and aerospace technology (ICECA)}}. IEEE, \bibinfo{pages}{910--914}.
\newblock


\bibitem[Sharma et~al\mbox{.}(2020)]%
        {sharma2020machine}
\bibfield{author}{\bibinfo{person}{Abhinav Sharma}, \bibinfo{person}{Arpit Jain}, \bibinfo{person}{Prateek Gupta}, {and} \bibinfo{person}{Vinay Chowdary}.} \bibinfo{year}{2020}\natexlab{}.
\newblock \showarticletitle{Machine learning applications for precision agriculture: A comprehensive review}.
\newblock \bibinfo{journal}{\emph{IEEe Access}}  \bibinfo{volume}{9} (\bibinfo{year}{2020}), \bibinfo{pages}{4843--4873}.
\newblock


\bibitem[Shazeer(2019)]%
        {shazeer2019fast}
\bibfield{author}{\bibinfo{person}{Noam Shazeer}.} \bibinfo{year}{2019}\natexlab{}.
\newblock \showarticletitle{Fast transformer decoding: One write-head is all you need}.
\newblock \bibinfo{journal}{\emph{arXiv preprint arXiv:1911.02150}} (\bibinfo{year}{2019}).
\newblock


\bibitem[She et~al\mbox{.}(2024)]%
        {she2024llmdiff}
\bibfield{author}{\bibinfo{person}{Lei She}, \bibinfo{person}{Chenghong Zhang}, \bibinfo{person}{Xin Man}, {and} \bibinfo{person}{Jie Shao}.} \bibinfo{year}{2024}\natexlab{}.
\newblock \showarticletitle{LLMDiff: Diffusion Model Using Frozen LLM Transformers for Precipitation Nowcasting}.
\newblock \bibinfo{journal}{\emph{Sensors}} \bibinfo{volume}{24}, \bibinfo{number}{18} (\bibinfo{year}{2024}), \bibinfo{pages}{6049}.
\newblock


\bibitem[Shen(2024)]%
        {shen2024llm}
\bibfield{author}{\bibinfo{person}{Zhuocheng Shen}.} \bibinfo{year}{2024}\natexlab{}.
\newblock \showarticletitle{Llm with tools: A survey}.
\newblock \bibinfo{journal}{\emph{arXiv preprint arXiv:2409.18807}} (\bibinfo{year}{2024}).
\newblock


\bibitem[Sheng et~al\mbox{.}(2025)]%
        {sheng2025beam}
\bibfield{author}{\bibinfo{person}{Yucheng Sheng}, \bibinfo{person}{Kai Huang}, \bibinfo{person}{Le Liang}, \bibinfo{person}{Peng Liu}, \bibinfo{person}{Shi Jin}, {and} \bibinfo{person}{Geoffrey~Ye Li}.} \bibinfo{year}{2025}\natexlab{}.
\newblock \showarticletitle{Beam prediction based on large language models}.
\newblock \bibinfo{journal}{\emph{IEEE Wireless Communications Letters}} (\bibinfo{year}{2025}).
\newblock


\bibitem[Shi et~al\mbox{.}(2024a)]%
        {shi2024large}
\bibfield{author}{\bibinfo{person}{Dan Shi}, \bibinfo{person}{Tianhao Shen}, \bibinfo{person}{Yufei Huang}, \bibinfo{person}{Zhigen Li}, \bibinfo{person}{Yongqi Leng}, \bibinfo{person}{Renren Jin}, \bibinfo{person}{Chuang Liu}, \bibinfo{person}{Xinwei Wu}, \bibinfo{person}{Zishan Guo}, \bibinfo{person}{Linhao Yu}, {et~al\mbox{.}}} \bibinfo{year}{2024}\natexlab{a}.
\newblock \showarticletitle{Large language model safety: A holistic survey}.
\newblock \bibinfo{journal}{\emph{arXiv preprint arXiv:2412.17686}} (\bibinfo{year}{2024}).
\newblock


\bibitem[Shi et~al\mbox{.}(2024b)]%
        {shi2024ehragent}
\bibfield{author}{\bibinfo{person}{Wenqi Shi}, \bibinfo{person}{Ran Xu}, \bibinfo{person}{Yuchen Zhuang}, \bibinfo{person}{Yue Yu}, \bibinfo{person}{Jieyu Zhang}, \bibinfo{person}{Hang Wu}, \bibinfo{person}{Yuanda Zhu}, \bibinfo{person}{Joyce Ho}, \bibinfo{person}{Carl Yang}, {and} \bibinfo{person}{May~D Wang}.} \bibinfo{year}{2024}\natexlab{b}.
\newblock \showarticletitle{Ehragent: Code empowers large language models for few-shot complex tabular reasoning on electronic health records}. In \bibinfo{booktitle}{\emph{Proceedings of the Conference on Empirical Methods in Natural Language Processing. Conference on Empirical Methods in Natural Language Processing}}, Vol.~\bibinfo{volume}{2024}. \bibinfo{pages}{22315}.
\newblock


\bibitem[Siam et~al\mbox{.}(2025)]%
        {siam2025artificial}
\bibfield{author}{\bibinfo{person}{Shakhrul~Iman Siam}, \bibinfo{person}{Hyunho Ahn}, \bibinfo{person}{Li Liu}, \bibinfo{person}{Samiul Alam}, \bibinfo{person}{Hui Shen}, \bibinfo{person}{Zhichao Cao}, \bibinfo{person}{Ness Shroff}, \bibinfo{person}{Bhaskar Krishnamachari}, \bibinfo{person}{Mani Srivastava}, {and} \bibinfo{person}{Mi Zhang}.} \bibinfo{year}{2025}\natexlab{}.
\newblock \showarticletitle{Artificial intelligence of things: A survey}.
\newblock \bibinfo{journal}{\emph{ACM Transactions on Sensor Networks}} \bibinfo{volume}{21}, \bibinfo{number}{1} (\bibinfo{year}{2025}), \bibinfo{pages}{1--75}.
\newblock


\bibitem[Singh et~al\mbox{.}(2022)]%
        {singh2022progprompt}
\bibfield{author}{\bibinfo{person}{Ishika Singh}, \bibinfo{person}{Valts Blukis}, \bibinfo{person}{Arsalan Mousavian}, \bibinfo{person}{Ankit Goyal}, \bibinfo{person}{Danfei Xu}, \bibinfo{person}{Jonathan Tremblay}, \bibinfo{person}{Dieter Fox}, \bibinfo{person}{Jesse Thomason}, {and} \bibinfo{person}{Animesh Garg}.} \bibinfo{year}{2022}\natexlab{}.
\newblock \showarticletitle{Progprompt: Generating situated robot task plans using large language models}.
\newblock \bibinfo{journal}{\emph{arXiv preprint arXiv:2209.11302}} (\bibinfo{year}{2022}).
\newblock


\bibitem[Song et~al\mbox{.}(2023b)]%
        {song2023llm}
\bibfield{author}{\bibinfo{person}{Chan~Hee Song}, \bibinfo{person}{Jiaman Wu}, \bibinfo{person}{Clayton Washington}, \bibinfo{person}{Brian~M Sadler}, \bibinfo{person}{Wei-Lun Chao}, {and} \bibinfo{person}{Yu Su}.} \bibinfo{year}{2023}\natexlab{b}.
\newblock \showarticletitle{Llm-planner: Few-shot grounded planning for embodied agents with large language models}. In \bibinfo{booktitle}{\emph{Proceedings of the IEEE/CVF international conference on computer vision}}. \bibinfo{pages}{2998--3009}.
\newblock


\bibitem[Song et~al\mbox{.}(2023c)]%
        {song2023pre}
\bibfield{author}{\bibinfo{person}{Lei Song}, \bibinfo{person}{Chuheng Zhang}, \bibinfo{person}{Li Zhao}, {and} \bibinfo{person}{Jiang Bian}.} \bibinfo{year}{2023}\natexlab{c}.
\newblock \showarticletitle{Pre-trained large language models for industrial control}.
\newblock \bibinfo{journal}{\emph{arXiv preprint arXiv:2308.03028}} (\bibinfo{year}{2023}).
\newblock


\bibitem[Song et~al\mbox{.}(2023a)]%
        {song2023comprehensive}
\bibfield{author}{\bibinfo{person}{Yisheng Song}, \bibinfo{person}{Ting Wang}, \bibinfo{person}{Puyu Cai}, \bibinfo{person}{Subrota~K Mondal}, {and} \bibinfo{person}{Jyoti~Prakash Sahoo}.} \bibinfo{year}{2023}\natexlab{a}.
\newblock \showarticletitle{A comprehensive survey of few-shot learning: Evolution, applications, challenges, and opportunities}.
\newblock \bibinfo{journal}{\emph{Comput. Surveys}} \bibinfo{volume}{55}, \bibinfo{number}{13s} (\bibinfo{year}{2023}), \bibinfo{pages}{1--40}.
\newblock


\bibitem[Sooriya~Patabandige et~al\mbox{.}(2023)]%
        {sooriya2023poster}
\bibfield{author}{\bibinfo{person}{Pramuka~Medaranga Sooriya~Patabandige}, \bibinfo{person}{Steven Antya~Orvala Waskito}, \bibinfo{person}{Kunjun Li}, \bibinfo{person}{Kai~Jie Leow}, \bibinfo{person}{Shantanu Chakrabarty}, {and} \bibinfo{person}{Ambuj Varshney}.} \bibinfo{year}{2023}\natexlab{}.
\newblock \showarticletitle{Poster: rethinking embedded sensor data processing and analysis with large language models}. In \bibinfo{booktitle}{\emph{Proceedings of the 21st Annual International Conference on Mobile Systems, Applications and Services}}. \bibinfo{pages}{561--562}.
\newblock


\bibitem[Sun and Ortiz(2024)]%
        {sun2024ai}
\bibfield{author}{\bibinfo{person}{Yuan Sun} {and} \bibinfo{person}{Jorge Ortiz}.} \bibinfo{year}{2024}\natexlab{}.
\newblock \showarticletitle{An ai-based system utilizing iot-enabled ambient sensors and llms for complex activity tracking}.
\newblock \bibinfo{journal}{\emph{arXiv preprint arXiv:2407.02606}} (\bibinfo{year}{2024}).
\newblock


\bibitem[Tay et~al\mbox{.}(2022)]%
        {tay2022efficient}
\bibfield{author}{\bibinfo{person}{Yi Tay}, \bibinfo{person}{Mostafa Dehghani}, \bibinfo{person}{Dara Bahri}, {and} \bibinfo{person}{Donald Metzler}.} \bibinfo{year}{2022}\natexlab{}.
\newblock \showarticletitle{Efficient transformers: A survey}.
\newblock \bibinfo{journal}{\emph{Comput. Surveys}} \bibinfo{volume}{55}, \bibinfo{number}{6} (\bibinfo{year}{2022}), \bibinfo{pages}{1--28}.
\newblock


\bibitem[Tian et~al\mbox{.}(2025)]%
        {tian2025dailyllm}
\bibfield{author}{\bibinfo{person}{Ye Tian}, \bibinfo{person}{Xiaoyuan Ren}, \bibinfo{person}{Zihao Wang}, \bibinfo{person}{Onat Gungor}, \bibinfo{person}{Xiaofan Yu}, {and} \bibinfo{person}{Tajana Rosing}.} \bibinfo{year}{2025}\natexlab{}.
\newblock \showarticletitle{DailyLLM: Context-Aware Activity Log Generation Using Multi-Modal Sensors and LLMs}.
\newblock \bibinfo{journal}{\emph{arXiv preprint arXiv:2507.13737}} (\bibinfo{year}{2025}).
\newblock


\bibitem[Tian et~al\mbox{.}(2024)]%
        {tian2024edge}
\bibfield{author}{\bibinfo{person}{Yuqing Tian}, \bibinfo{person}{Zhaoyang Zhang}, \bibinfo{person}{Yuzhi Yang}, \bibinfo{person}{Zirui Chen}, \bibinfo{person}{Zhaohui Yang}, \bibinfo{person}{Richeng Jin}, \bibinfo{person}{Tony~QS Quek}, {and} \bibinfo{person}{Kai-Kit Wong}.} \bibinfo{year}{2024}\natexlab{}.
\newblock \showarticletitle{An edge-cloud collaboration framework for generative AI service provision with synergetic big cloud model and small edge models}.
\newblock \bibinfo{journal}{\emph{IEEE Network}} (\bibinfo{year}{2024}).
\newblock


\bibitem[Touvron et~al\mbox{.}(2023)]%
        {touvron2023llama}
\bibfield{author}{\bibinfo{person}{Hugo Touvron}, \bibinfo{person}{Louis Martin}, \bibinfo{person}{Kevin Stone}, \bibinfo{person}{Peter Albert}, \bibinfo{person}{Amjad Almahairi}, \bibinfo{person}{Yasmine Babaei}, \bibinfo{person}{Nikolay Bashlykov}, \bibinfo{person}{Soumya Batra}, \bibinfo{person}{Prajjwal Bhargava}, \bibinfo{person}{Shruti Bhosale}, {et~al\mbox{.}}} \bibinfo{year}{2023}\natexlab{}.
\newblock \showarticletitle{Llama 2: Open foundation and fine-tuned chat models}.
\newblock \bibinfo{journal}{\emph{arXiv preprint arXiv:2307.09288}} (\bibinfo{year}{2023}).
\newblock


\bibitem[Ullah et~al\mbox{.}(2020)]%
        {ullah2020applications}
\bibfield{author}{\bibinfo{person}{Zaib Ullah}, \bibinfo{person}{Fadi Al-Turjman}, \bibinfo{person}{Leonardo Mostarda}, {and} \bibinfo{person}{Roberto Gagliardi}.} \bibinfo{year}{2020}\natexlab{}.
\newblock \showarticletitle{Applications of artificial intelligence and machine learning in smart cities}.
\newblock \bibinfo{journal}{\emph{Computer Communications}}  \bibinfo{volume}{154} (\bibinfo{year}{2020}), \bibinfo{pages}{313--323}.
\newblock


\bibitem[Vaswani et~al\mbox{.}(2017)]%
        {vaswani2017attention}
\bibfield{author}{\bibinfo{person}{Ashish Vaswani}, \bibinfo{person}{Noam Shazeer}, \bibinfo{person}{Niki Parmar}, \bibinfo{person}{Jakob Uszkoreit}, \bibinfo{person}{Llion Jones}, \bibinfo{person}{Aidan~N Gomez}, \bibinfo{person}{{\L}ukasz Kaiser}, {and} \bibinfo{person}{Illia Polosukhin}.} \bibinfo{year}{2017}\natexlab{}.
\newblock \showarticletitle{Attention is all you need}.
\newblock \bibinfo{journal}{\emph{Advances in neural information processing systems}}  \bibinfo{volume}{30} (\bibinfo{year}{2017}).
\newblock


\bibitem[Wan et~al\mbox{.}(2024)]%
        {wan2024meit}
\bibfield{author}{\bibinfo{person}{Zhongwei Wan}, \bibinfo{person}{Che Liu}, \bibinfo{person}{Xin Wang}, \bibinfo{person}{Chaofan Tao}, \bibinfo{person}{Hui Shen}, \bibinfo{person}{Zhenwu Peng}, \bibinfo{person}{Jie Fu}, \bibinfo{person}{Rossella Arcucci}, \bibinfo{person}{Huaxiu Yao}, {and} \bibinfo{person}{Mi Zhang}.} \bibinfo{year}{2024}\natexlab{}.
\newblock \showarticletitle{MEIT: Multi-modal electrocardiogram instruction tuning on large language models for report generation}.
\newblock \bibinfo{journal}{\emph{arXiv preprint arXiv:2403.04945}} (\bibinfo{year}{2024}).
\newblock


\bibitem[Wang et~al\mbox{.}(2024a)]%
        {wang2024privacyoracle}
\bibfield{author}{\bibinfo{person}{Brian Wang}, \bibinfo{person}{Luis~Antonio Garcia}, {and} \bibinfo{person}{Mani Srivastava}.} \bibinfo{year}{2024}\natexlab{a}.
\newblock \showarticletitle{PrivacyOracle: Configuring Sensor Privacy Firewalls with Large Language Models in Smart Built Environments}. In \bibinfo{booktitle}{\emph{2024 IEEE Security and Privacy Workshops (SPW)}}. IEEE, \bibinfo{pages}{239--245}.
\newblock


\bibitem[Wang et~al\mbox{.}(2023)]%
        {wang2023voyager}
\bibfield{author}{\bibinfo{person}{Guanzhi Wang}, \bibinfo{person}{Yuqi Xie}, \bibinfo{person}{Yunfan Jiang}, \bibinfo{person}{Ajay Mandlekar}, \bibinfo{person}{Chaowei Xiao}, \bibinfo{person}{Yuke Zhu}, \bibinfo{person}{Linxi Fan}, {and} \bibinfo{person}{Anima Anandkumar}.} \bibinfo{year}{2023}\natexlab{}.
\newblock \showarticletitle{Voyager: An open-ended embodied agent with large language models}.
\newblock \bibinfo{journal}{\emph{arXiv preprint arXiv:2305.16291}} (\bibinfo{year}{2023}).
\newblock


\bibitem[Wang(2024)]%
        {wang2024hallucination}
\bibfield{author}{\bibinfo{person}{Jue Wang}.} \bibinfo{year}{2024}\natexlab{}.
\newblock \showarticletitle{Hallucination Reduction and Optimization for Large Language Model-Based Autonomous Driving}.
\newblock \bibinfo{journal}{\emph{Symmetry}} \bibinfo{volume}{16}, \bibinfo{number}{9} (\bibinfo{year}{2024}), \bibinfo{pages}{1196}.
\newblock


\bibitem[Wang et~al\mbox{.}(2024b)]%
        {wang2024survey}
\bibfield{author}{\bibinfo{person}{Lei Wang}, \bibinfo{person}{Chen Ma}, \bibinfo{person}{Xueyang Feng}, \bibinfo{person}{Zeyu Zhang}, \bibinfo{person}{Hao Yang}, \bibinfo{person}{Jingsen Zhang}, \bibinfo{person}{Zhiyuan Chen}, \bibinfo{person}{Jiakai Tang}, \bibinfo{person}{Xu Chen}, \bibinfo{person}{Yankai Lin}, {et~al\mbox{.}}} \bibinfo{year}{2024}\natexlab{b}.
\newblock \showarticletitle{A survey on large language model based autonomous agents}.
\newblock \bibinfo{journal}{\emph{Frontiers of Computer Science}} \bibinfo{volume}{18}, \bibinfo{number}{6} (\bibinfo{year}{2024}), \bibinfo{pages}{186345}.
\newblock


\bibitem[Wang et~al\mbox{.}(2024d)]%
        {wang2024omnidrive}
\bibfield{author}{\bibinfo{person}{Shihao Wang}, \bibinfo{person}{Zhiding Yu}, \bibinfo{person}{Xiaohui Jiang}, \bibinfo{person}{Shiyi Lan}, \bibinfo{person}{Min Shi}, \bibinfo{person}{Nadine Chang}, \bibinfo{person}{Jan Kautz}, \bibinfo{person}{Ying Li}, {and} \bibinfo{person}{Jose~M Alvarez}.} \bibinfo{year}{2024}\natexlab{d}.
\newblock \showarticletitle{Omnidrive: A holistic llm-agent framework for autonomous driving with 3d perception, reasoning and planning}.
\newblock \bibinfo{journal}{\emph{arXiv preprint arXiv:2405.01533}} (\bibinfo{year}{2024}).
\newblock


\bibitem[Wang and Qin(2024)]%
        {wang2024intelligent}
\bibfield{author}{\bibinfo{person}{Zihao Wang} {and} \bibinfo{person}{Huijian Qin}.} \bibinfo{year}{2024}\natexlab{}.
\newblock \showarticletitle{Intelligent industrial production process automatic regulation system based on LLM agents}. In \bibinfo{booktitle}{\emph{2024 5th International Conference on Artificial Intelligence and Electromechanical Automation (AIEA)}}. IEEE, \bibinfo{pages}{133--137}.
\newblock


\bibitem[Wang et~al\mbox{.}(2024c)]%
        {wang2024agent}
\bibfield{author}{\bibinfo{person}{Zora~Zhiruo Wang}, \bibinfo{person}{Jiayuan Mao}, \bibinfo{person}{Daniel Fried}, {and} \bibinfo{person}{Graham Neubig}.} \bibinfo{year}{2024}\natexlab{c}.
\newblock \showarticletitle{Agent workflow memory}.
\newblock \bibinfo{journal}{\emph{arXiv preprint arXiv:2409.07429}} (\bibinfo{year}{2024}).
\newblock


\bibitem[Wei et~al\mbox{.}(2024a)]%
        {wei2024systematic}
\bibfield{author}{\bibinfo{person}{Hui Wei}, \bibinfo{person}{Shenghua He}, \bibinfo{person}{Tian Xia}, \bibinfo{person}{Fei Liu}, \bibinfo{person}{Andy Wong}, \bibinfo{person}{Jingyang Lin}, {and} \bibinfo{person}{Mei Han}.} \bibinfo{year}{2024}\natexlab{a}.
\newblock \showarticletitle{Systematic evaluation of llm-as-a-judge in llm alignment tasks: Explainable metrics and diverse prompt templates}.
\newblock \bibinfo{journal}{\emph{arXiv preprint arXiv:2408.13006}} (\bibinfo{year}{2024}).
\newblock


\bibitem[Wei et~al\mbox{.}(2024b)]%
        {wei2024temporally}
\bibfield{author}{\bibinfo{person}{Hui Wei}, \bibinfo{person}{Maxwell~A Xu}, \bibinfo{person}{Colin Samplawski}, \bibinfo{person}{James~M Rehg}, \bibinfo{person}{Santosh Kumar}, {and} \bibinfo{person}{Benjamin~M Marlin}.} \bibinfo{year}{2024}\natexlab{b}.
\newblock \showarticletitle{Temporally Multi-Scale Sparse Self-Attention for Physical Activity Data Imputation}.
\newblock \bibinfo{journal}{\emph{Proceedings of machine learning research}}  \bibinfo{volume}{248} (\bibinfo{year}{2024}), \bibinfo{pages}{137}.
\newblock


\bibitem[Wei et~al\mbox{.}(2025)]%
        {wei2025plangenllms}
\bibfield{author}{\bibinfo{person}{Hui Wei}, \bibinfo{person}{Zihao Zhang}, \bibinfo{person}{Shenghua He}, \bibinfo{person}{Tian Xia}, \bibinfo{person}{Shijia Pan}, {and} \bibinfo{person}{Fei Liu}.} \bibinfo{year}{2025}\natexlab{}.
\newblock \showarticletitle{Plangenllms: A modern survey of llm planning capabilities}.
\newblock \bibinfo{journal}{\emph{arXiv preprint arXiv:2502.11221}} (\bibinfo{year}{2025}).
\newblock


\bibitem[Wei et~al\mbox{.}(2022)]%
        {wei2022chain}
\bibfield{author}{\bibinfo{person}{Jason Wei}, \bibinfo{person}{Xuezhi Wang}, \bibinfo{person}{Dale Schuurmans}, \bibinfo{person}{Maarten Bosma}, \bibinfo{person}{Fei Xia}, \bibinfo{person}{Ed Chi}, \bibinfo{person}{Quoc~V Le}, \bibinfo{person}{Denny Zhou}, {et~al\mbox{.}}} \bibinfo{year}{2022}\natexlab{}.
\newblock \showarticletitle{Chain-of-thought prompting elicits reasoning in large language models}.
\newblock \bibinfo{journal}{\emph{Advances in neural information processing systems}}  \bibinfo{volume}{35} (\bibinfo{year}{2022}), \bibinfo{pages}{24824--24837}.
\newblock


\bibitem[Wen et~al\mbox{.}(2024)]%
        {wen2024poster}
\bibfield{author}{\bibinfo{person}{Hao Wen}, \bibinfo{person}{Wenjie Du}, \bibinfo{person}{Yuanchun Li}, {and} \bibinfo{person}{Yunxin Liu}.} \bibinfo{year}{2024}\natexlab{}.
\newblock \showarticletitle{Poster: Enabling Agent-centric Interaction on Smartphones with LLM-based UI Reassembling}. In \bibinfo{booktitle}{\emph{Proceedings of the 22nd Annual International Conference on Mobile Systems, Applications and Services}}. \bibinfo{pages}{706--707}.
\newblock


\bibitem[Weng(2023)]%
        {weng2023agent}
\bibfield{author}{\bibinfo{person}{Lilian Weng}.} \bibinfo{year}{2023}\natexlab{}.
\newblock \showarticletitle{LLM-powered Autonomous Agents}.
\newblock \bibinfo{journal}{\emph{lilianweng.github.io}} (\bibinfo{date}{Jun} \bibinfo{year}{2023}).
\newblock
\urldef\tempurl%
\url{https://lilianweng.github.io/posts/2023-06-23-agent/}
\showURL{%
\tempurl}


\bibitem[Weng et~al\mbox{.}(2024)]%
        {weng2024large}
\bibfield{author}{\bibinfo{person}{Yuxuan Weng}, \bibinfo{person}{Guoquan Wu}, \bibinfo{person}{Tianyue Zheng}, \bibinfo{person}{Yanbing Yang}, {and} \bibinfo{person}{Jun Luo}.} \bibinfo{year}{2024}\natexlab{}.
\newblock \showarticletitle{Large Model for Small Data: Foundation Model for Cross-Modal RF Human Activity Recognition}. In \bibinfo{booktitle}{\emph{Proceedings of the 22nd ACM Conference on Embedded Networked Sensor Systems}}. \bibinfo{pages}{436--449}.
\newblock


\bibitem[Worae et~al\mbox{.}(2024)]%
        {worae2024unified}
\bibfield{author}{\bibinfo{person}{Daniel~Adu Worae}, \bibinfo{person}{Athar Sheikh}, {and} \bibinfo{person}{Spyridon Mastorakis}.} \bibinfo{year}{2024}\natexlab{}.
\newblock \showarticletitle{A Unified Framework for Context-Aware IoT Management and State-of-the-Art IoT Traffic Anomaly Detection}.
\newblock \bibinfo{journal}{\emph{arXiv preprint arXiv:2412.19830}} (\bibinfo{year}{2024}).
\newblock


\bibitem[Xiao et~al\mbox{.}(2024)]%
        {xiao2024efficient}
\bibfield{author}{\bibinfo{person}{Bin Xiao}, \bibinfo{person}{Burak Kantarci}, \bibinfo{person}{Jiawen Kang}, \bibinfo{person}{Dusit Niyato}, {and} \bibinfo{person}{Mohsen Guizani}.} \bibinfo{year}{2024}\natexlab{}.
\newblock \showarticletitle{Efficient prompting for llm-based generative internet of things}.
\newblock \bibinfo{journal}{\emph{IEEE Internet of Things Journal}} (\bibinfo{year}{2024}).
\newblock


\bibitem[Xiong et~al\mbox{.}(2024)]%
        {xiong2024novel}
\bibfield{author}{\bibinfo{person}{Fuhai Xiong}, \bibinfo{person}{Junxian Wang}, \bibinfo{person}{Yushi Liu}, \bibinfo{person}{Xudong Yan}, \bibinfo{person}{Kamen Ivanov}, \bibinfo{person}{Lei Wang}, {and} \bibinfo{person}{Yan Yan}.} \bibinfo{year}{2024}\natexlab{}.
\newblock \showarticletitle{A Novel Human Activity Recognition Framework Based on Pre-Trained Foundation Model}. In \bibinfo{booktitle}{\emph{2024 IEEE International Conference on Bioinformatics and Biomedicine (BIBM)}}. IEEE, \bibinfo{pages}{5712--5718}.
\newblock


\bibitem[Xiong et~al\mbox{.}(2025)]%
        {xiong2025memory}
\bibfield{author}{\bibinfo{person}{Zidi Xiong}, \bibinfo{person}{Yuping Lin}, \bibinfo{person}{Wenya Xie}, \bibinfo{person}{Pengfei He}, \bibinfo{person}{Jiliang Tang}, \bibinfo{person}{Himabindu Lakkaraju}, {and} \bibinfo{person}{Zhen Xiang}.} \bibinfo{year}{2025}\natexlab{}.
\newblock \showarticletitle{How Memory Management Impacts LLM Agents: An Empirical Study of Experience-Following Behavior}.
\newblock \bibinfo{journal}{\emph{arXiv preprint arXiv:2505.16067}} (\bibinfo{year}{2025}).
\newblock


\bibitem[Xu et~al\mbox{.}(2024a)]%
        {xu2024penetrative}
\bibfield{author}{\bibinfo{person}{Huatao Xu}, \bibinfo{person}{Liying Han}, \bibinfo{person}{Qirui Yang}, \bibinfo{person}{Mo Li}, {and} \bibinfo{person}{Mani Srivastava}.} \bibinfo{year}{2024}\natexlab{a}.
\newblock \showarticletitle{Penetrative ai: Making llms comprehend the physical world}. In \bibinfo{booktitle}{\emph{Proceedings of the 25th International Workshop on Mobile Computing Systems and Applications}}. \bibinfo{pages}{1--7}.
\newblock


\bibitem[Xu et~al\mbox{.}(2021)]%
        {xu2021limu}
\bibfield{author}{\bibinfo{person}{Huatao Xu}, \bibinfo{person}{Pengfei Zhou}, \bibinfo{person}{Rui Tan}, \bibinfo{person}{Mo Li}, {and} \bibinfo{person}{Guobin Shen}.} \bibinfo{year}{2021}\natexlab{}.
\newblock \showarticletitle{Limu-bert: Unleashing the potential of unlabeled data for imu sensing applications}. In \bibinfo{booktitle}{\emph{Proceedings of the 19th ACM Conference on Embedded Networked Sensor Systems}}. \bibinfo{pages}{220--233}.
\newblock


\bibitem[Xu et~al\mbox{.}(2024d)]%
        {xu2024large}
\bibfield{author}{\bibinfo{person}{Shengzhe Xu}, \bibinfo{person}{Christo~Kurisummoottil Thomas}, \bibinfo{person}{Omar Hashash}, \bibinfo{person}{Nikhil Muralidhar}, \bibinfo{person}{Walid Saad}, {and} \bibinfo{person}{Naren Ramakrishnan}.} \bibinfo{year}{2024}\natexlab{d}.
\newblock \showarticletitle{Large multi-modal models (LMMs) as universal foundation models for AI-native wireless systems}.
\newblock \bibinfo{journal}{\emph{IEEE Network}} (\bibinfo{year}{2024}).
\newblock


\bibitem[Xu et~al\mbox{.}(2024b)]%
        {xu2024assuring}
\bibfield{author}{\bibinfo{person}{Weizhe Xu}, \bibinfo{person}{Mengyu Liu}, \bibinfo{person}{Steven Drager}, \bibinfo{person}{Matthew Anderson}, {and} \bibinfo{person}{Fanxin Kong}.} \bibinfo{year}{2024}\natexlab{b}.
\newblock \showarticletitle{Assuring LLM-Enabled Cyber-Physical Systems}. In \bibinfo{booktitle}{\emph{2024 ACM/IEEE 15th International Conference on Cyber-Physical Systems (ICCPS)}}. IEEE, \bibinfo{pages}{287--288}.
\newblock


\bibitem[Xu et~al\mbox{.}(2024c)]%
        {xu2024llm}
\bibfield{author}{\bibinfo{person}{Weizhe Xu}, \bibinfo{person}{Mengyu Liu}, \bibinfo{person}{Oleg Sokolsky}, \bibinfo{person}{Insup Lee}, {and} \bibinfo{person}{Fanxin Kong}.} \bibinfo{year}{2024}\natexlab{c}.
\newblock \showarticletitle{LLM-enabled cyber-physical systems: survey, research opportunities, and challenges}. In \bibinfo{booktitle}{\emph{2024 IEEE International Workshop on Foundation Models for Cyber-Physical Systems \& Internet of Things (FMSys)}}. IEEE, \bibinfo{pages}{50--55}.
\newblock


\bibitem[Xu et~al\mbox{.}(2025)]%
        {xu2025mem}
\bibfield{author}{\bibinfo{person}{Wujiang Xu}, \bibinfo{person}{Kai Mei}, \bibinfo{person}{Hang Gao}, \bibinfo{person}{Juntao Tan}, \bibinfo{person}{Zujie Liang}, {and} \bibinfo{person}{Yongfeng Zhang}.} \bibinfo{year}{2025}\natexlab{}.
\newblock \showarticletitle{A-mem: Agentic memory for llm agents}.
\newblock \bibinfo{journal}{\emph{arXiv preprint arXiv:2502.12110}} (\bibinfo{year}{2025}).
\newblock


\bibitem[Xue et~al\mbox{.}(2024)]%
        {xue2024leveraging}
\bibfield{author}{\bibinfo{person}{Dinghao Xue}, \bibinfo{person}{Xiaoran Fan}, \bibinfo{person}{Tao Chen}, \bibinfo{person}{Guohao Lan}, {and} \bibinfo{person}{Qun Song}.} \bibinfo{year}{2024}\natexlab{}.
\newblock \showarticletitle{Leveraging Foundation Models for Zero-Shot IoT Sensing}.
\newblock \bibinfo{journal}{\emph{arXiv preprint arXiv:2407.19893}} (\bibinfo{year}{2024}).
\newblock


\bibitem[Yacchirema et~al\mbox{.}(2019)]%
        {yacchirema2019fall}
\bibfield{author}{\bibinfo{person}{Diana Yacchirema}, \bibinfo{person}{Jara~Su{\'a}rez de Puga}, \bibinfo{person}{Carlos Palau}, {and} \bibinfo{person}{Manuel Esteve}.} \bibinfo{year}{2019}\natexlab{}.
\newblock \showarticletitle{Fall detection system for elderly people using IoT and ensemble machine learning algorithm}.
\newblock \bibinfo{journal}{\emph{Personal and Ubiquitous Computing}} \bibinfo{volume}{23}, \bibinfo{number}{5} (\bibinfo{year}{2019}), \bibinfo{pages}{801--817}.
\newblock


\bibitem[Yan et~al\mbox{.}(2024)]%
        {yan2024language}
\bibfield{author}{\bibinfo{person}{Hua Yan}, \bibinfo{person}{Heng Tan}, \bibinfo{person}{Yi Ding}, \bibinfo{person}{Pengfei Zhou}, \bibinfo{person}{Vinod Namboodiri}, {and} \bibinfo{person}{Yu Yang}.} \bibinfo{year}{2024}\natexlab{}.
\newblock \showarticletitle{Language-centered Human Activity Recognition}.
\newblock \bibinfo{journal}{\emph{arXiv preprint arXiv:2410.00003}} (\bibinfo{year}{2024}).
\newblock


\bibitem[Yang et~al\mbox{.}(2025a)]%
        {yang2025socialmind}
\bibfield{author}{\bibinfo{person}{Bufang Yang}, \bibinfo{person}{Yunqi Guo}, \bibinfo{person}{Lilin Xu}, \bibinfo{person}{Zhenyu Yan}, \bibinfo{person}{Hongkai Chen}, \bibinfo{person}{Guoliang Xing}, {and} \bibinfo{person}{Xiaofan Jiang}.} \bibinfo{year}{2025}\natexlab{a}.
\newblock \showarticletitle{Socialmind: Llm-based proactive ar social assistive system with human-like perception for in-situ live interactions}.
\newblock \bibinfo{journal}{\emph{Proceedings of the ACM on Interactive, Mobile, Wearable and Ubiquitous Technologies}} \bibinfo{volume}{9}, \bibinfo{number}{1} (\bibinfo{year}{2025}), \bibinfo{pages}{1--30}.
\newblock


\bibitem[Yang et~al\mbox{.}(2023a)]%
        {yang2023edgefm}
\bibfield{author}{\bibinfo{person}{Bufang Yang}, \bibinfo{person}{Lixing He}, \bibinfo{person}{Neiwen Ling}, \bibinfo{person}{Zhenyu Yan}, \bibinfo{person}{Guoliang Xing}, \bibinfo{person}{Xian Shuai}, \bibinfo{person}{Xiaozhe Ren}, {and} \bibinfo{person}{Xin Jiang}.} \bibinfo{year}{2023}\natexlab{a}.
\newblock \showarticletitle{Edgefm: Leveraging foundation model for open-set learning on the edge}. In \bibinfo{booktitle}{\emph{Proceedings of the 21st ACM Conference on Embedded Networked Sensor Systems}}. \bibinfo{pages}{111--124}.
\newblock


\bibitem[Yang et~al\mbox{.}(2024a)]%
        {yang2024drhouse}
\bibfield{author}{\bibinfo{person}{Bufang Yang}, \bibinfo{person}{Siyang Jiang}, \bibinfo{person}{Lilin Xu}, \bibinfo{person}{Kaiwei Liu}, \bibinfo{person}{Hai Li}, \bibinfo{person}{Guoliang Xing}, \bibinfo{person}{Hongkai Chen}, \bibinfo{person}{Xiaofan Jiang}, {and} \bibinfo{person}{Zhenyu Yan}.} \bibinfo{year}{2024}\natexlab{a}.
\newblock \showarticletitle{Drhouse: An llm-empowered diagnostic reasoning system through harnessing outcomes from sensor data and expert knowledge}.
\newblock \bibinfo{journal}{\emph{Proceedings of the ACM on Interactive, Mobile, Wearable and Ubiquitous Technologies}} \bibinfo{volume}{8}, \bibinfo{number}{4} (\bibinfo{year}{2024}), \bibinfo{pages}{1--29}.
\newblock


\bibitem[Yang et~al\mbox{.}(2025b)]%
        {yang2025contextagent}
\bibfield{author}{\bibinfo{person}{Bufang Yang}, \bibinfo{person}{Lilin Xu}, \bibinfo{person}{Liekang Zeng}, \bibinfo{person}{Kaiwei Liu}, \bibinfo{person}{Siyang Jiang}, \bibinfo{person}{Wenrui Lu}, \bibinfo{person}{Hongkai Chen}, \bibinfo{person}{Xiaofan Jiang}, \bibinfo{person}{Guoliang Xing}, {and} \bibinfo{person}{Zhenyu Yan}.} \bibinfo{year}{2025}\natexlab{b}.
\newblock \showarticletitle{ContextAgent: Context-Aware Proactive {LLM} Agents with Open-World Sensory Perceptions}.
\newblock \bibinfo{journal}{\emph{CoRR}}  \bibinfo{volume}{abs/2505.14668} (\bibinfo{year}{2025}).
\newblock
\href{https://doi.org/10.48550/ARXIV.2505.14668}{doi:\nolinkurl{10.48550/ARXIV.2505.14668}}
\showeprint[arXiv]{2505.14668}


\bibitem[Yang et~al\mbox{.}(2024c)]%
        {yang2024llm}
\bibfield{author}{\bibinfo{person}{Hanqing Yang}, \bibinfo{person}{Marie Siew}, {and} \bibinfo{person}{Carlee Joe-Wong}.} \bibinfo{year}{2024}\natexlab{c}.
\newblock \showarticletitle{An llm-based digital twin for optimizing human-in-the loop systems}. In \bibinfo{booktitle}{\emph{2024 IEEE International Workshop on Foundation Models for Cyber-Physical Systems \& Internet of Things (FMSys)}}. IEEE, \bibinfo{pages}{26--31}.
\newblock


\bibitem[Yang et~al\mbox{.}(2023b)]%
        {yang2023diffusion}
\bibfield{author}{\bibinfo{person}{Ling Yang}, \bibinfo{person}{Zhilong Zhang}, \bibinfo{person}{Yang Song}, \bibinfo{person}{Shenda Hong}, \bibinfo{person}{Runsheng Xu}, \bibinfo{person}{Yue Zhao}, \bibinfo{person}{Wentao Zhang}, \bibinfo{person}{Bin Cui}, {and} \bibinfo{person}{Ming-Hsuan Yang}.} \bibinfo{year}{2023}\natexlab{b}.
\newblock \showarticletitle{Diffusion models: A comprehensive survey of methods and applications}.
\newblock \bibinfo{journal}{\emph{Comput. Surveys}} \bibinfo{volume}{56}, \bibinfo{number}{4} (\bibinfo{year}{2023}), \bibinfo{pages}{1--39}.
\newblock


\bibitem[Yang and Ardakanian(2023)]%
        {yang2023privacy}
\bibfield{author}{\bibinfo{person}{Xin Yang} {and} \bibinfo{person}{Omid Ardakanian}.} \bibinfo{year}{2023}\natexlab{}.
\newblock \showarticletitle{Privacy through Diffusion: A White-listing Approach to Sensor Data Anonymization}. In \bibinfo{booktitle}{\emph{Proceedings of the 5th Workshop on CPS\&IoT Security and Privacy}}. \bibinfo{pages}{101--107}.
\newblock


\bibitem[Yang et~al\mbox{.}(2024d)]%
        {yang2024joint}
\bibfield{author}{\bibinfo{person}{Yunhao Yang}, \bibinfo{person}{William Ward}, \bibinfo{person}{Zichao Hu}, \bibinfo{person}{Joydeep Biswas}, {and} \bibinfo{person}{Ufuk Topcu}.} \bibinfo{year}{2024}\natexlab{d}.
\newblock \showarticletitle{Joint Verification and Refinement of Language Models for Safety-Constrained Planning}.
\newblock \bibinfo{journal}{\emph{arXiv preprint arXiv:2410.14865}} (\bibinfo{year}{2024}).
\newblock


\bibitem[Yang et~al\mbox{.}(2024b)]%
        {yang2024plug}
\bibfield{author}{\bibinfo{person}{Ziyi Yang}, \bibinfo{person}{Shreyas~S Raman}, \bibinfo{person}{Ankit Shah}, {and} \bibinfo{person}{Stefanie Tellex}.} \bibinfo{year}{2024}\natexlab{b}.
\newblock \showarticletitle{Plug in the safety chip: Enforcing constraints for llm-driven robot agents}. In \bibinfo{booktitle}{\emph{2024 IEEE International Conference on Robotics and Automation (ICRA)}}. IEEE, \bibinfo{pages}{14435--14442}.
\newblock


\bibitem[Yao et~al\mbox{.}(2022)]%
        {yao2022edge}
\bibfield{author}{\bibinfo{person}{Jiangchao Yao}, \bibinfo{person}{Shengyu Zhang}, \bibinfo{person}{Yang Yao}, \bibinfo{person}{Feng Wang}, \bibinfo{person}{Jianxin Ma}, \bibinfo{person}{Jianwei Zhang}, \bibinfo{person}{Yunfei Chu}, \bibinfo{person}{Luo Ji}, \bibinfo{person}{Kunyang Jia}, \bibinfo{person}{Tao Shen}, {et~al\mbox{.}}} \bibinfo{year}{2022}\natexlab{}.
\newblock \showarticletitle{Edge-cloud polarization and collaboration: A comprehensive survey for ai}.
\newblock \bibinfo{journal}{\emph{IEEE Transactions on Knowledge and Data Engineering}} \bibinfo{volume}{35}, \bibinfo{number}{7} (\bibinfo{year}{2022}), \bibinfo{pages}{6866--6886}.
\newblock


\bibitem[Yao et~al\mbox{.}(2023)]%
        {yao2023tree}
\bibfield{author}{\bibinfo{person}{Shunyu Yao}, \bibinfo{person}{Dian Yu}, \bibinfo{person}{Jeffrey Zhao}, \bibinfo{person}{Izhak Shafran}, \bibinfo{person}{Tom Griffiths}, \bibinfo{person}{Yuan Cao}, {and} \bibinfo{person}{Karthik Narasimhan}.} \bibinfo{year}{2023}\natexlab{}.
\newblock \showarticletitle{Tree of thoughts: Deliberate problem solving with large language models}.
\newblock \bibinfo{journal}{\emph{Advances in neural information processing systems}}  \bibinfo{volume}{36} (\bibinfo{year}{2023}), \bibinfo{pages}{11809--11822}.
\newblock


\bibitem[Ye et~al\mbox{.}(2024)]%
        {ye2024justice}
\bibfield{author}{\bibinfo{person}{Jiayi Ye}, \bibinfo{person}{Yanbo Wang}, \bibinfo{person}{Yue Huang}, \bibinfo{person}{Dongping Chen}, \bibinfo{person}{Qihui Zhang}, \bibinfo{person}{Nuno Moniz}, \bibinfo{person}{Tian Gao}, \bibinfo{person}{Werner Geyer}, \bibinfo{person}{Chao Huang}, \bibinfo{person}{Pin-Yu Chen}, {et~al\mbox{.}}} \bibinfo{year}{2024}\natexlab{}.
\newblock \showarticletitle{Justice or prejudice? quantifying biases in llm-as-a-judge}.
\newblock \bibinfo{journal}{\emph{arXiv preprint arXiv:2410.02736}} (\bibinfo{year}{2024}).
\newblock


\bibitem[Yonekura et~al\mbox{.}(2024)]%
        {yonekura2024generating}
\bibfield{author}{\bibinfo{person}{Haruki Yonekura}, \bibinfo{person}{Fukuharu Tanaka}, \bibinfo{person}{Teruhiro Mizumoto}, {and} \bibinfo{person}{Hirozumi Yamaguchi}.} \bibinfo{year}{2024}\natexlab{}.
\newblock \showarticletitle{Generating Human Daily Activities with LLM for Smart Home Simulator Agents}. In \bibinfo{booktitle}{\emph{2024 International Conference on Intelligent Environments (IE)}}. IEEE, \bibinfo{pages}{93--96}.
\newblock


\bibitem[Yuan et~al\mbox{.}(2023)]%
        {yuan2023distilling}
\bibfield{author}{\bibinfo{person}{Siyu Yuan}, \bibinfo{person}{Jiangjie Chen}, \bibinfo{person}{Ziquan Fu}, \bibinfo{person}{Xuyang Ge}, \bibinfo{person}{Soham Shah}, \bibinfo{person}{Charles~Robert Jankowski}, \bibinfo{person}{Yanghua Xiao}, {and} \bibinfo{person}{Deqing Yang}.} \bibinfo{year}{2023}\natexlab{}.
\newblock \showarticletitle{Distilling script knowledge from large language models for constrained language planning}.
\newblock \bibinfo{journal}{\emph{arXiv preprint arXiv:2305.05252}} (\bibinfo{year}{2023}).
\newblock


\bibitem[Yuan et~al\mbox{.}(2024)]%
        {yuan2024wip}
\bibfield{author}{\bibinfo{person}{Yizhen Yuan}, \bibinfo{person}{Rui Kong}, \bibinfo{person}{Yuanchun Li}, {and} \bibinfo{person}{Yunxin Liu}.} \bibinfo{year}{2024}\natexlab{}.
\newblock \showarticletitle{Wip: An on-device llm-based approach to query privacy protection}. In \bibinfo{booktitle}{\emph{Proceedings of the Workshop on Edge and Mobile Foundation Models}}. \bibinfo{pages}{7--9}.
\newblock


\bibitem[Zeng et~al\mbox{.}(2023)]%
        {zeng2023transformers}
\bibfield{author}{\bibinfo{person}{Ailing Zeng}, \bibinfo{person}{Muxi Chen}, \bibinfo{person}{Lei Zhang}, {and} \bibinfo{person}{Qiang Xu}.} \bibinfo{year}{2023}\natexlab{}.
\newblock \showarticletitle{Are transformers effective for time series forecasting?}. In \bibinfo{booktitle}{\emph{Proceedings of the AAAI conference on artificial intelligence}}, Vol.~\bibinfo{volume}{37}. \bibinfo{pages}{11121--11128}.
\newblock


\bibitem[Zeng et~al\mbox{.}(2024)]%
        {zeng2024structural}
\bibfield{author}{\bibinfo{person}{Ruihong Zeng}, \bibinfo{person}{Jinyuan Fang}, \bibinfo{person}{Siwei Liu}, {and} \bibinfo{person}{Zaiqiao Meng}.} \bibinfo{year}{2024}\natexlab{}.
\newblock \showarticletitle{On the structural memory of llm agents}.
\newblock \bibinfo{journal}{\emph{arXiv preprint arXiv:2412.15266}} (\bibinfo{year}{2024}).
\newblock


\bibitem[Zhang and Tao(2020)]%
        {zhang2020empowering}
\bibfield{author}{\bibinfo{person}{Jing Zhang} {and} \bibinfo{person}{Dacheng Tao}.} \bibinfo{year}{2020}\natexlab{}.
\newblock \showarticletitle{Empowering things with intelligence: a survey of the progress, challenges, and opportunities in artificial intelligence of things}.
\newblock \bibinfo{journal}{\emph{IEEE Internet of Things Journal}} \bibinfo{volume}{8}, \bibinfo{number}{10} (\bibinfo{year}{2020}), \bibinfo{pages}{7789--7817}.
\newblock


\bibitem[Zhang et~al\mbox{.}(2024d)]%
        {zhang2024mambareid}
\bibfield{author}{\bibinfo{person}{Ruijuan Zhang}, \bibinfo{person}{Lizhong Xu}, \bibinfo{person}{Song Yang}, {and} \bibinfo{person}{Li Wang}.} \bibinfo{year}{2024}\natexlab{d}.
\newblock \showarticletitle{MambaReID: Exploiting Vision Mamba for Multi-Modal Object Re-Identification}.
\newblock \bibinfo{journal}{\emph{Sensors}} \bibinfo{volume}{24}, \bibinfo{number}{14} (\bibinfo{year}{2024}), \bibinfo{pages}{4639}.
\newblock


\bibitem[Zhang et~al\mbox{.}(2024b)]%
        {zhang2024large}
\bibfield{author}{\bibinfo{person}{Xiyuan Zhang}, \bibinfo{person}{Ranak~Roy Chowdhury}, \bibinfo{person}{Rajesh~K Gupta}, {and} \bibinfo{person}{Jingbo Shang}.} \bibinfo{year}{2024}\natexlab{b}.
\newblock \showarticletitle{Large language models for time series: A survey}.
\newblock \bibinfo{journal}{\emph{arXiv preprint arXiv:2402.01801}} (\bibinfo{year}{2024}).
\newblock


\bibitem[Zhang et~al\mbox{.}(2024c)]%
        {zhang2024llm}
\bibfield{author}{\bibinfo{person}{Yadong Zhang}, \bibinfo{person}{Shaoguang Mao}, \bibinfo{person}{Tao Ge}, \bibinfo{person}{Xun Wang}, \bibinfo{person}{Adrian de Wynter}, \bibinfo{person}{Yan Xia}, \bibinfo{person}{Wenshan Wu}, \bibinfo{person}{Ting Song}, \bibinfo{person}{Man Lan}, {and} \bibinfo{person}{Furu Wei}.} \bibinfo{year}{2024}\natexlab{c}.
\newblock \showarticletitle{Llm as a mastermind: A survey of strategic reasoning with large language models}.
\newblock \bibinfo{journal}{\emph{arXiv preprint arXiv:2404.01230}} (\bibinfo{year}{2024}).
\newblock


\bibitem[Zhang et~al\mbox{.}(2024a)]%
        {zhang2024survey}
\bibfield{author}{\bibinfo{person}{Zeyu Zhang}, \bibinfo{person}{Xiaohe Bo}, \bibinfo{person}{Chen Ma}, \bibinfo{person}{Rui Li}, \bibinfo{person}{Xu Chen}, \bibinfo{person}{Quanyu Dai}, \bibinfo{person}{Jieming Zhu}, \bibinfo{person}{Zhenhua Dong}, {and} \bibinfo{person}{Ji-Rong Wen}.} \bibinfo{year}{2024}\natexlab{a}.
\newblock \showarticletitle{A survey on the memory mechanism of large language model based agents}.
\newblock \bibinfo{journal}{\emph{arXiv preprint arXiv:2404.13501}} (\bibinfo{year}{2024}).
\newblock


\bibitem[Zheng et~al\mbox{.}(2023)]%
        {zheng2023judging}
\bibfield{author}{\bibinfo{person}{Lianmin Zheng}, \bibinfo{person}{Wei-Lin Chiang}, \bibinfo{person}{Ying Sheng}, \bibinfo{person}{Siyuan Zhuang}, \bibinfo{person}{Zhanghao Wu}, \bibinfo{person}{Yonghao Zhuang}, \bibinfo{person}{Zi Lin}, \bibinfo{person}{Zhuohan Li}, \bibinfo{person}{Dacheng Li}, \bibinfo{person}{Eric Xing}, {et~al\mbox{.}}} \bibinfo{year}{2023}\natexlab{}.
\newblock \showarticletitle{Judging llm-as-a-judge with mt-bench and chatbot arena}.
\newblock \bibinfo{journal}{\emph{Advances in Neural Information Processing Systems}}  \bibinfo{volume}{36} (\bibinfo{year}{2023}), \bibinfo{pages}{46595--46623}.
\newblock


\bibitem[Zhong et~al\mbox{.}(2024b)]%
        {zhong2024casit}
\bibfield{author}{\bibinfo{person}{Ningze Zhong}, \bibinfo{person}{Yi Wang}, \bibinfo{person}{Rui Xiong}, \bibinfo{person}{Yingyue Zheng}, \bibinfo{person}{Yang Li}, \bibinfo{person}{Mingjun Ouyang}, \bibinfo{person}{Dan Shen}, {and} \bibinfo{person}{Xiangwei Zhu}.} \bibinfo{year}{2024}\natexlab{b}.
\newblock \showarticletitle{Casit: Collective intelligent agent system for internet of things}.
\newblock \bibinfo{journal}{\emph{IEEE Internet of Things Journal}} \bibinfo{volume}{11}, \bibinfo{number}{11} (\bibinfo{year}{2024}), \bibinfo{pages}{19646--19656}.
\newblock


\bibitem[Zhong et~al\mbox{.}(2024a)]%
        {zhong2024memorybank}
\bibfield{author}{\bibinfo{person}{Wanjun Zhong}, \bibinfo{person}{Lianghong Guo}, \bibinfo{person}{Qiqi Gao}, \bibinfo{person}{He Ye}, {and} \bibinfo{person}{Yanlin Wang}.} \bibinfo{year}{2024}\natexlab{a}.
\newblock \showarticletitle{Memorybank: Enhancing large language models with long-term memory}. In \bibinfo{booktitle}{\emph{Proceedings of the AAAI Conference on Artificial Intelligence}}, Vol.~\bibinfo{volume}{38}. \bibinfo{pages}{19724--19731}.
\newblock


\bibitem[Zhou et~al\mbox{.}(2024)]%
        {zhou2024comprehensive}
\bibfield{author}{\bibinfo{person}{Ce Zhou}, \bibinfo{person}{Qian Li}, \bibinfo{person}{Chen Li}, \bibinfo{person}{Jun Yu}, \bibinfo{person}{Yixin Liu}, \bibinfo{person}{Guangjing Wang}, \bibinfo{person}{Kai Zhang}, \bibinfo{person}{Cheng Ji}, \bibinfo{person}{Qiben Yan}, \bibinfo{person}{Lifang He}, {et~al\mbox{.}}} \bibinfo{year}{2024}\natexlab{}.
\newblock \showarticletitle{A comprehensive survey on pretrained foundation models: A history from bert to chatgpt}.
\newblock \bibinfo{journal}{\emph{International Journal of Machine Learning and Cybernetics}} (\bibinfo{year}{2024}), \bibinfo{pages}{1--65}.
\newblock


\bibitem[Zhuang et~al\mbox{.}(2024)]%
        {zhuang2024litemoe}
\bibfield{author}{\bibinfo{person}{Yan Zhuang}, \bibinfo{person}{Zhenzhe Zheng}, \bibinfo{person}{Fan Wu}, {and} \bibinfo{person}{Guihai Chen}.} \bibinfo{year}{2024}\natexlab{}.
\newblock \showarticletitle{LiteMoE: Customizing On-device LLM Serving via Proxy Submodel Tuning}. In \bibinfo{booktitle}{\emph{Proceedings of the 22nd ACM Conference on Embedded Networked Sensor Systems}}. \bibinfo{pages}{521--534}.
\newblock


\bibitem[Zhuge et~al\mbox{.}(2024)]%
        {zhuge2024agent}
\bibfield{author}{\bibinfo{person}{Mingchen Zhuge}, \bibinfo{person}{Changsheng Zhao}, \bibinfo{person}{Dylan Ashley}, \bibinfo{person}{Wenyi Wang}, \bibinfo{person}{Dmitrii Khizbullin}, \bibinfo{person}{Yunyang Xiong}, \bibinfo{person}{Zechun Liu}, \bibinfo{person}{Ernie Chang}, \bibinfo{person}{Raghuraman Krishnamoorthi}, \bibinfo{person}{Yuandong Tian}, {et~al\mbox{.}}} \bibinfo{year}{2024}\natexlab{}.
\newblock \showarticletitle{Agent-as-a-judge: Evaluate agents with agents}.
\newblock \bibinfo{journal}{\emph{arXiv preprint arXiv:2410.10934}} (\bibinfo{year}{2024}).
\newblock


\bibitem[Zong et~al\mbox{.}(2025)]%
        {zong2025integrating}
\bibfield{author}{\bibinfo{person}{Mingyu Zong}, \bibinfo{person}{Arvin Hekmati}, \bibinfo{person}{Michael Guastalla}, \bibinfo{person}{Yiyi Li}, {and} \bibinfo{person}{Bhaskar Krishnamachari}.} \bibinfo{year}{2025}\natexlab{}.
\newblock \showarticletitle{Integrating large language models with internet of things: applications}.
\newblock \bibinfo{journal}{\emph{Discover Internet of Things}} \bibinfo{volume}{5}, \bibinfo{number}{1} (\bibinfo{year}{2025}), \bibinfo{pages}{2}.
\newblock


\end{thebibliography}
